\documentclass[10pt,twocolumn,letterpaper]{article}

\usepackage[pagenumbers]{iccv} %

\def\paperID{4759} 
\def\confName{ICCV}
\def\confYear{2025}
\definecolor{cvprblue}{rgb}{0.21,0.49,0.74}

\def\shortTitle{C2OT}
\def\paperTitle{The Curse of Conditions: Analyzing and Improving Optimal Transport for Conditional Flow-Based Generation}

\newcommand*\ourmethod{C\(^2\)OT}
\newcommand*\generatedsample{\tilde{x}_1}
\newcommand*\condition{c}
\newcommand*\flow{v_\theta}
\newcommand*\conditionedflow{v_{\theta, \condition}}
\newcommand*\jointdistribution{\pi}
\newcommand*\batchsize{b}
\newcommand*\netbatchsize{b_n}
\newcommand*\datadim{e}
\newcommand*\conditiondim{h}
\newcommand*\otjointdistribution{\pi_{\text{ot}}}
\newcommand*\diff{\mathop{}\!\mathrm{d}}
\newcommand*\ourjointdistribution{\pi_{\text{c}^2\text{ot}}}
\newcommand*\numlabels{k}
\newcommand*\discretef{f_{\text{d}}}

\newcommand*\conditionweight{w}
\newcommand*\conditionratio{r}
\newcommand*\targetratio{r_{\text{tar}}}
\newcommand*\continuousf{f_{\text{c}}}
\newcommand*\cosinedist{\text{cdist}}

\usepackage[utf8]{inputenc} %
\usepackage[T1]{fontenc}    %
\usepackage[pagebackref,breaklinks,colorlinks,allcolors=cvprblue]{hyperref}
\usepackage{url}            %
\usepackage{booktabs}       %
\usepackage{amsfonts}       %
\usepackage{nicefrac}       %
\usepackage{microtype}      %
\usepackage[dvipsnames]{xcolor}         %
\usepackage{notoccite}
\usepackage{titletoc}
\usepackage[bb=dsserif]{mathalpha}
\usepackage{algorithm}
\usepackage{algorithmicx}
\usepackage[noend]{algpseudocode}
\usepackage{nicefrac}

\usepackage{colortbl}
\usepackage{graphicx,overpic}
\usepackage[export]{adjustbox}
\usepackage{amsmath, amssymb}
\usepackage{multirow}
\usepackage[super]{nth}
\usepackage{comment}
\usepackage{nicematrix}

\usepackage{enumitem}
\usepackage{caption}

\usepackage{titlesec}

\titleformat{\section}
{\normalfont\large\bfseries}{\thesection.~}{0pt}{}{}
\titleformat{\subsection}
{\normalfont\large\bfseries}{\thesubsection.~}{0pt}{}{}
\titleformat{\subsubsection}
{\normalfont\normalsize\bfseries}{\thesubsubsection.~}{0pt}{}{}
\titleformat{\paragraph}[runin]
{\normalfont\normalsize\bfseries}{\theparagraph\vspace{1em}}{}{}

\titlespacing*{\paragraph}{0em}{.5ex plus .5ex minus .3ex}{1em}

\titlespacing*{\section}{0pt}{3.5ex plus 1ex minus .2ex}{2.3ex plus .2ex}
\titlespacing*{\subsection}{0pt}{3.25ex plus 1ex minus .2ex}{1.5ex plus .2ex}
\titlespacing*{\subsubsection}{0pt}{3.25ex plus 1ex minus .2ex}{1.5ex plus .2ex}

\usepackage{tikz}
\usepackage{pgfplots}
\usepackage{pgfplotstable}
\pgfplotsset{compat=1.13}
\usetikzlibrary{pgfplots.groupplots}
\usetikzlibrary{calc,arrows}
\usepackage{mathtools}
\usepgfplotslibrary{fillbetween}

\usepackage{tabularx,stackengine,collcell}
\let\endminwd\relax
\newcolumntype{L}[1]{>{\collectcell\xminwd l{#1}}l<{\endminwd\endcollectcell}}
\newcolumntype{C}[1]{>{\collectcell\xminwd c{#1}}c<{\endminwd\endcollectcell}}
\newcolumntype{R}[1]{>{\collectcell\xminwd r{#1}}r<{\endminwd\endcollectcell}}
\def\minwd#1#2#3\endminwd{\stackengine{0pt}{#3}{\rule{#2}{0pt}}{O}{#1}{F}{F}{L}}
\newcommand\xminwd[1]{\minwd#1}

\hypersetup{
	colorlinks,
	linkcolor={red!80!black},
	citecolor={cvprblue},
	urlcolor={cvprblue},
    pdftitle={\shortTitle{}: \paperTitle{}},
}

\definecolor{defaultColor}{RGB}{230, 244, 252}

\newcommand{\beginsupplement}{
    \appendix
	\setcounter{table}{0}
	\renewcommand{\thetable}{A\arabic{table}}%
	\setcounter{figure}{0}
	\renewcommand{\thefigure}{A\arabic{figure}}%
	\setcounter{equation}{0}
	\renewcommand{\theequation}{A\arabic{equation}}
}

\title{\paperTitle{}}

\author{
Ho Kei Cheng\hspace{2em}
Alexander Schwing
\\
University of Illinois Urbana-Champaign
\\
{\tt\scriptsize
\{hokeikc2,aschwing\}@illinois.edu
}
}

\usepackage{etoolbox}

\makeatletter
\apptocmd{\@maketitle}{\centering
\vspace{-4ex}
\captionsetup{type=figure}
\begin{NiceTabular}{c@{}c@{}c@{}|@{}c@{}c@{}c@{}|@{}c@{}c@{}c}
& \Block{1-2}{\small Unconditional} & & \Block{1-3}{\small With discrete conditions} & & & \Block{1-3}{\small With continuous conditions \\ \small (target \(x\) coordinates)}
\\
\cmidrule(lr{\dimexpr 4\tabcolsep-16pt}){2-3}
\cmidrule(lr{\dimexpr 4\tabcolsep-16pt}){4-6}
\cmidrule(lr{\dimexpr 4\tabcolsep-16pt}){7-9}
\Block{2-1}{\rotate \small Euler-1 step}
&
\includegraphics[width=0.118\linewidth]{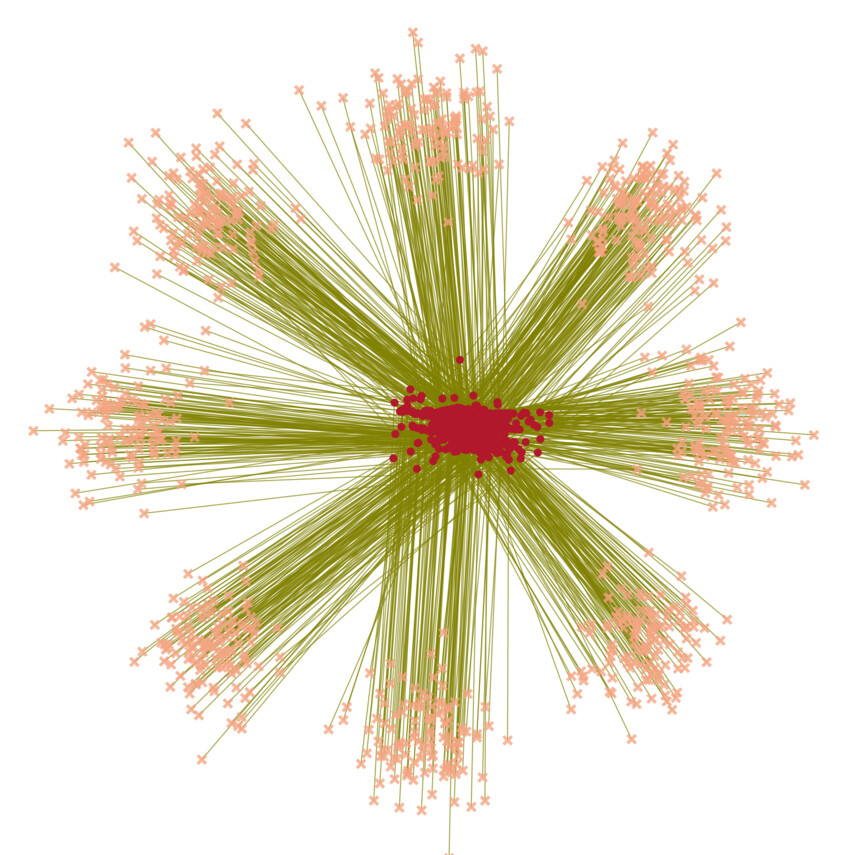} & 
\includegraphics[width=0.118\linewidth]{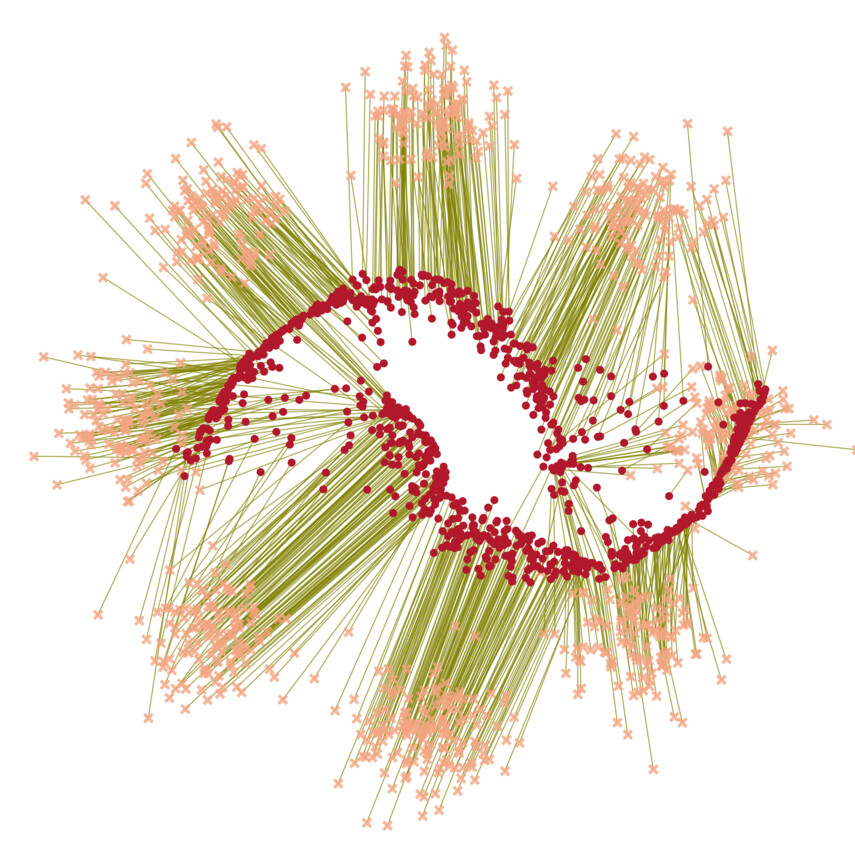} & 
\includegraphics[width=0.118\linewidth]{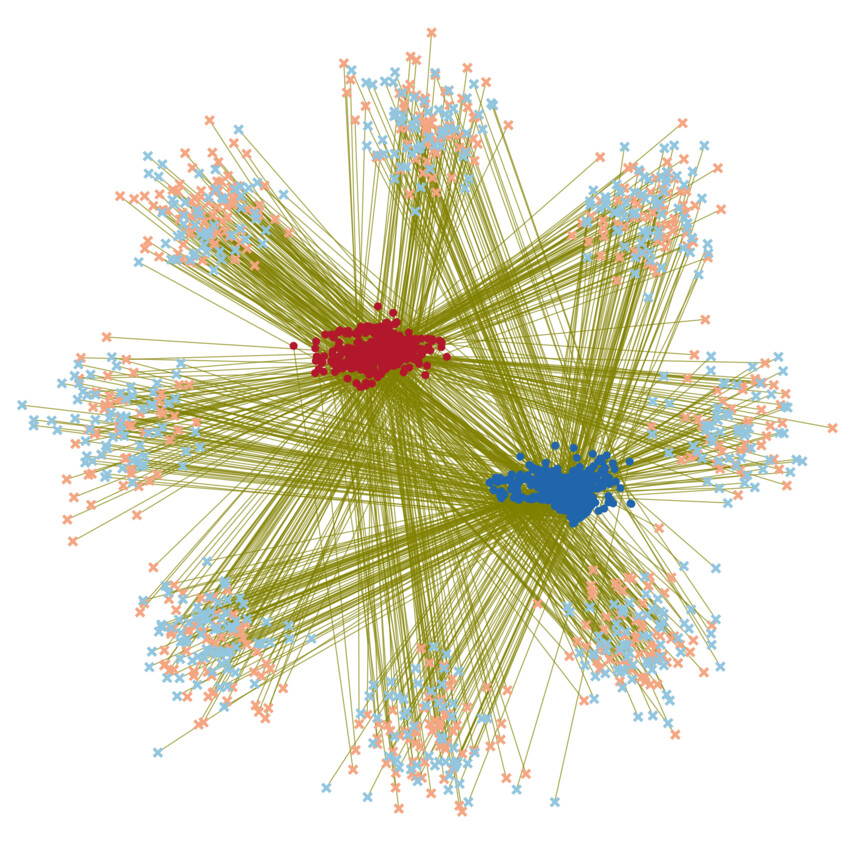} & 
\includegraphics[width=0.118\linewidth]{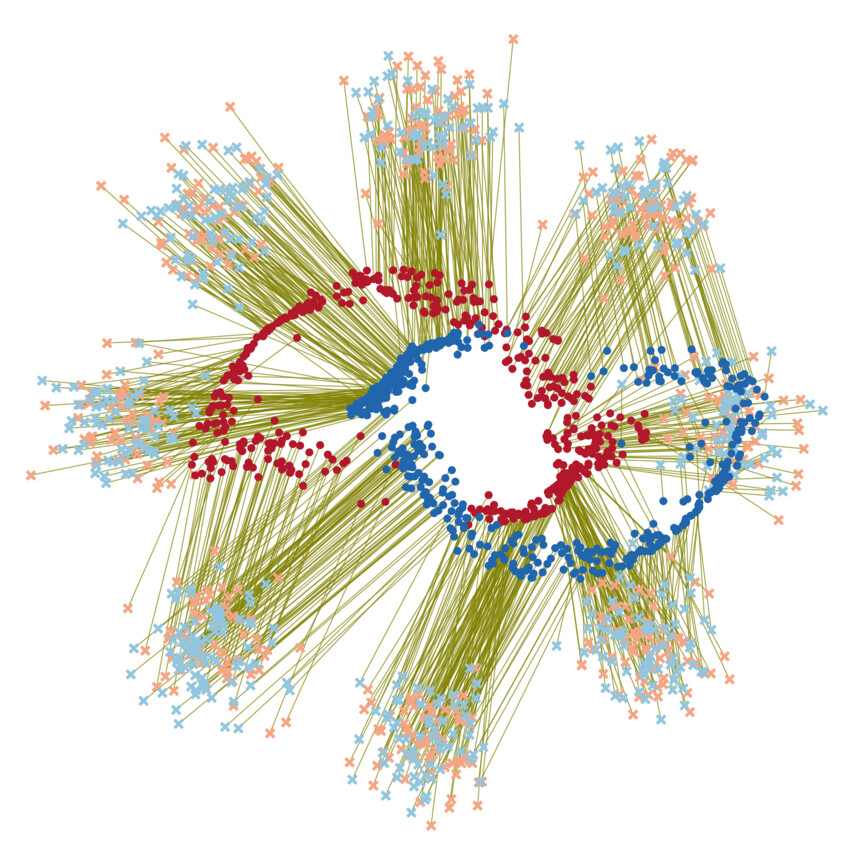} & 
\includegraphics[width=0.118\linewidth]{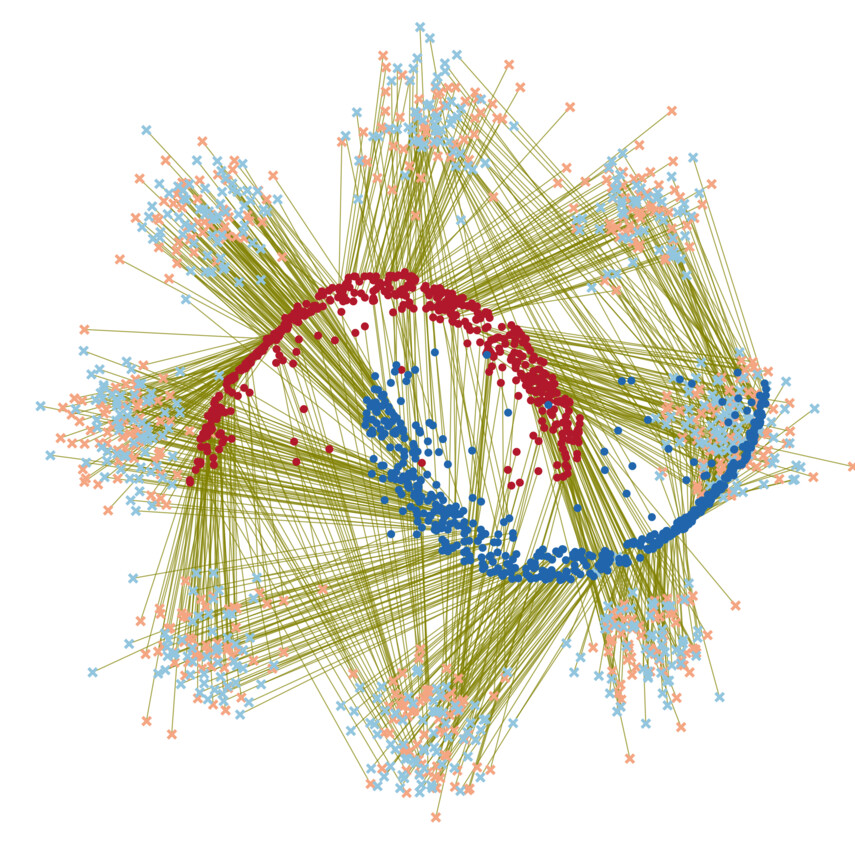} & 
\includegraphics[width=0.118\linewidth]{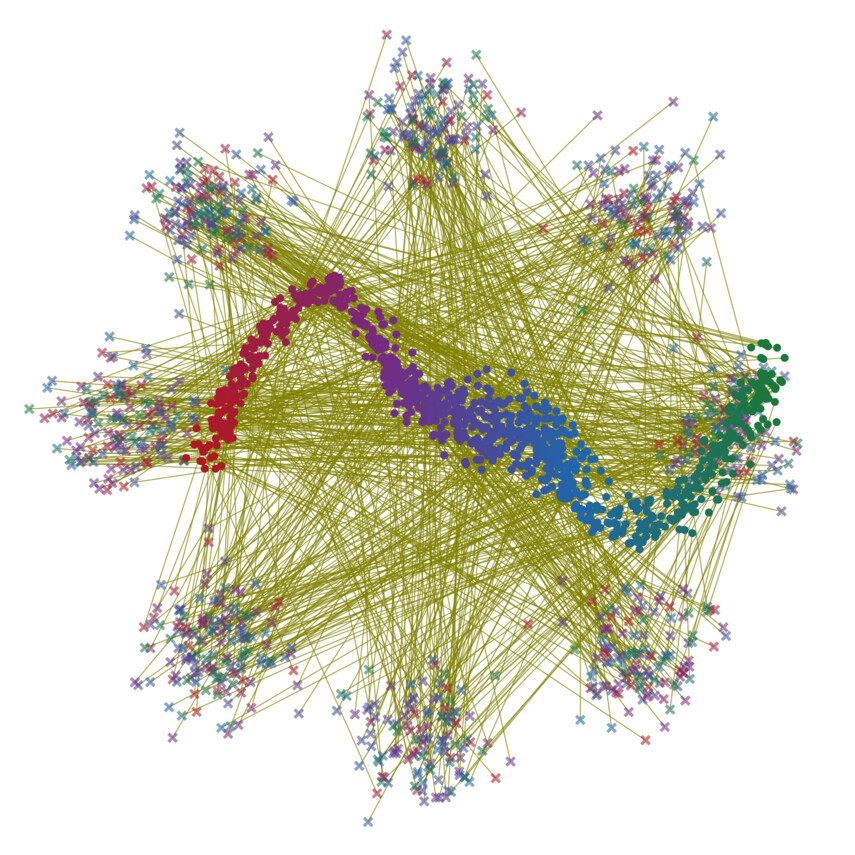} & 
\includegraphics[width=0.118\linewidth]{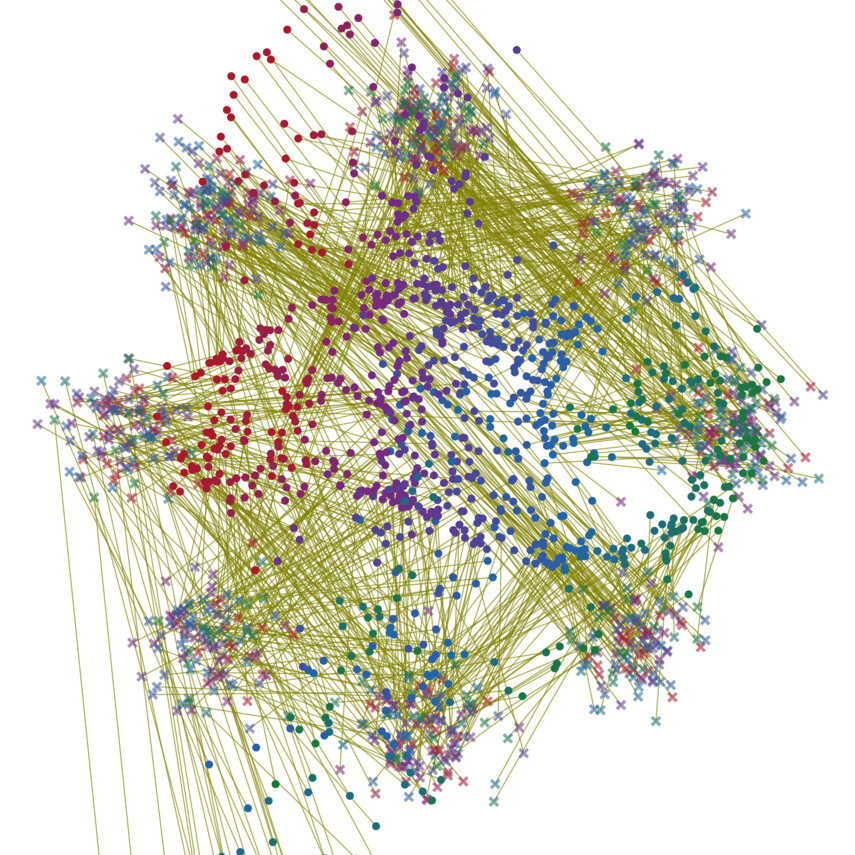} & 
\includegraphics[width=0.118\linewidth]{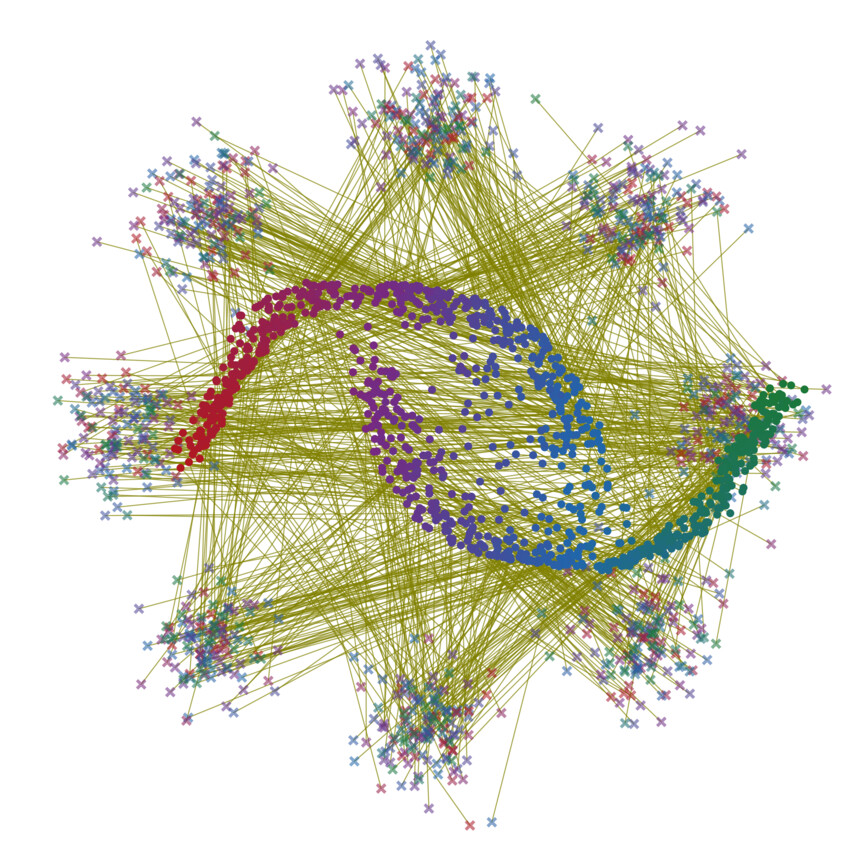}
\\
& \small 6.232\(\pm\)0.186
& \small \textbf{0.072\(\pm\)0.025}
& \small 2.573\(\pm\)0.112
& \small 0.483\(\pm\)0.059
& \small \textbf{0.048\(\pm\)0.010}
& \small 0.732\(\pm\)0.052
& \small 8.276\(\pm\)6.510
& \small \textbf{0.077\(\pm\)0.024}
\\
\Block{2-1}{\rotate \small Adaptive}
&
\includegraphics[width=0.118\linewidth]{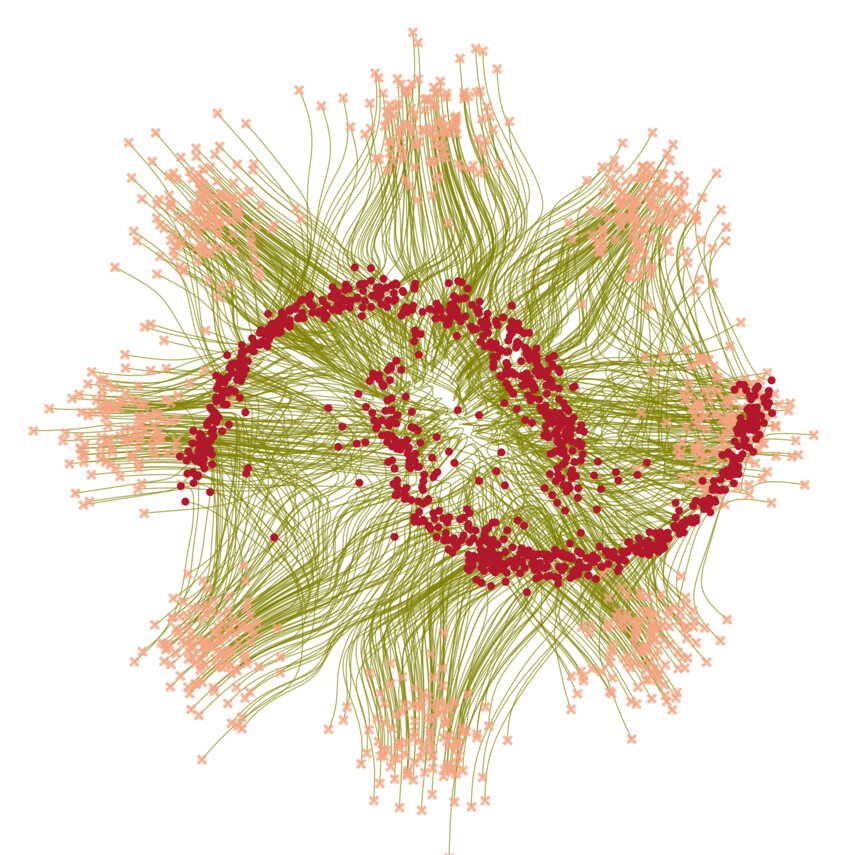} & 
\includegraphics[width=0.118\linewidth]{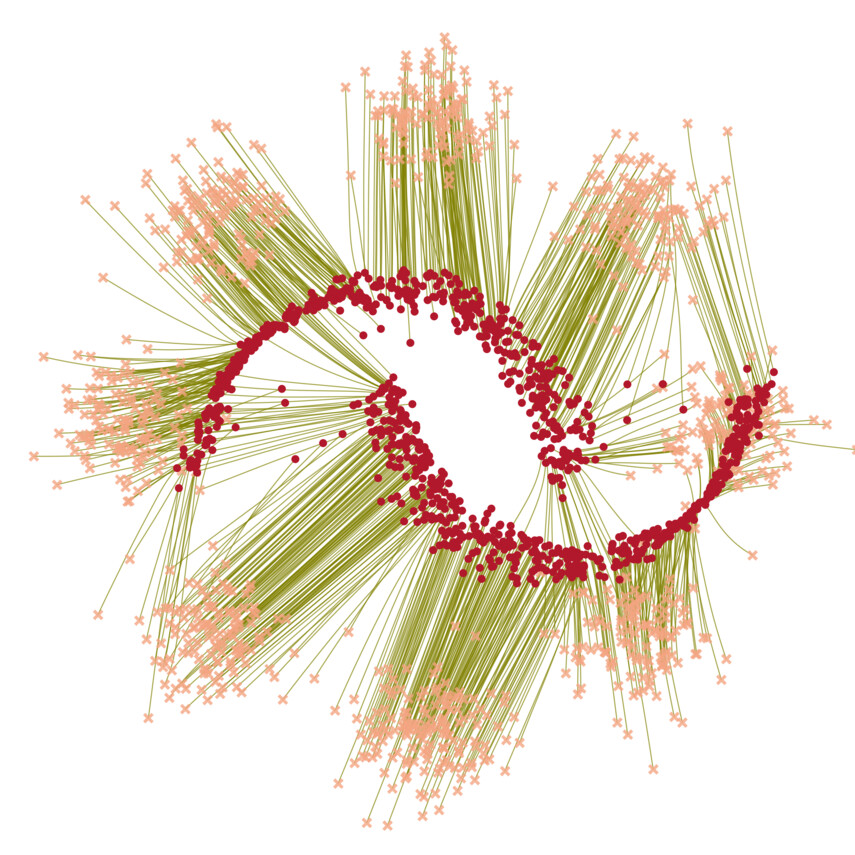} & 
\includegraphics[width=0.118\linewidth]{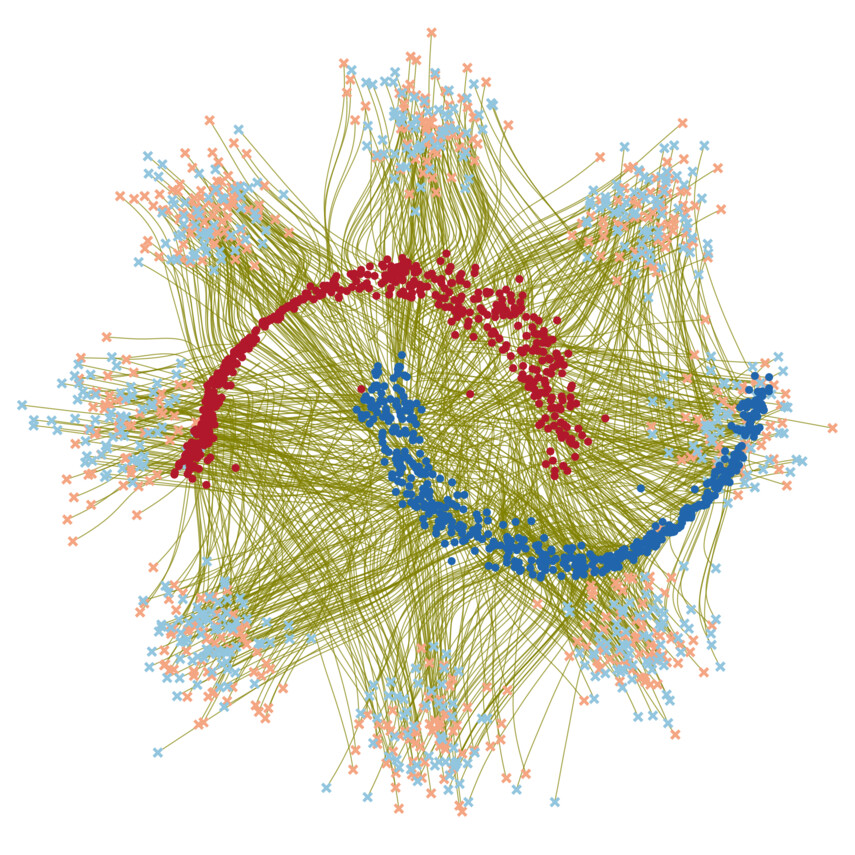} & 
\includegraphics[width=0.118\linewidth]{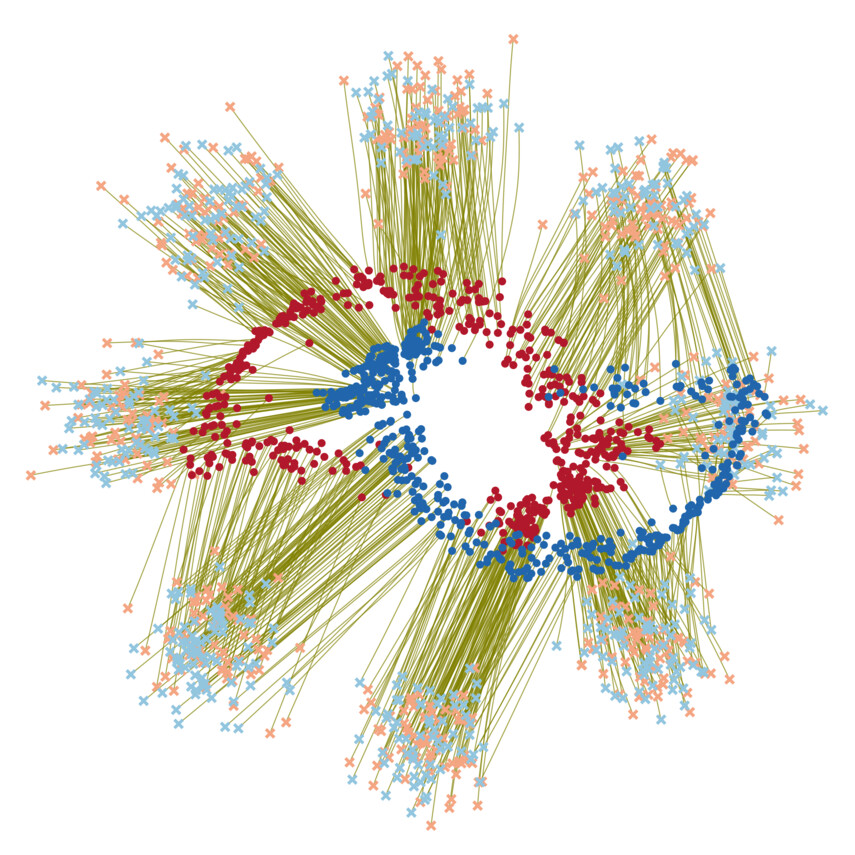} & 
\includegraphics[width=0.118\linewidth]{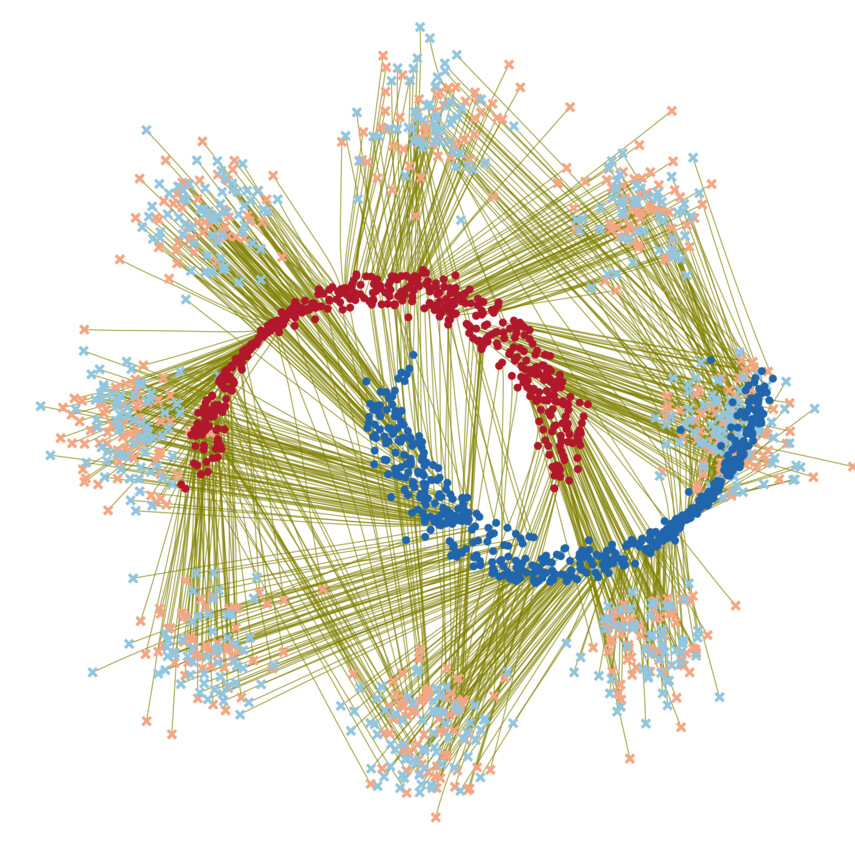} & 
\includegraphics[width=0.118\linewidth]{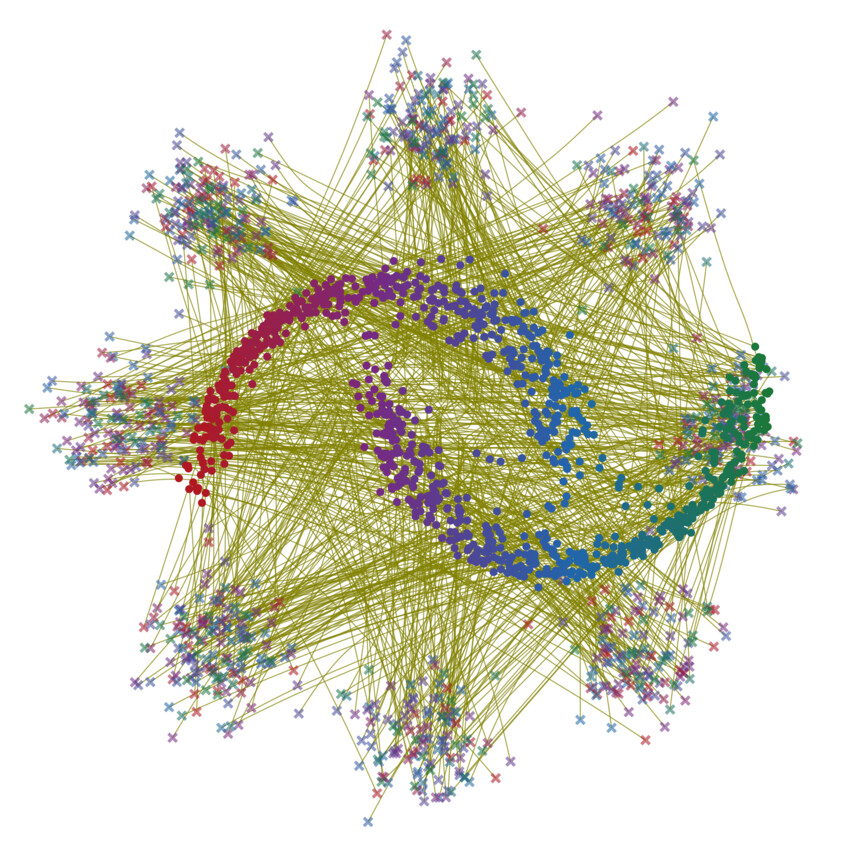} & 
\includegraphics[width=0.118\linewidth]{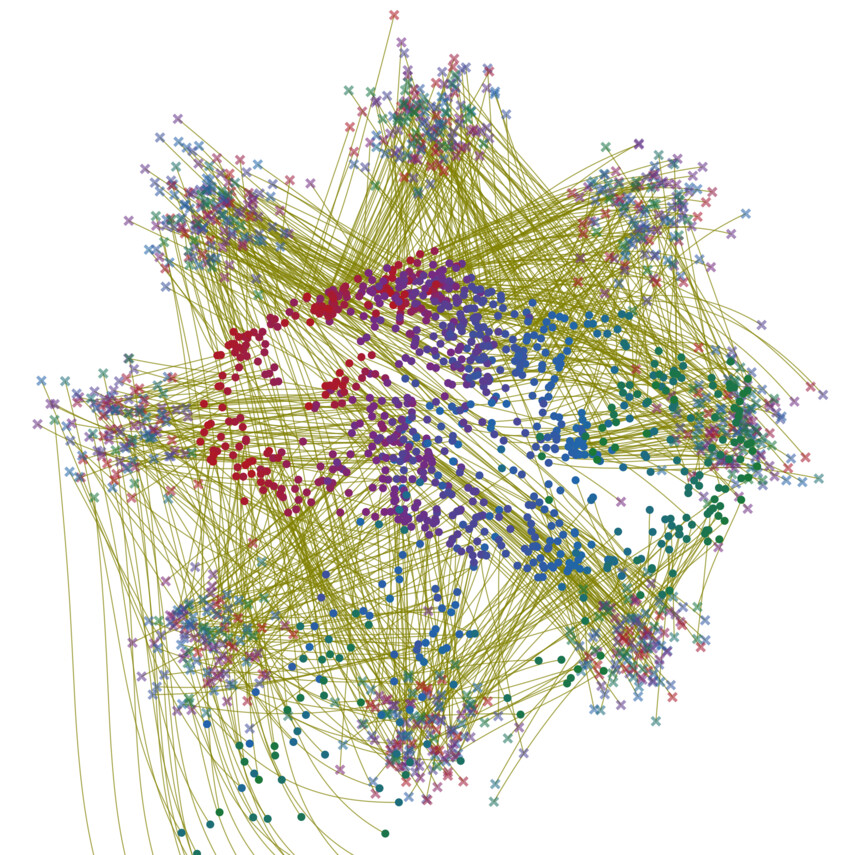} & 
\includegraphics[width=0.118\linewidth]{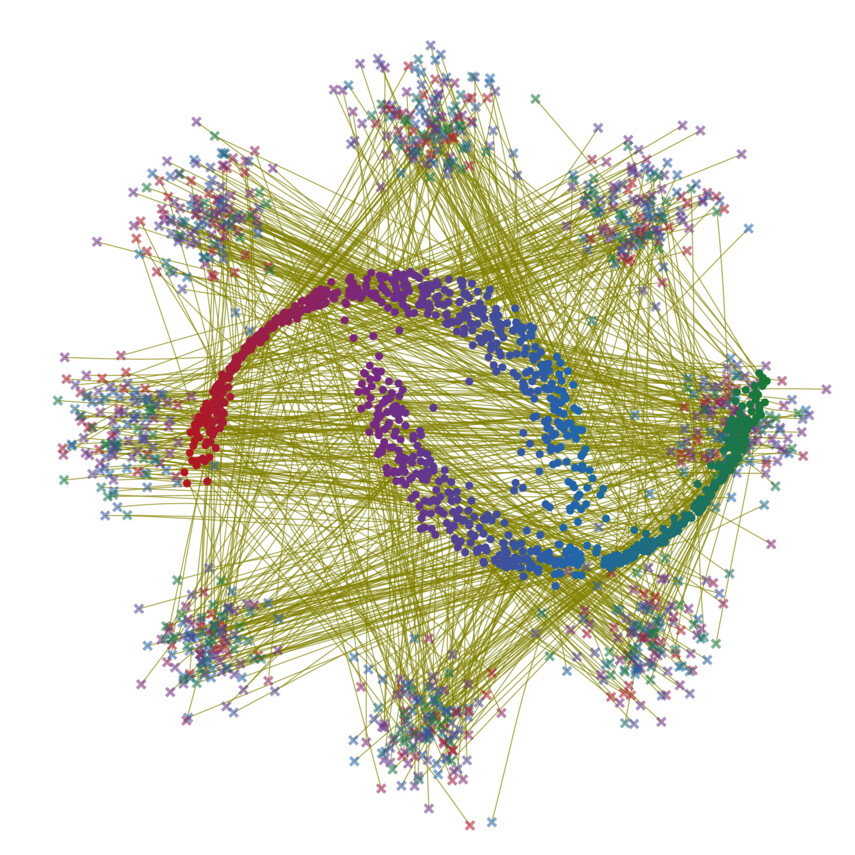}
\\
& \small 0.120\(\pm\)0.045 
& \small \textbf{0.050\(\pm\)0.040} 
& \small 0.059\(\pm\)0.029 
& \small 0.462\(\pm\)0.025 
& \small \textbf{0.018\(\pm\)0.006} 
& \small 0.028\(\pm\)0.010 
& \small 2.143\(\pm\)1.993 
& \small \textbf{0.013\(\pm\)0.003} 
\\
& \small FM
& \small OT
& \small \small FM
& \small OT
& \small C\(^2\)OT (ours)
& \small FM
& \small OT
& \small C\(^2\)OT (ours)
\end{NiceTabular}

\captionof{figure}{
\small 
We visualize the flows learned by different algorithms using the {\tt 8gaussians\(\to\)moons} dataset.
We compare flow matching (FM), minibatch optimal transport FM (OT), and our proposed conditional optimal transport FM (\ourmethod) using one Euler step and an adaptive ODE solver, respectively.
Below each plot, we show the 2-Wasserstein distance (lower is better; mean\(\pm\)std over 10 runs).
For unconditional generation, OT achieves significantly straighter flows,  outperforming FM.
However, paradoxically, OT performs worse when conditions are introduced.
This paper analyzes this degradation and finds that it occurs because optimal transport disregards conditions, leading to a train-test gap. 
To address this, we propose a simple fix (\ourmethod) that outperforms FM and OT in conditional generation. 
}
\label{fig:teaser}
\vspace{2ex}
}{}{}%
\makeatother

\begin{document}
\maketitle

\begin{abstract}
Minibatch optimal transport coupling straightens paths in unconditional flow matching. 
This leads to computationally less demanding inference as fewer integration steps and less complex numerical solvers can be employed when numerically solving an ordinary differential equation at test time. 
However, in the conditional setting, minibatch optimal transport falls short.
This is because the default optimal transport mapping disregards conditions, resulting in a conditionally skewed prior distribution during training.
In contrast, at test time, we have no access to the skewed prior, and instead sample from the full, unbiased prior distribution.
This gap between training and testing leads to a subpar performance.
To bridge this gap, we propose conditional optimal transport (\ourmethod{}) that adds a conditional weighting term in the cost matrix when computing the optimal transport assignment. 
Experiments demonstrate that this simple fix works with both discrete and continuous conditions in 8gaussians\(\to\)moons, CIFAR-10, ImageNet-32\(\times\)32, and ImageNet-256\(\times\)256.
Our method performs better overall compared to the existing baselines across different function evaluation budgets.
Code is available at {\href{https://hkchengrex.github.io/C2OT}{\nolinkurl{hkchengrex.github.io/C2OT}.}}

\end{abstract}

\section{Introduction}
\label{sec:intro}

We focus on flow-matching-based conditional generative models, \ie, generation guided by an input condition.\footnote{
We use condition to denote an \emph{input condition}, not to be confused with the `condition' in conditional flow matching (CFM)~\cite{lipman2022flow} (where the second `C' in our acronym comes from), which refers to a data sample.
To avoid ambiguity, we refer to CFM simply as flow matching (FM) in this paper.
}
Examples of such conditions include class labels, text, or video.
Recently, flow matching has been applied in many areas of computer vision, including text-to-image/video~\cite{esser2024scaling,polyak2025moviegencastmedia}, vision-language applications~\cite{black2024pi_0}, image restoration~\cite{martin2024pnp}, and video-to-audio~\cite{cheng2024taming}.
However, these methods can be slow at test-time -- obtaining a solution requires numerically integrating the flow with an ODE solver, typically involving many steps, each requiring a deep network forward pass.
Na\"ively reducing the number of steps degrades performance, 
as the underlying flow is often curved (\Cref{fig:teaser}, leftmost column), necessitating a small step size for accurate numerical integration.
One way to address this issue is by straightening the flow.
In the context of unconditional generation, \citet{tong2023improving} and \citet{pooladian2023multisample} concurrently proposed ``minibatch optimal transport'' (OT), which deterministically couples data and samples from the prior within a minibatch via optimal transport (replacing random coupling) to minimize flow path lengths and to straighten the flow.

While OT improves unconditional generation, we find that it paradoxically and consistently harms conditional generation (\Cref{fig:teaser}).
This occurs because OT disregards the conditions when computing the coupling.
As a result, during training, OT samples from a prior distribution that is \emph{skewed} by the condition, as we show in \Cref{sec:ot-ignores-conditions}.
However, at test time, we have no access to this skewed distribution and instead sample from the full prior distribution.
This mismatch creates a gap between training and testing, leading to performance degradation.
To bridge this gap, we propose conditional optimal transport FM (\ourmethod{}) which introduces a simple yet effective condition-aware weighting term when computing the cost matrix for OT.
Additionally, we propose two techniques: adaptive weight finding to simplify hyperparameter tuning, and efficient oversampling of OT batches to counteract the reduced effective OT batch size introduced by the weighting term.

We conduct extensive experiments on both two-dimensional synthetic data and high-dimensional image data, including CIFAR-10, ImageNet-32\(\times\)32, and ImageNet-256\(\times\)256.
The results demonstrate that our method performs well across different condition types (discrete and continuous), datasets, network architectures (UNets and transformers), and data spaces (image space and latent space).
\ourmethod{} achieves better overall performance than the existing baselines across different inference computation budgets, including with few-steps Euler's method and with an adaptive ordinary differential equation (ODE) solver~\cite{dormand1980family}.

\section{Related Works}
\label{sec:related}

\paragraph{Flow Matching.}
Recently, flow matching~\cite{lipman2022flow,albergo2023building,albergo2023stochastic} has become a popular choice for generative modeling, in part due to its simple training objective and its ability to generate high-quality samples.
However, at test time, flow-based methods typically require many computationally expensive forward passes through a deep net to numerically integrate the flow.
This is because the underlying flow is usually curved and therefore cannot be approximated well with a few integration steps.

\paragraph{Optimal Transport Coupling.}
To straighten the flow, in the context of unconditional generation, \citet{tong2023improving} and \citet{pooladian2023multisample} concurrently proposed minibatch optimal transport (OT).
While OT has proven effective in the unconditional setting, we show in \Cref{sec:ot-ignores-conditions} that it skews the prior distribution and fails in the conditional setting if used na\"ively.
Our method, \ourmethod{}, specifically addresses this failure, aiming to extend the success of OT in unconditional generation to conditional generation.
Note, OT has also been used in equivariant flow matching~\cite{song2023equivariant,klein2023equivariant}, which exploits domain-specific data symmetry (\eg, in molecules). 
In contrast, our method targets general data.
Further, \citet{hui2025not} find that mixing of OT and independent coupling benefits shape completion,  which is an orthogonal contribution to this paper.
Concurrently, \citet{davtyan2025faster} improve data-noise coupling for unconditional generation, while leaving data coupling for conditional generation to future work.

\paragraph{Straightening Flow Paths.}
Other methods of straightening flow paths exist.
Reflow~\cite{liu2022flow} achieves this by retraining a flow matching network multiple times, at the cost of additional training overhead.
Variational flow matching~\cite{guo2025variational} reduces flow ambiguity with a latent variable and hierarchical flow matching~\cite{zhang2025towards} straightens flows by modeling higher-order flows.
These approaches are orthogonal to our focus on prior-data coupling and, in principle, can be combined with our method.

\paragraph{Learning the Prior Distribution.}
An alternative approach to improving flow is to jointly learn a prior distribution with the flow matching network.
However, these methods~\cite{lee2023minimizing,liu2024flowing,silvestri2025training} typically rely on a variational autoencoder (VAE)~\cite{kingma2013auto,higgins2017beta} to learn the prior.
This introduces yet another density estimation problem, which complicates training. 
In this work, we focus on improving training-free coupling plans without learning a new prior.

\paragraph{Consistency Models.}
Consistency models~\cite{song2023consistency} have been proposed to accelerate diffusion model sampling, and can be extended to flow matching~\cite{yang2024consistency}.
Recent advancements have improved training efficiency~\cite{geng2024consistency}, stability~\cite{lu2024simplifying}, and quality~\cite{kim2023consistency}.
These methods are orthogonal to our contribution, which focuses on improving prior-data coupling, and can be integrated, as demonstrated by \citet{silvestri2025training} in their combination of consistency models with OT.
In this paper, we focus on the base form of flow matching to avoid distractions from other formulations.

\section{Method}
\label{sec:method}

\begin{figure*}
    \includegraphics[width=\linewidth]{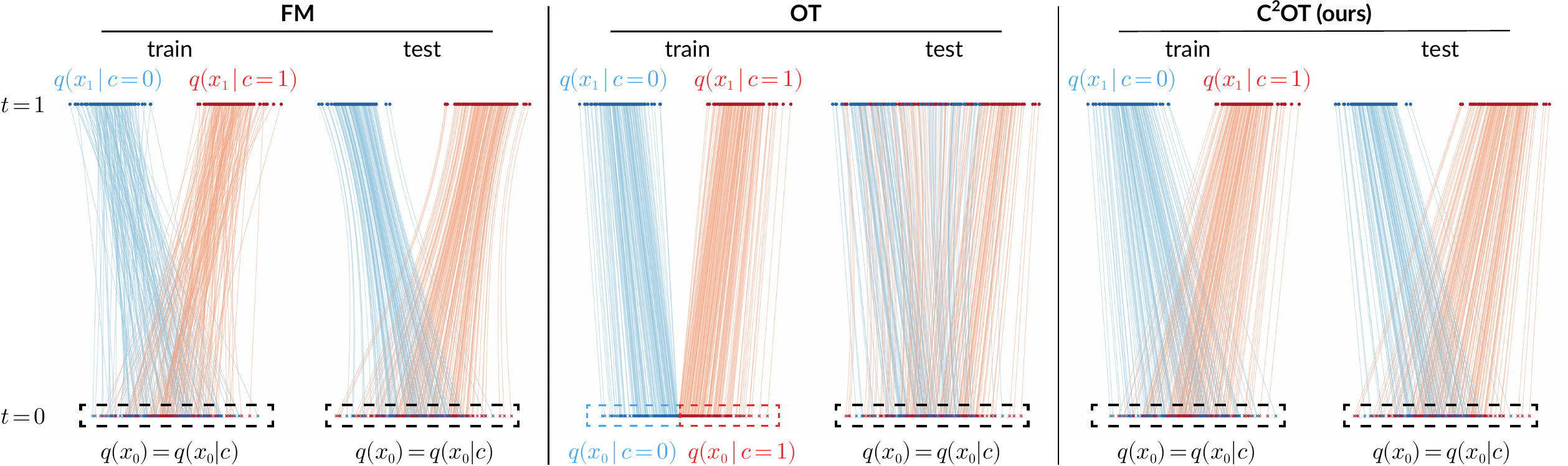}
    \caption{
    We visualize the coupling during training and the learned flow from \(t=0\) to \(t=1\) during testing for FM, OT, and our proposed method \ourmethod.
    The prior is defined as \(q(x_0)=\mathcal{N}(0, 1)\), while the target distribution is given by \(q(x_1)=\tfrac{1}{2}q(x_1|c=0) + \tfrac{1}{2}q(x_1|c=1)) = \tfrac{1}{2}\mathcal{N}(-2, 0.5) + \tfrac{1}{2}\mathcal{N}(2, 0.5)\).
    Note, FM results in curved flows during testing, while OT degenerates due to skewed training samples (\(q(x_0|c)\neq q(x_0)\)).
    In contrast, our method successfully learns straight flows without degeneration.
    }
    \label{fig:1d_example}
\end{figure*}

\subsection{Preliminaries}

\paragraph{Flow Matching (FM).}
We base our discussion on FM~\cite{lipman2022flow,tong2023improving} (also called rectified flow~\cite{liu2022flow}) for generative modeling and refer readers to \cite{tong2023improving,liu2022flow} for details. 
At test-time, a sample \(\generatedsample\) is characterized by an ordinary differential equation (ODE) with an initial boundary value specified via noise \(x_0\), randomly drawn from a prior distribution (\eg, the standard normal). To compute the sample \(\generatedsample\), this ODE is commonly solved by numerically integrating from time $t=0$ to time $t=1$ a learned flow/velocity field \(\flow\), where \(\theta\) denotes the set of trainable deep net parameters.
For conditional generation, we additionally obtain a condition \(\condition\), \eg, a class label from user input, which adjusts the flow. We use \(\conditionedflow\) to denote this conditional flow.
At training time, we find the deep net parameters \(\theta\) by minimizing the  objective
\begin{equation}
    \mathbb{E}_{t, (x_0,x_1,\condition)\sim\jointdistribution(x_0, x_1, \condition)} \lVert \conditionedflow(t, x_t) - u(x_t|x_0, x_1) \rVert^2, 
\label{eq:cfm_objective}
\end{equation}
where \(t\) is uniformly drawn from the interval \([0, 1]\), and where \(\jointdistribution(x_0, x_1, \condition)\) denotes the joint distribution of the prior and the training data.
In practice, often, \emph{independent coupling} is used, \ie, the prior and the training data are sampled independently, such that
\begin{equation}
    \jointdistribution(x_0, x_1, \condition) = q(x_0)q(x_1, c).
\label{eq:independent_coupling}
\end{equation}
Further,  
\begin{equation}
    x_t = tx_1 + (1-t)x_0
\end{equation}
defines a linear interpolation between noise and data, and
\begin{equation}
    u(x_t|x_0, x_1) = x_1 - x_0
\end{equation}
denotes its corresponding flow velocity at \(x_t\). Note, simply dropping the condition \(c\) leads to the unconditional flow matching formulation.

Although rectified flow matching is trained with straight paths, the learned flow  is often curved (\Cref{fig:teaser}, bottom-left) since there are many possible paths induced by the independently sampled couplings \(x_0, x_1\) through any \(x_t\) at time \(t\)~\cite{pooladian2023multisample}.
Curved paths require more numerical integration and network evaluation steps, which is not ideal.

\paragraph{Optimal Transport (OT) Coupling.}
To address this issue for \textbf{unconditional generation}, \citet{tong2023improving} and \citet{pooladian2023multisample} concurrently proposed to use 
minibatch optimal transport to deterministically couple the sampled prior and a sampled batch of the data. This minimizes flow path lengths and straightens the flow.
Concretely, given a minibatch of \(\batchsize\) \(\datadim\)-dimensional samples, an OT coupling seeks a \(\batchsize \times \batchsize\) \emph{permutation matrix} \(P\) such that the squared Euclidean distance is minimized, \ie, 
\begin{equation}
    \min_P \|X_0 - PX_1\|_2^2.
\label{eq:ot_optim}
\end{equation}
Here, \(X_0, X_1\in\mathbb{R}^{\batchsize\times\datadim}\) are matrices containing a minibatch of samples from the prior and the data.
Instead of an ``independent coupling'', in the unconditional setting we then optimize a variant of \Cref{eq:cfm_objective} using the coupled tuple \((X_0, PX_1)\). 
Note that this algorithm corresponds to sampling tuples \((x_0,x_1)\) from a joint distribution \(\otjointdistribution(x_0, x_1)\).

Importantly, for \textbf{unconditional generation}, \ie, when using the joint distribution \(\otjointdistribution(x_0, x_1)\), the marginals remain unchanged from independent coupling (see~\cite{tong2023improving}):
\begin{align}
    \int \otjointdistribution(x_0, x_1)q(x_0)\diff x_0 &= q(x_1), \text{ and } \nonumber\\
    \int \otjointdistribution(x_0, x_1)q(x_1)\diff x_1 &= q(x_0).
\end{align}
Hence, during test-time, we can sample from the prior \(q(x_0)\) and integrate the flow \(v_\theta\) to simulate data from the data distribution \(q(x_1)\).

Unfortunately, as shown in the next section, this does not hold for \textbf{conditional generation}. Said differently, na\"ively extending the unconditional joint distribution \(\otjointdistribution(x_0, x_1)\) to a conditional joint distribution  \(\otjointdistribution(x_0, x_1, c)\) by using the coupled tuple \((X_0, PX_1, PC)\), where \(C\in\mathbb{R}^{\batchsize\times\conditiondim}\) is a batch of \(\conditiondim\)-dimensional conditions, does not maintain marginals and leads to a gap between training time and test time. We discuss this next.

\subsection{OT Coupling Skews The Prior}
\label{sec:ot-ignores-conditions}

\paragraph{Intuition.}
Consider the one-dimensional example illustrated in \Cref{fig:1d_example}: the prior distribution \(q(x_0)\) follows a Gaussian distribution, and the target distribution \(q(x_1)\) consists of a mixture of two  Gaussians translated left and right, with the translation direction indicated by a binary condition~\(c\).
When training with a \(\otjointdistribution(x_0, x_1,c)\) that represents the coupled tuple \((X_0, PX_1, PC)\), samples from the prior and the sampled conditions become correlated through the OT coupling -- samples from the left half of the prior distribution are always associated with \(c=0\), while those from the right half always correspond to \(c=1\).
As a result, the model overfits on data pairs \((x_0, c)\) where either \((x_0<0, c=0)\) or \((x_0>0, c=1)\).
However, during testing, this skewed distribution is no longer accessible, and instead, \(x_0\) and \(c\) are sampled independently.
Consequently, the model encounters previously unseen data pairs, \ie, \((x_0>0, c=0)\) and \((x_0<0, c=1)\) -- leading to failure.
A similar phenomenon can be observed in the two-dimensional example illustrated in \Cref{fig:teaser}.

\paragraph{Skewed Prior.}
Formally, imagine that we consider the most general OT coupling joint distribution \(\otjointdistribution(x_0, x_1, c)\) at training time. 
At test time, the user wants to specify a condition \(c_1\) to obtain samples from the conditional distribution \(q(x_1|c=c_1)\). 
What is the prior distribution that we need to sample from such that we arrive at the correct conditional distribution \(q(x_1|c=c_1)\)? 
To compute this, imagine, we are given an arbitrary condition \(c_1\), \ie, we look at the induced distribution 
\(\otjointdistribution(x_0, x_1, c | c=c_1)\).
Marginalizing this induced distribution over \(x_1\) yields
\begin{align}
    \int \otjointdistribution(x_0, x_1, c| c=c_1)\diff x_1 
    &= \int \otjointdistribution(x_0, x_1| c=c_1)\diff x_1 \nonumber\\
    &= q(x_0 | c=c_1), 
\end{align}
which implies that the prior required to arrive at the conditional distribution \(q(x_1 | c=c_1)\) is \(q(x_0 | c=c_1)\) -- a distribution that we cannot access at test time. 
In general, the prior that we need to sample from, given condition \(c\), is skewed from \(q(x_0)\) to \(q(x_0 | c)\) which is inaccessible at test time.
As a result, accurately capturing \(q(x_1 | c)\) at test time becomes infeasible unless \(q(x_0)\) and \(q(c)\) are independent, \ie, if \(q(x_0| c)=q(x_0)\).
However, this condition is generally not satisfied by the most general OT coupling -- for instance, in \Cref{fig:1d_example}, clearly \(q(x_0|c)\neq q(x_0)\).
To address this issue, we propose a method \ourmethod{} to unskew the prior. We discuss this in the next section.

\begin{figure*}
\vspace{-1em}
\centering
\input{fig/algo}
\end{figure*}

\subsection{Conditional Optimal Transport FM}
\label{sec:c2ot}

\subsubsection{Formulation}

We propose conditional optimal transport FM (\ourmethod{}) to \emph{unskew} the prior \(q(x_0|c)\) such that \(q(x_0|c)=q(x_0)\).
This ensures that at test time, we can sample from the full prior \(q(x_0)\), irrespective of the condition \(c\). %
Importantly, at the same time, we also aim to preserve the straight flow paths provided by OT.
Conceptually, we construct a prior distribution independently for each condition, \ie, \(q(x_0|c_1)=q(x_0|c_2)=q(x_0), \forall c_1, c_2\).
This can be achieved by sampling from the prior and computing OT independently for each condition, as shown in \Cref{fig:1d_example} (right).

Formally, we construct the joint distribution as %
\begin{equation}
    \ourjointdistribution(x_0, x_1, c) \coloneq q(x_0)q(c)\otjointdistribution(x_1|x_0, c).
\label{eq:our_joint_distribution}
\end{equation}
Here, the prior \(q(x_0)\) and the condition \(q(c)\) are sampled independently, while the data \(x_1\) is conditioned on \(x_0\) (via OT) and \(c\) (via the dataset construction). 
Different from independent coupling in \Cref{eq:independent_coupling}, \(x_1\) and \(x_0\) are associated through OT and therefore lead to straighter flow paths.
Also different from the most general OT coupling \(\otjointdistribution(x_0, x_1, c)\), we explicitly enforce the independence between \(x_0\) and \(c\).

To see that this joint distribution provides an unskewed prior, imagine we are given an arbitrary input condition \(c_1\), marginalizing over \(x_1\) yields
\begin{align}
    \int &q(x_0)q(c|c=c_1)\otjointdistribution(x_1|x_0, c=c_1) \diff x_1 \nonumber\\
    =\int &q(x_0)\otjointdistribution(x_1|x_0, c=c_1) \diff x_1 = q(x_0).
\end{align}
This implies that, at test time, we can sample from the entire prior \(q(x_0)\), regardless of the condition \(c\), to arrive at the desired data distribution \(q(x_1|c)\).
During training, in practice, we sample from \(\ourjointdistribution\) by modifying the optimal transport cost function in \Cref{eq:ot_optim}.
This process is exact for discrete conditions (\Cref{sub:method_discrete_cond}) and approximate for continuous conditions (\Cref{sub:method_continuous_cond}).
Specifically, given a minibatch of \(\batchsize\) \(\datadim\)-dimensional samples and a minibatch of \(\batchsize\) \(\conditiondim\)-dimensional conditions \(C\in\mathbb{R}^{\batchsize\times\conditiondim}\), \ourmethod{} seeks a \(\batchsize \times \batchsize\) permutation matrix \(P\) that minimizes the following cost function:
\begin{equation}
\min_P \|X_0 - PX_1\|_2^2 \color{MidnightBlue} + \sum_{i=1}^\batchsize f(C_i, [PC]_i) \color{black}, 
\label{eq:mod_ot_optim}
\end{equation}
where \(f\) is a symmetric and non-negative cost function, satisfying the property \(f(c, c)=0\) \(\forall c\).
The key differences between independent coupling, OT coupling, and \ourmethod{} coupling are highlighted in \Cref{algo:cfm,algo:ot,algo:ours}.
In the following two sections, we discuss our choice of \(f\) for both discrete and continuous conditions.

\subsubsection{Discrete Conditions}
\label{sub:method_discrete_cond}

For discrete conditions, we assume the conditions correspond to class labels from a finite, discrete set, \ie, \(c\in\{0, 1, \dots, \numlabels-1\} \).
For example, in \Cref{fig:1d_example}, the labels \(0\) and \(1\) represent the left and right groups in the target distribution, respectively.
We define \(f=\discretef\) such that no transport is allowed to modify the conditions (\ie, \(PC = C\)). This is achieved via 
\begin{equation}
    \discretef (c_1, c_2) = 
    \begin{cases}
    \infty, c_1\neq c_2, \\
    0, \text{otherwise}.
    \end{cases}
\label{eq:discrete_f}
\end{equation}
As a result, the unordered set of prior samples for any given condition \(c\) remains unchanged: \( \{{X_0}_i | c_i=c\} \overset{\mathrm{iid}}{\sim} q(x_0) \).
This formulation samples exactly from our ideal (\Cref{eq:our_joint_distribution}), as OT is computed independently within each class, and the data sample from each class is exposed to the full prior without skew.
This effect is illustrated in \Cref{fig:teaser} (middle) and \Cref{fig:1d_example} (right). 
As expected, the results differ from FM's curved flow paths and OT's inability to properly capture the target distribution.

It is important to note that enforcing \(PC = C\) does not necessarily imply \(P=I\) since we can still swap row \(i\) and \(j\) in \(P\) as long as \(c_i=c_j\).
Setting \(P=I\) would correspond to a degeneration back to independent coupling in \Cref{eq:independent_coupling} and forfeit the benefits of straightened flow provided by OT.

Unfortunately, using \(\discretef\) with continuous conditions leads to exactly this degeneration, since, in general, no \(c_i=c_j\) unless \(i=j\).
Hence, in the next section, we devise a relaxed cost for handling continuous conditions.

\subsubsection{Continuous Conditions}
\label{sub:method_continuous_cond}

For continuous conditions, the condition is typically a feature embedding, \eg, computed from a text prompt.
Directly applying \(\discretef\) in \Cref{eq:discrete_f} leads to a degeneration back to independent coupling as discussed earlier. To avoid this, we propose a relaxed penalty function \(\continuousf\) based on the cosine distance:
\begin{equation}
    \continuousf(c_1, c_2) = w\left(1 - \frac{c_1\cdot c_2}{\|c_1\| \|c_2\|}\right) = w\cdot\text{cdist}(c_1, c_2).
\end{equation}
Here, \(\conditionweight>0\) is a scaling hyperparameter.
Note, when \(w=0\), this formulation degenerates to regular OT; when \(w\to\infty\), this formulation approaches \Cref{eq:discrete_f}, \ie, independent coupling if no two conditions are equal.
The right section of \Cref{fig:teaser} illustrates an example where the \(x\)-coordinate of the target data point serves as the condition.
Notably, \ourmethod{} preserves straight flow paths without exhibiting OT's apparent degradation.

\paragraph{Finding \(\conditionweight\).}
The optimal \(\conditionweight\) is highly dependent on the data distribution and the magnitude of features. 
To alleviate the need for tuning the hyperparameter \(\conditionweight\), we propose to find \(\conditionweight\) adaptively for each minibatch.
For this, we develop a ratio \(\conditionratio(\conditionweight)\) which measures the proportion of samples that are considered ``potential optimal transport candidates'', \ie, 
\begin{multline}
    \conditionratio(\conditionweight) = \frac{1}{B^2} \sum_{i, j} \mathbb{1}
    \left(
    \| {X_0}_i - {X_1}_j \|_2^2 
    + w\cdot\cosinedist(c_i, c_j)
    \right. \\ \left.
    \leq
    \| {X_0}_i - {X_1}_i \|_2^2
    \right), 
\label{eq:ratio}
\end{multline}
where \(\mathbb{1}(\cdot)\) is an indicator function that evaluates to one if the condition is satisfied and to zero otherwise.
Then, we introduce a hyperparameter \(\targetratio\) as the target ratio and find \(\conditionweight\) such that \(\conditionratio(\conditionweight) \approx \targetratio\).
Note that setting \(\targetratio\) is invariant to the scaling of distances between \(x_0\) and \(x_1\).
Namely, if \(\| {X_0}_i - {X_1}_j \|_2^2 \) is scaled by \(s\), \(\conditionweight\) would also be scaled by \(s\) to preserve the same \(\targetratio\).
This invariance is particularly useful when transitioning to a new dataset, such as changing image resolutions in image generation tasks.
In our implementation, we search for \(\conditionweight\) in two steps: first with exponential search to establish the upper/lower bounds, then with binary search to locate a precise value. 
The final \(\conditionweight\) is used as the initial value in the subsequent minibatch. 
In practice, we observe \(\conditionweight\) differs very little between minibatches (<1\%) and the search converges quickly within 10 iterations with a negligible overhead.
Next, we introduce \emph{oversampling}, a simple technique to obtain more accurate OT couplings.

\subsubsection{Oversampling OT Batches}
\label{sec:oversampling}

\paragraph{OT \vs Deep Net Batch Sizes.}
Typically, the ``OT batch size'' \(\batchsize\) used to compute optimal transport is set equal to the ``deep net batch size'' \(\netbatchsize\) used to compute forward/backward passes of the deep net~\cite{tong2023improving,pooladian2023multisample}.
While the OT batch size controls the dynamics of prior-to-data assignments, the deep net batch size affects gradient variances, training efficiency, and memory usage.
There is no reason why setting them equal should be optimal.
Particularly, with our proposed \ourmethod, the effective OT batch size is reduced since we restrict transport between samples with divergent conditions, which calls for an \emph{increased OT batch size \(\batchsize\)}. At the same time, keeping the deep net batch size \(\netbatchsize\) unchanged prevents unwarranted side effects in other aspects of training.

\paragraph{Reduced Effective OT Batch Size.}
In the case of discrete conditions, OT is effectively performed within \( \{{X_0}_i, {X_1}_i | c_i=c\}\) for each condition \(c\).
With \(\numlabels\) classes, the expected effective OT batch size is reduced to \( \mathbb{E}(\lvert \{i | c_i=c\} \rvert) = \batchsize / \numlabels \).
Empirically, we find that using a much larger OT batch size \(\batchsize\) than the network batch size \(\netbatchsize\) is helpful.
This observation motivates oversampling.

\paragraph{Efficiency.}
A natural concern with increasing \(\batchsize\) is the computational overhead, 
since optimal transport with Hungarian matching~\cite{kuhn1955hungarian} has a time complexity of \(O(\batchsize^3)\).
However, 
\begin{itemize}
    \item Since each OT batch can be used across \((\batchsize / \netbatchsize)\) forward/backward passes, the amortized cost per minibatch reduces to \(O(\batchsize^2 \netbatchsize)\).
    \item We offload OT computation from the main process to the data workers, enabling data preparation on the CPU in parallel with forward/backward passes on the GPU. 
\end{itemize}
With a modest choice of 8 data workers per GPU, we observe \textbf{no wall-clock overhead} for OT batch sizes up to 6,400 when training on CIFAR-10 and ImageNet, as data preparation is completed faster than the GPU can process batches.

\section{Experiments}
\label{sec:expr}

\begin{table}
\small
    \centering
    \caption{Results on the {\tt 8gaussians\(\to\)moons} dataset with different training coupling methods and different conditioning.
    Note that our method \ourmethod{} does not apply in the unconditional setting.
    }
    \begin{NiceTabular}{l@{\hspace{2mm}}c@{\hspace{2mm}}c@{\hspace{2mm}}c}
\toprule
Method & Euler-1 (\(W_2^2\)↓) & Adaptive (\(W_2^2\)↓) & NFE ↓ \\
\bottomrule
\multicolumn{4}{l}{\emph{Unconditional}} \\
\midrule
FM & 6.232\(\pm\)0.186 & 0.120\(\pm\)0.045 &  93.22\(\pm\)1.81 \\
OT & \textbf{0.072}\(\pm\)0.025 & \textbf{0.050}\(\pm\)0.040 & \textbf{36.08}\(\pm\)2.57 \\
\bottomrule
\multicolumn{4}{l}{\emph{Binary conditions}} \\
\midrule
FM & 2.573\(\pm\)0.112 & 0.059\(\pm\)0.029 &  91.10\(\pm\)2.77 \\
OT & 0.483\(\pm\)0.059 & 0.462\(\pm\)0.025 &  \textbf{32.89}\(\pm\)0.95 \\
\rowcolor{defaultColor}
\ourmethod{} (ours) & \textbf{0.048}\(\pm\)0.010 & \textbf{0.018}\(\pm\)0.006 &  58.88\(\pm\)1.48 \\
\bottomrule
\multicolumn{4}{l}{\emph{Continuous conditions}} \\
\midrule
FM & 0.732\(\pm\)0.052 & 0.028\(\pm\)0.010 &  93.86\(\pm\)1.76 \\
OT & 8.276\(\pm\)6.510 & 2.143\(\pm\)1.993 &  \textbf{42.07}\(\pm\)2.59 \\
\rowcolor{defaultColor}
\ourmethod{} (ours) & \textbf{0.077}\(\pm\)0.024 & \textbf{0.013}\(\pm\)0.003 & 89.87\(\pm\)1.53 \\
\midrule
\bottomrule
\end{NiceTabular}

    \label{tab:two_moons}
\end{table}

We experimentally verify our proposed method \ourmethod{} in two sections: first on the two-dimensional synthetic dataset {\tt 8gaussians\(\to\)moons}, and then on high-dimensional image data, including CIFAR-10~\cite{krizhevsky2009learning} and ImageNet-1K~\cite{deng2009imagenet}. 
For ImageNet-1K, we conduct experiments on both 32\(\times\)32 images in image space and 256\(\times\)256 images in latent space. 
Implementation details are provided in \Cref{sec:app:implementation}.

\subsection{Two-Dimensional Data}

The two-dimensional {\tt 8gaussians\(\to\)moons} dataset maps a mixture of eight Gaussian distributions to two interleaving half-circles, as visualized in \Cref{fig:teaser}.
We consider three conditioning scenarios: (a) unconditional; (b) binary class conditions, where each class corresponds to one of the half-circles; and (c) continuous conditions, where the \(x\)-coordinate of the desired target point is given as the condition.
For (b), we apply our setup for discrete conditions (\Cref{sub:method_discrete_cond}); for (c), we apply our setup for continuous conditions (\Cref{sub:method_continuous_cond}) with \(\targetratio=0.01\).

\paragraph{Metrics.}
We report the 2-Wasserstein distance (\(W^2_2\)) (following~\cite{tong2023improving}) between 10K generated data points and 10K samples from the ground-truth {\tt moons} distribution.
We experiment with single-step Euler's method (Euler-1) and the adaptive Dormand–Prince method~\cite{dormand1980family} (dopri5, referred to as adaptive) for numerical integration.
Additionally, we report the number of function evaluations (NFE) in the adaptive method, which is the number of forward passes through the neural network.
We average the results over ten training runs with different random seeds and report mean\(\pm\)std.

\paragraph{Results.}
The results are summarized in \Cref{tab:two_moons} and visualized in \Cref{fig:teaser}.
In the \emph{unconditional} setting, OT produces straighter flows, leading to better distribution matching with fewer NFEs, and our method does not apply.
However, in the \emph{conditional} setting, OT degrades and performs worse than regular FM in the adaptive setting due to the aforementioned train-test gap in OT coupling.
Our method, \ourmethod{}, effectively models the target distribution with both Euler-1 and the adaptive method, improving upon the baselines.

\begin{table*}[t]
\small
    \centering
    \caption{
    Results of different training algorithms on image generation tasks.
    We bold the best-performing entry and underline the second-best.
    Our proposed method, \ourmethod{}, achieves the best overall performance -- better than FM with few sampling steps and better than OT with more sampling steps.
    In cases where \ourmethod{} performs second, our performance is often comparable to the best entry.
    Additionally, \ourmethod{} has the most stable performance, with the smallest deviations across runs in most entries.
    Variations in CLIP scores are minimal.
    }
    \begin{NiceTabular}{l@{\hspace{1.1mm}}l@{\hspace{1.6mm}}r@{}l@{\hspace{1.6mm}}r@{}l@{\hspace{1.6mm}}r@{}l@{\hspace{1.6mm}}r@{}l@{\hspace{1.6mm}}r@{}l@{\hspace{1.6mm}}r@{}l@{\hspace{1.6mm}}r@{}l@{\hspace{1.6mm}}r@{}l}
\CodeBefore
\rectanglecolor{defaultColor}{5-2}{5-18}
\rectanglecolor{defaultColor}{8-2}{8-18}
\rectanglecolor{defaultColor}{12-2}{12-18}
\rectanglecolor{defaultColor}{15-2}{15-18}
\rectanglecolor{defaultColor}{19-2}{19-18}
\rectanglecolor{defaultColor}{22-2}{22-18}
\Body
    \toprule
    \multicolumn{18}{c}{\textbf{CIFAR-10 Class-Conditioned Generation}}\\
    \toprule
    & Method & \multicolumn{2}{c}{Euler-2} & \multicolumn{2}{c}{Euler-5} & \multicolumn{2}{c}{Euler-10} & \multicolumn{2}{c}{Euler-25} & \multicolumn{2}{c}{Euler-50} & \multicolumn{2}{c}{Euler-100} & \multicolumn{2}{c}{Adaptive} & \multicolumn{2}{c}{NFE ↓} \\
    \midrule
    \parbox[t]{3mm}{\multirow{3}{*}{\rotatebox[origin=c]{90}{FID ↓}}} 
    & FM & 105.481&\(\pm\)1.619 & 21.968&\(\pm\)0.400 & \underline{10.479}&\(\pm\)0.237 & \textbf{5.640}&\(\pm\)0.117 & \textbf{4.128}&\(\pm\)0.069 & \textbf{3.429}&\(\pm\)0.054 & \underline{2.922}&\(\pm\)0.039 & \textbf{124.0}&\(\pm\)2.5 \\
    & OT & \textbf{64.417}&\(\pm\)0.434 & \underline{18.194}&\(\pm\)0.409 & 10.778&\(\pm\)0.270 & 6.821&\(\pm\)0.186 & 5.368&\(\pm\)0.113 & 4.671&\(\pm\)0.066 & 4.144&\(\pm\)0.018 & \underline{125.7}&\(\pm\)0.4 \\
    & \ourmethod{} (ours) & \underline{64.694}&\(\pm\)0.379 & \textbf{17.565}&\(\pm\)0.202 & \textbf{9.762}&\(\pm\)0.136 & \underline{5.644}&\(\pm\)0.081 & \underline{4.152}&\(\pm\)0.061 & \underline{3.447}&\(\pm\)0.049 & \textbf{2.904}&\(\pm\)0.035 & 127.7&\(\pm\)0.7 \\
    \midrule
    \parbox[t]{3mm}{\multirow{3}{*}{\rotatebox[origin=c]{90}{CE ↓}}} 
    & FM & 3.343 &\(\pm\)0.022 & \underline{0.546} &\(\pm\)0.020 & \underline{0.323} &\(\pm\)0.007 & \textbf{0.272} &\(\pm\)0.004 & \textbf{0.267} &\(\pm\)0.002 & \textbf{0.269} &\(\pm\)0.003 & \textbf{0.276} &\(\pm\)0.004 & \textbf{124.0} &\(\pm\)2.5 \\
    & OT & \underline{2.529} &\(\pm\)0.010 & 0.624 &\(\pm\)0.012 & 0.426 &\(\pm\)0.005 & 0.369 &\(\pm\)0.004 & 0.363 &\(\pm\)0.004 & 0.364 &\(\pm\)0.005 & 0.372 &\(\pm\)0.005 & \underline{125.7} &\(\pm\)0.4 \\
    & \ourmethod{} (ours) & \textbf{2.264} &\(\pm\)0.022 & \textbf{0.475} &\(\pm\)0.009 & \textbf{0.322} &\(\pm\)0.004 & \underline{0.276} &\(\pm\)0.005 & \underline{0.271} &\(\pm\)0.003 & \underline{0.272} &\(\pm\)0.003 & \underline{0.279} &\(\pm\)0.004 & 127.7 &\(\pm\)0.7 \\
    \toprule
    \multicolumn{18}{c}{\textbf{ImageNet 32\(\times\)32 Caption-Conditioned Generation}}\\
    \midrule
    \parbox[t]{3mm}{\multirow{3}{*}{\rotatebox[origin=c]{90}{FID ↓}}} 
    & FM & 113.250 &\(\pm\)0.793 & 23.015 &\(\pm\)0.123 & \underline{11.141} &\(\pm\)0.051 & \underline{7.165} &\(\pm\)0.087 & \underline{6.142} &\(\pm\)0.085 & \underline{5.673} &\(\pm\)0.074 & \underline{5.358} &\(\pm\)0.059 & 146.9 &\(\pm\)3.5 \\
    & OT & \textbf{81.317} &\(\pm\)0.369 & \textbf{20.763} &\(\pm\)0.154 & 11.360 &\(\pm\)0.089 & 7.854 &\(\pm\)0.077 & 6.899 &\(\pm\)0.079 & 6.469 &\(\pm\)0.071 & 6.220 &\(\pm\)0.065 & \underline{137.7} &\(\pm\)0.2 \\
    & \ourmethod{} (ours) & \underline{102.380} &\(\pm\)0.279 & \underline{21.965} &\(\pm\)0.035 & \textbf{10.897} &\(\pm\)0.033 & \textbf{7.069} &\(\pm\)0.027 & \textbf{6.084} &\(\pm\)0.016 & \textbf{5.638} &\(\pm\)0.018 & \textbf{5.350} &\(\pm\)0.017 & \textbf{134.2} &\(\pm\)1.4 \\
    \midrule
    \parbox[t]{3mm}{\multirow{3}{*}{\rotatebox[origin=c]{90}{CLIP ↑}}} 
    & FM & 0.067&\(\pm\)0.000 & \textbf{0.076}&\(\pm\)0.000 & \textbf{0.074}&\(\pm\)0.000 & \textbf{0.071}&\(\pm\)0.000 & 0.071&\(\pm\)0.000 & \textbf{0.070}&\(\pm\)0.000 & \textbf{0.069}&\(\pm\)0.000 & 146.9&\(\pm\)3.5 \\
    & OT & 0.067&\(\pm\)0.000 & 0.071&\(\pm\)0.000 & 0.070&\(\pm\)0.000 & 0.068&\(\pm\)0.000 & 0.068&\(\pm\)0.000 & 0.067&\(\pm\)0.000 & 0.067&\(\pm\)0.000 & 137.7&\(\pm\)0.2 \\
    & \ourmethod{} (ours) & \textbf{0.068}&\(\pm\)0.000 & \underline{0.075}&\(\pm\)0.000 & \underline{0.073}&\(\pm\)0.000 & \textbf{0.071}&\(\pm\)0.000 & \underline{0.070}&\(\pm\)0.000 & \textbf{0.070}&\(\pm\)0.000 & \textbf{0.069}&\(\pm\)0.000 & \textbf{134.2}&\(\pm\)1.4 \\
    \toprule
    \multicolumn{18}{c}{\textbf{ImageNet 256\(\times\)256 Caption-Conditioned Generation in Latent Space}}\\
    \midrule
    \parbox[t]{3mm}{\multirow{3}{*}{\rotatebox[origin=c]{90}{FID ↓}}} 
    & FM & 
    202.519 &\(\pm\)0.606 & \underline{31.266} &\(\pm\)0.355 & \underline{9.996} &\(\pm\)0.106 & \underline{5.189} &\(\pm\)0.082 & \underline{3.886} &\(\pm\)0.066 & \underline{3.484} &\(\pm\)0.055 & \underline{3.498} &\(\pm\)0.054 & 131.1 &\(\pm\)2.3 \\
    & OT & 
    \textbf{193.376} &\(\pm\)1.665 & 42.780 &\(\pm\)2.132 & 15.638 &\(\pm\)1.510 & 7.514 &\(\pm\)1.034 & 5.693 &\(\pm\)0.998 & 5.360 &\(\pm\)1.016 & 5.726 &\(\pm\)1.074 & \underline{125.6} &\(\pm\)6.5  \\
    & \ourmethod{} (ours) & 
    \underline{200.811} &\(\pm\)0.385 & \textbf{30.715} &\(\pm\)0.221 & \textbf{10.000} &\(\pm\)0.072 & \textbf{5.069} &\(\pm\)0.037 & \textbf{3.722} &\(\pm\)0.037 & \textbf{3.320} &\(\pm\)0.032 & \textbf{3.369} &\(\pm\)0.034 & \underline{128.2} &\(\pm\)2.6 \\
    \midrule
    \parbox[t]{3mm}{\multirow{3}{*}{\rotatebox[origin=c]{90}{CLIP ↑}}} 
    & FM & 
    0.028 &\(\pm\)0.000 & \textbf{0.120} &\(\pm\)0.000 & \textbf{0.134} &\(\pm\)0.000 & \textbf{0.137 }&\(\pm\)0.000 & \textbf{0.138} &\(\pm\)0.000 & \textbf{0.139 }&\(\pm\)0.000 & \textbf{0.140} &\(\pm\)0.000 & 131.2 &\(\pm\)2.3 \\
    & OT & 
    \textbf{0.031} &\(\pm\)0.000 & 0.107 &\(\pm\)0.003 & 0.120 &\(\pm\)0.003 & 0.123 &\(\pm\)0.003 & 0.124 &\(\pm\)0.003 & 0.124 &\(\pm\)0.004 & 0.125 &\(\pm\)0.004 & \textbf{125.6} &\(\pm\)6.5 \\
    & \ourmethod{} (ours) & 
    \underline{0.029} &\(\pm\)0.000 & \textbf{0.120} &\(\pm\)0.000 & \underline{0.133} &\(\pm\)0.000 & \textbf{0.137} &\(\pm\)0.000 & \textbf{0.138} &\(\pm\)0.000 & \underline{0.138} &\(\pm\)0.000 & \underline{0.139} &\(\pm\)0.000 & \underline{128.2} &\(\pm\)2.6 \\
    \bottomrule
\end{NiceTabular}

    \label{tab:main}
\end{table*}

\subsection{High-Dimensional Image Data}

We conduct experiments on CIFAR-10~\cite{krizhevsky2009learning} (image space, class-conditioned), ImageNet-32\(\times\)32~\cite{deng2009imagenet} (image space, caption-conditioned), and ImageNet-256\(\times\)256~\cite{deng2009imagenet} (latent space, caption-conditioned).
For ImageNet, we use the captions provided by~\cite{imagenetcaptioned} and encode text features using a CLIP~\cite{radford2021learning}-like DFN~\cite{fang2023data} text encoder as input conditions.
Our results are summarized in \Cref{tab:main}.

\paragraph{Metrics.}
We report Fréchet inception distance (FID)~\cite{heusel2017gans} and the number of function evaluations (NFE) in the adaptive solver for all three datasets.
To assess how well the generations adhere to the input conditions, we additionally compute condition adherence metrics.
For CIFAR-10, we run a pretrained classifier~\cite{cifarmodels} on the generated images to obtain logits and report the average cross entropy (CE) with the input class label.
For ImageNet, we compute CLIP features~\cite{radford2021learning} using SigLIP-2~\cite{tschannen2025siglip} from the generated images and the input captions respectively, and report the average cosine similarities (CLIP).

\paragraph{CIFAR-10.}
We base our UNet architecture on~\citet{tong2023improving}, with details provided in \Cref{sec:app:cifar}. 
The CIFAR-10 training set contains 50,000 32\(\times\)32 color images across 10 object classes.
During testing, we generate 50,000 images evenly distributed among classes and assess FID against the training set following~\cite{tong2023improving}.
Each model is trained five times with different random seeds and the results are reported as mean\(\pm\)std.
We use an OT batch size \(\batchsize\) of 640 and a ratio \(\targetratio\) of 0.01.
Compared to FM, our method achieves better performance (FID and CE) in few-steps settings and comparable performance in many-steps settings.
Compared to OT, our method significantly outperforms in the many-step setting. 
Although OT achieves slightly better FID in Euler-2, it performs worse in CE, indicating that it fails to follow the condition.

\paragraph{ImageNet 32\(\times\)32.}
We base our UNet architecture on~\citet{pooladian2023multisample}, and list details in \Cref{sec:app:imagenet32}.
We train and evaluate on the face-blurred~\cite{yang2022study} ImageNet, following~\cite{pooladian2023multisample}.
In contrast to prior works that assess FID against the training set, we assess FID against the validation set (49,997 images) %
, as the fine-grained nature of captions may lead to overfitting on the training set.
Each model is trained three times with different random seeds and the results are reported as mean\(\pm\)std.
Compared to CIFAR-10, there are much more variations in this dataset (\eg, 1000 classes \vs 10 classes in CIFAR-10). 
Thus, we adopt a larger OT batch size \(\batchsize\) of 6400 while keeping the same target ratio \(\targetratio\) 0.01 as before.
Similar to our findings in CIFAR-10, our method performs better or is comparable to FM; in few-steps settings, OT achieves better FID but worse condition following (CLIP).
Our method strikes the best overall balance.

\begin{figure}
    \centering
    \includegraphics[width=\linewidth]{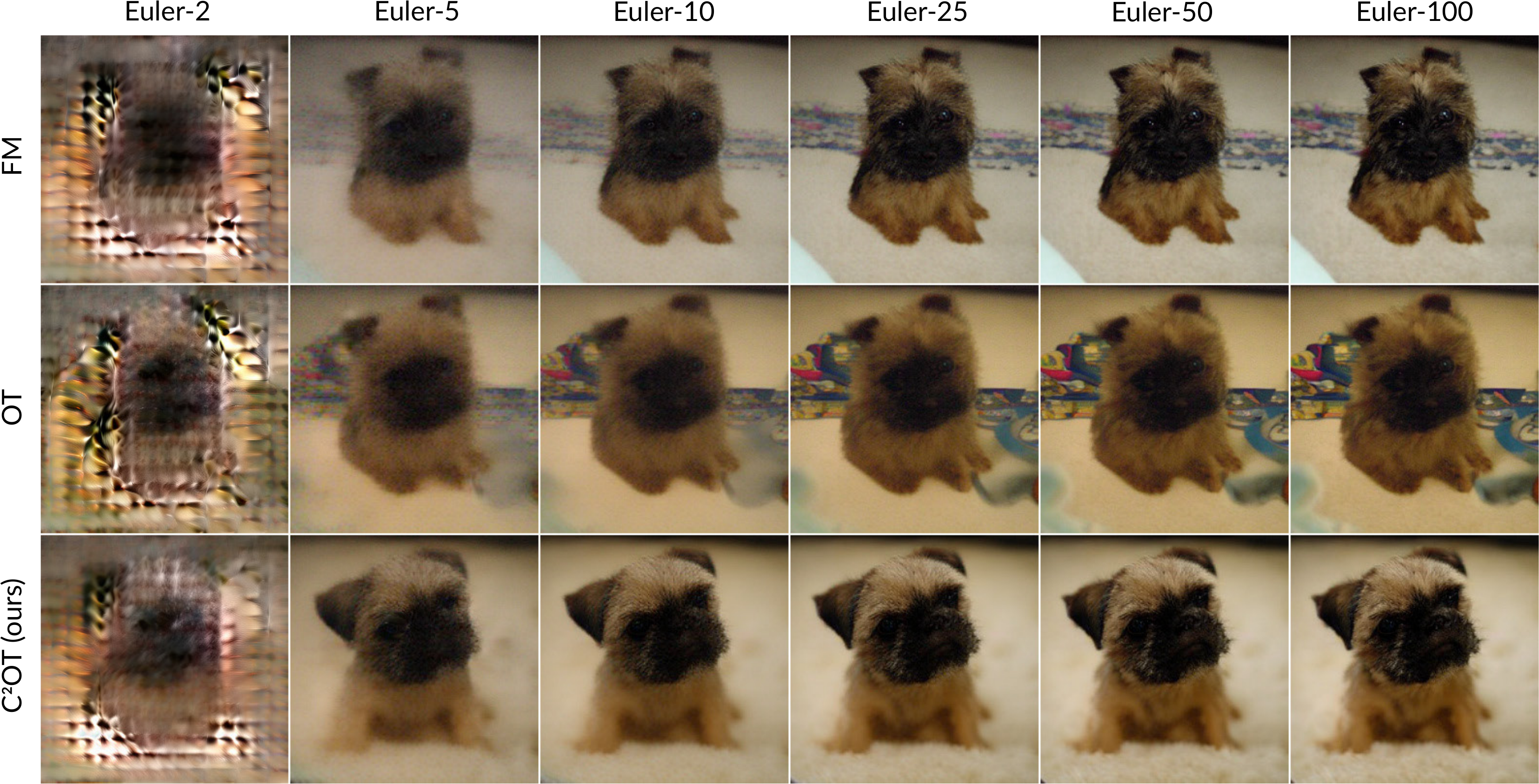}
    \caption{Visual comparisons of 256\(\times\)256 images generated by the baselines and our approach with different amounts of sampling steps.
    Our approach converges faster and produces cleaner outputs.
    The input caption is ``a small dog sitting on a carpet''.
    }
    \label{fig:visualize_imagenet}
\end{figure}

\paragraph{ImageNet 256\(\times\)256 in Latent Space.}
Here, we explore latent flow matching models~\cite{rombach2022high}.
During training, the flow matching model learns to generate data in the latent space, which is encoded from images.
At test-time, the generated latents are decoded into images using a pretrained decoder.
We base our experiment on a recent transformer-based model, LightningDiT~\cite{yao2025reconstruction}, demonstrating that our method is effective across different network architectures and data spaces.
We use the officially provided ``64 epochs'' configuration for all experiments (due to computational constraints), which is not directly comparable to methods trained with 800 epochs.
We follow the same evaluation protocol as ImageNet-32\(\times\)32, and adopt the same OT batch size \(\batchsize\) of 6400 and the target ratio \(\targetratio\) of 0.01.
Each model is trained five times with different random seeds, and the results are reported as mean\(\pm\)std.
Again, \ourmethod{} has the best overall performance compared to both FM and OT.
We visualize the generations in \Cref{fig:visualize_imagenet}. %

\paragraph{Limitations in High-Dimensional Data}
We observed that the benefits of \ourmethod{} are less pronounced, though still visible, in higher-dimensional image data compared to low-dimensional data. 
We hypothesize this is because image data is more entangled, making the prior skew induced by OT less detrimental.
We also observe statistically straighter paths in \ourmethod{} compared to FM, with details given in \Cref{tab:app:path_straightness}.

\subsection{Ablation Studies}

\subsubsection[Varying OT Batch Size]{Varying OT Batch Size \(\batchsize\)}
\label{sec:vary_ot_batch_size}
Here, we analyze the effect of varying the OT batch size \(\batchsize\).
We conduct these experiments on the CIFAR-10 dataset, training each model three times with different random seeds for every choice of \(\batchsize\). 
We plot the results in \Cref{fig:ot_batch_size}.
Increasing the OT batch size clearly improves FID in the few-sampling-steps setting (Euler-2).
This is intuitive, because a larger OT batch size enables the coupling to approach the \emph{true OT} results, \ie, dataset-level optimal transport. 
In turn, this leads to straighter paths.
However, in the adaptive setting, increasing the OT batch size slightly worsens FID, though the effect is much less significant (the range of the blue y-axis is much smaller than the range of the red y-axis).
We think this occurs because a more accurate OT coupling reduces variations in the training data (similar to having less data augmentations) which harms performance as the benefit of straighter flows diminishes in the adaptive setting.
This observation aligns with the findings of~\citet{pooladian2023multisample}:  computing OT coupling across GPUs (\ie, having a larger effective OT batch size) slightly harms results.
In contrast to~\citet{pooladian2023multisample}, though, we seek to counteract the reduced effective OT batch size induced by conditional optimal transport (as discussed in \Cref{sec:oversampling}), and we do find reasonably large (640 for CIFAR, 6400 for ImageNet) OT batch sizes useful while not adding wall-clock overhead.

\begin{figure}
    \centering
    \begin{adjustbox}{width=\linewidth}
    \begin{tikzpicture}
    \pgfplotsset{
        log ticks with fixed point,
        xmin=128, xmax=6400,
        xtick = {
            128, 640, 1280, 2560, 6400
        },
    }
    
    \begin{axis}[
      scale only axis,
      axis y line*=left,
      xlabel={OT batch size \(\batchsize\)},
      ylabel={\textcolor{red}{Euler-2 FID ↓}},
      enlarge x limits=0.05,   %
      grid=both,               %
      ymin=50, ymax=75,       %
      restrict y to domain=50:120,
      every axis plot/.append style={thick},
      xmode=log,
      minor tick num=4,
      every y tick label/.append style={red},
      ytick={50,55,60,65,70,75},
      width=9cm,height=6cm,
    ]

    \addplot[name path=upper1, draw=none] coordinates{
        (128,74.5172 + 0.0614)
        (640,64.8487 + 0.3867)
        (1280,61.8505 + 0.3042)
        (2560,59.3612 + 0.2646)
        (6400,56.0108 + 0.4537)
    };

    \addplot[name path=lower1, draw=none] coordinates{
        (128,74.5172 - 0.0614)
        (640,64.8487 - 0.3867)
        (1280,61.8505 - 0.3042)
        (2560,59.3612 - 0.2646)
        (6400,56.0108 - 0.4537)
    };
    
    \addplot[mark=x,red]
      coordinates{
        (128,74.5172)
        (640,64.8487)
        (1280,61.8505)
        (2560,59.3612)
        (6400,56.0108)
    }; \label{euler2}

    \addplot[blue, fill=red, fill opacity=0.2] 
      fill between[of=upper1 and lower1];

    \addplot[only marks, mark=*, mark options={fill=red}] coordinates {(128,64.694)};
    \node[above] at (axis cs:180,64.694) {\small OT (Euler-2)};
        
    \end{axis}
    
    \begin{axis}[
      scale only axis,
      at={(current axis.south west)},
      anchor=south west,
      legend pos=south east,
      axis y line*=right,
      axis x line=none,
      ylabel={\textcolor{blue}{Adaptive FID ↓}},
      grid=both,              %
        enlarge x limits=0.05,   %
      ymin=2.85, ymax=3.0,      %
      restrict y to domain=2.8:3.0,
      grid=none,
      tick label style={
                /pgf/number format/fixed,
                /pgf/number format/fixed zerofill,
                /pgf/number format/precision=2
            },
        ytick={
          2.85,
          2.88,
          2.91,
          2.94,
          2.97,
          3.0
        },
        every axis plot/.append style={thick},
        xmode=log,
        every y tick label/.append style={blue},
        width=9cm,height=6cm,
    ]
    \addlegendimage{/pgfplots/refstyle=euler2}\addlegendentry{Ours, Euler-2}
    \addlegendimage{/pgfplots/refstyle=adaptive}\addlegendentry{Ours, adaptive}

    \addplot[name path=upper2, draw=none] coordinates{
        (128,2.8804 + 0.0481)
        (640,2.9051 + 0.0381)
        (1280,2.9511 + 0.0386)
        (2560,2.9413 + 0.0243)
        (6400,2.9449 + 0.0174)
    };

    \addplot[name path=lower2, draw=none] coordinates{
        (128,2.8804 - 0.0481)
        (640,2.9051 - 0.0381)
        (1280,2.9511 - 0.0386)
        (2560,2.9413 - 0.0243)
        (6400,2.9449 - 0.0174)
    };
    
    \addplot[mark=o,blue]
      coordinates{
        (128,2.8804)
        (640,2.9051)
        (1280,2.9511)
        (2560,2.9413)
        (6400,2.9449)
    }; \label{adaptive}
    \addplot[blue, fill=blue, fill opacity=0.2] 
      fill between[of=upper2 and lower2];

    \addplot [domain=128:6400, samples=2, dashed, color=blue] {2.9224};
    \node[above] at (axis cs:4500,2.9224) {\small FM (adaptive)};
      
    \end{axis}
    \end{tikzpicture}
\end{adjustbox}
    \vspace{-1em}
    \caption{
    Changes in FID with respect to varying OT batch sizes \(\batchsize\). 
    We plot mean\(\pm\)std over three runs and represent std with a shaded region.
    OT (adaptive) and FM (Euler-2) perform worse than all other results in this plot, hence results are not shown (full plot in \Cref{sec:app:extended_plots}).
    }
    \label{fig:ot_batch_size}
\end{figure}

\subsubsection[Varying Target Ratio]{Varying Target Ratio \(\targetratio\)}
\label{sec:vary_target_r}
Here, we analyze the effects of varying the target ratio \(\targetratio\), as defined in \Cref{eq:ratio}.
We conduct these experiments on the ImageNet-32\(\times\)32 dataset, training each model three times with different random seeds for every choice of \(\targetratio\). 
We plot the results in \Cref{fig:target_r}.
Recall that when \(r\to0\), \(w\to\infty\) and the method approaches FM; when \(r\to0.5\),\footnote{There is a 0.5 chance that the distance between a random prior-data pair is closer to that of another random prior-data pair.} \(w\to0\) and the method approaches OT (\Cref{sub:method_continuous_cond}).
Our findings are similar to those in \Cref{sec:vary_ot_batch_size}: as we approach \emph{true OT}, whether by increasing OT batch size or increasing \(\targetratio\)), FID improves in the few-step setting and worsens in the adaptive setting.

\begin{figure}
    \centering
    \begin{adjustbox}{width=\linewidth}
    \begin{tikzpicture}
    \pgfplotsset{
        xmode=log,
        log ticks with fixed point,
        xmin=0.0005, xmax=0.1,
        xtick = {
            0.0005, 0.001, 0.002, 0.005, 0.01, 0.02, 0.05, 0.1
        },
    }
    
    \begin{axis}[
      scale only axis,
      axis y line*=left,
      xlabel={Target ratio \(\targetratio\)},
      ylabel={\textcolor{red}{Euler-2 FID ↓}},
      enlarge x limits=0.05,   %
      grid=both,               %
      ymin=80, ymax=115,       %
      restrict y to domain=80:115,
      every axis plot/.append style={thick},
      xmode=log,
      minor tick num=4,
      every y tick label/.append style={red},
      ytick={80,85,90,95,100,105,110,115},
      width=9cm,height=6cm,
    ]

    \addplot[name path=upper3, draw=none] coordinates{
        (0.0005,110.148 + 0.820)
        (0.001,109.555 + 0.579)
        (0.002,107.825 + 0.588)
        (0.005,105.148 + 0.832)
        (0.01,102.380 + 0.279)
        (0.02,97.667 + 0.639)
        (0.05,90.114 + 0.181)
        (0.1,82.456 + 0.343)
    };

    \addplot[name path=lower3, draw=none] coordinates{
        (0.0005,110.148 - 0.820)
        (0.001,109.555 - 0.579)
        (0.002,107.825 - 0.588)
        (0.005,105.148 - 0.832)
        (0.01,102.380 - 0.279)
        (0.02,97.667 - 0.639)
        (0.05,90.114 - 0.181)
        (0.1,82.456 - 0.343)
    };
    
    \addplot[mark=x,red]
      coordinates{
        (0.0005,110.148)
        (0.001,109.555)
        (0.002,107.825)
        (0.005,105.148)
        (0.01,102.380)
        (0.02,97.667)
        (0.05,90.114)
        (0.1,82.456)
    }; \label{r_euler2}

    \addplot[blue, fill=red, fill opacity=0.2] 
      fill between[of=upper3 and lower3];

    \addplot [domain=0.0005:0.1, samples=2, dashed, color=red] {81.3169};
    \node[above, anchor=south west] at (axis cs:0.0005,81.3169) {\small OT (Euler-2)};

    \addplot [domain=0.0005:0.1, samples=2, dashed, color=red] {113.250};
    \node[below, anchor=north west] at (axis cs:0.0005,113.250) {\small FM (Euler-2)};
    
    \end{axis}
    
    \begin{axis}[
      scale only axis,
      at={(current axis.south west)},
      anchor=south west,
      legend style={at={(0.88,0.87)}},
      axis y line*=right,
      axis x line=none,
      ylabel={\textcolor{blue}{Adaptive FID ↓}},
      every tick label/.append style={blue},
      grid=both,              %
        enlarge x limits=0.05,   %
      ymin=5.25, ymax=5.9,      %
      restrict y to domain=5.25:5.9,
      grid=none,
      tick label style={
                /pgf/number format/fixed,
                /pgf/number format/fixed zerofill,
                /pgf/number format/precision=2
            },
        ytick={5.25,5.38,5.51,5.64,5.77,5.9},
        every axis plot/.append style={thick},
        xmode=log,
        width=9cm,height=6cm,
    ]
    \addlegendimage{/pgfplots/refstyle=r_euler2}\addlegendentry{Ours, Euler-2}
    \addlegendimage{/pgfplots/refstyle=r_adaptive}\addlegendentry{Ours, adaptive}

    \addplot[name path=upper4, draw=none] coordinates{
        (0.0005,5.341 + 0.015)
        (0.001,5.334 + 0.002)
        (0.002,5.338 + 0.027)
        (0.005,5.316 + 0.044)
        (0.01,5.350 + 0.017)
        (0.02,5.425 + 0.016)
        (0.05,5.579 + 0.038)
        (0.1,5.843 + 0.054)
    };

    \addplot[name path=lower4, draw=none] coordinates{
        (0.0005,5.341 - 0.015)
        (0.001,5.334 - 0.002)
        (0.002,5.338 - 0.027)
        (0.005,5.316 - 0.044)
        (0.01,5.350 - 0.017)
        (0.02,5.425 - 0.016)
        (0.05,5.579 - 0.038)
        (0.1,5.843 - 0.054)
    };
    
    \addplot[mark=o,blue]
      coordinates{
        (0.0005,5.341)
        (0.001,5.334)
        (0.002,5.338)
        (0.005,5.316)
        (0.01,5.350)
        (0.02,5.425)
        (0.05,5.579)
        (0.1,5.843)
    }; \label{r_adaptive}
    \addplot[blue, fill=blue, fill opacity=0.2] 
      fill between[of=upper4 and lower4];

    \addplot [domain=0.0005:0.1, samples=2, dashed, color=blue] {5.358};
    \node[above, anchor=south west] at (axis cs:0.0005,5.358) {\small FM (adaptive)};
      
    \end{axis}
    \end{tikzpicture}
\end{adjustbox}
    \vspace{-1em}
    \caption{
    Changes in FID with respect to varying target ratio \(\targetratio\). 
    We plot mean\(\pm\)std over three runs and represent std with a shaded region.
    OT (adaptive) performs worse than all other results in this plot, hence results are not shown  (full plot in \Cref{sec:app:extended_plots}).
    }
    \label{fig:target_r}
\end{figure}

\vspace{-1em}
\section{Conclusion}
\label{sec:conclusion}
We first investigate and formalize the struggle of minibatch optimal transport in conditional generation.
Based on the findings, we propose \ourmethod{}, a simple yet effective addition that corrects the degeneration of OT while maintaining straight integration paths.
Extensive experiments verify the effectiveness of our method.
We believe that these findings, along with the proposed simple technique, close a gap in current research on prior-data coupling.

{
\scriptsize
\noindent\textbf{Acknowledgment.}
This work is supported in part by NSF grants 2008387, 2045586, 2106825, NIFA award 2020-67021-32799, and OAC 2320345.
\par
}

{
    \small
    \bibliographystyle{ieeenat_fullname}
    \bibliography{main}

\begin{thebibliography}{48}
\providecommand{\natexlab}[1]{#1}
\providecommand{\url}[1]{\texttt{#1}}
\expandafter\ifx\csname urlstyle\endcsname\relax
  \providecommand{\doi}[1]{doi: #1}\else
  \providecommand{\doi}{doi: \begingroup \urlstyle{rm}\Url}\fi

\bibitem[Albergo and Vanden-Eijnden(2023)]{albergo2023building}
Michael Albergo and Eric Vanden-Eijnden.
\newblock Building normalizing flows with stochastic interpolants.
\newblock In \emph{Proc. ICLR}, 2023.

\bibitem[Albergo et~al.(2023)Albergo, Boffi, and Vanden-Eijnden]{albergo2023stochastic}
Michael Albergo, Nicholas Boffi, and Eric Vanden-Eijnden.
\newblock Stochastic interpolants: A unifying framework for flows and diffusions.
\newblock \emph{arXiv preprint arXiv:2303.08797}, 2023.

\bibitem[Black et~al.(2024)Black, Brown, Driess, Esmail, Equi, Finn, Fusai, Groom, Hausman, Ichter, et~al.]{black2024pi_0}
Kevin Black, Noah Brown, Danny Driess, Adnan Esmail, Michael Equi, Chelsea Finn, Niccolo Fusai, Lachy Groom, Karol Hausman, Brian Ichter, et~al.
\newblock pi0: A vision-language-action flow model for general robot control.
\newblock \emph{arXiv}, 2024.

\bibitem[Cheng et~al.(2025)Cheng, Ishii, Hayakawa, Shibuya, Schwing, and Mitsufuji]{cheng2024taming}
Ho~Kei Cheng, Masato Ishii, Akio Hayakawa, Takashi Shibuya, Alexander Schwing, and Yuki Mitsufuji.
\newblock Taming multimodal joint training for high-quality video-to-audio synthesis.
\newblock \emph{CVPR}, 2025.

\bibitem[chenyaofo(2023)]{cifarmodels}
chenyaofo.
\newblock Pytorch cifar models.
\newblock \url{https://github.com/chenyaofo/pytorch-cifar-models}, 2023.

\bibitem[Chrabaszcz et~al.(2017)Chrabaszcz, Loshchilov, and Hutter]{chrabaszcz2017downsampled}
Patryk Chrabaszcz, Ilya Loshchilov, and Frank Hutter.
\newblock A downsampled variant of imagenet as an alternative to the cifar datasets.
\newblock \emph{arXiv}, 2017.

\bibitem[Davtyan et~al.(2025)Davtyan, Dadi, Cevher, and Favaro]{davtyan2025faster}
Aram Davtyan, Leello~Tadesse Dadi, Volkan Cevher, and Paolo Favaro.
\newblock Faster inference of flow-based generative models via improved data-noise coupling.
\newblock In \emph{ICLR}, 2025.

\bibitem[Deng et~al.(2009)Deng, Dong, Socher, Li, Li, and Fei-Fei]{deng2009imagenet}
Jia Deng, Wei Dong, Richard Socher, Li-Jia Li, Kai Li, and Li Fei-Fei.
\newblock Imagenet: A large-scale hierarchical image database.
\newblock In \emph{CVPR}, 2009.

\bibitem[Dhariwal and Nichol(2021)]{dhariwal2021diffusion}
Prafulla Dhariwal and Alexander Nichol.
\newblock Diffusion models beat gans on image synthesis.
\newblock \emph{NeurIPS}, 2021.

\bibitem[Dormand and Prince(1980)]{dormand1980family}
John~R Dormand and Peter~J Prince.
\newblock A family of embedded runge-kutta formulae.
\newblock \emph{Journal of computational and applied mathematics}, 1980.

\bibitem[Esser et~al.(2024)Esser, Kulal, Blattmann, Entezari, M{\"u}ller, Saini, Levi, Lorenz, Sauer, Boesel, et~al.]{esser2024scaling}
Patrick Esser, Sumith Kulal, Andreas Blattmann, Rahim Entezari, Jonas M{\"u}ller, Harry Saini, Yam Levi, Dominik Lorenz, Axel Sauer, Frederic Boesel, et~al.
\newblock Scaling rectified flow transformers for high-resolution image synthesis.
\newblock In \emph{ICML}, 2024.

\bibitem[Fang et~al.(2024)Fang, Jose, Jain, Schmidt, Toshev, and Shankar]{fang2023data}
Alex Fang, Albin~Madappally Jose, Amit Jain, Ludwig Schmidt, Alexander Toshev, and Vaishaal Shankar.
\newblock Data filtering networks.
\newblock \emph{ICLR}, 2024.

\bibitem[Geng et~al.(2025)Geng, Pokle, Luo, Lin, and Kolter]{geng2024consistency}
Zhengyang Geng, Ashwini Pokle, William Luo, Justin Lin, and J~Zico Kolter.
\newblock Consistency models made easy.
\newblock \emph{ICLR}, 2025.

\bibitem[Guo and Schwing(2025)]{guo2025variational}
Pengsheng Guo and Alexander~G Schwing.
\newblock Variational rectified flow matching.
\newblock \emph{arXiv}, 2025.

\bibitem[Hendrycks and Gimpel(2016)]{hendrycks2016gaussian}
Dan Hendrycks and Kevin Gimpel.
\newblock Gaussian error linear units (gelus).
\newblock \emph{arXiv}, 2016.

\bibitem[Heusel et~al.(2017)Heusel, Ramsauer, Unterthiner, Nessler, and Hochreiter]{heusel2017gans}
Martin Heusel, Hubert Ramsauer, Thomas Unterthiner, Bernhard Nessler, and Sepp Hochreiter.
\newblock Gans trained by a two time-scale update rule converge to a local nash equilibrium.
\newblock \emph{NeurIPS}, 2017.

\bibitem[Higgins et~al.(2017)Higgins, Matthey, Pal, Burgess, Glorot, Botvinick, Mohamed, and Lerchner]{higgins2017beta}
Irina Higgins, Loic Matthey, Arka Pal, Christopher Burgess, Xavier Glorot, Matthew Botvinick, Shakir Mohamed, and Alexander Lerchner.
\newblock beta-vae: Learning basic visual concepts with a constrained variational framework.
\newblock In \emph{ICLR}, 2017.

\bibitem[Hui et~al.(2025)Hui, Liu, Zeng, Fu, and Vahdat]{hui2025not}
Ka-Hei Hui, Chao Liu, Xiaohui Zeng, Chi-Wing Fu, and Arash Vahdat.
\newblock Not-so-optimal transport flows for 3d point cloud generation.
\newblock \emph{arXiv}, 2025.

\bibitem[Ilharco et~al.(2021)Ilharco, Wortsman, Wightman, Gordon, Carlini, Taori, Dave, Shankar, Namkoong, Miller, Hajishirzi, Farhadi, and Schmidt]{ilharco_gabriel_2021_5143773}
Gabriel Ilharco, Mitchell Wortsman, Ross Wightman, Cade Gordon, Nicholas Carlini, Rohan Taori, Achal Dave, Vaishaal Shankar, Hongseok Namkoong, John Miller, Hannaneh Hajishirzi, Ali Farhadi, and Ludwig Schmidt.
\newblock Openclip, 2021.

\bibitem[Kim et~al.(2024)Kim, Lai, Liao, Murata, Takida, Uesaka, He, Mitsufuji, and Ermon]{kim2023consistency}
Dongjun Kim, Chieh-Hsin Lai, Wei-Hsiang Liao, Naoki Murata, Yuhta Takida, Toshimitsu Uesaka, Yutong He, Yuki Mitsufuji, and Stefano Ermon.
\newblock Consistency trajectory models: Learning probability flow ode trajectory of diffusion.
\newblock \emph{ICLR}, 2024.

\bibitem[Kingma and Ba(2015)]{kingma2014adam}
Diederik~P Kingma and Jimmy Ba.
\newblock Adam: A method for stochastic optimization.
\newblock \emph{ICLR}, 2015.

\bibitem[Kingma et~al.(2014)Kingma, Welling, et~al.]{kingma2013auto}
Diederik~P Kingma, Max Welling, et~al.
\newblock Auto-encoding variational bayes.
\newblock In \emph{ICLR}, 2014.

\bibitem[Klein et~al.(2023)Klein, Kr{\"a}mer, and No{\'e}]{klein2023equivariant}
Leon Klein, Andreas Kr{\"a}mer, and Frank No{\'e}.
\newblock Equivariant flow matching.
\newblock \emph{NeurIPS}, 36, 2023.

\bibitem[Krizhevsky and Hinton(2009)]{krizhevsky2009learning}
Alex Krizhevsky and Geoffrey Hinton.
\newblock Learning multiple layers of features from tiny images, 2009.

\bibitem[Kuhn(1955)]{kuhn1955hungarian}
Harold~W. Kuhn.
\newblock The hungarian method for the assignment problem.
\newblock \emph{Naval research logistics quarterly}, 1955.

\bibitem[Layer(2024)]{imagenetcaptioned}
Visual Layer.
\newblock Imagenet-1k-vl-enriched.
\newblock \url{https://huggingface.co/datasets/visual-layer/imagenet-1k-vl-enriched}, 2024.

\bibitem[Lee et~al.(2023)Lee, Kim, and Ye]{lee2023minimizing}
Sangyun Lee, Beomsu Kim, and Jong~Chul Ye.
\newblock Minimizing trajectory curvature of ode-based generative models.
\newblock In \emph{ICML}, 2023.

\bibitem[Lipman et~al.(2023)Lipman, Chen, Ben-Hamu, Nickel, and Le]{lipman2022flow}
Yaron Lipman, Ricky~TQ Chen, Heli Ben-Hamu, Maximilian Nickel, and Matt Le.
\newblock Flow matching for generative modeling.
\newblock \emph{ICLR}, 2023.

\bibitem[Lipman et~al.(2024)Lipman, Havasi, Holderrieth, Shaul, Le, Karrer, Chen, Lopez-Paz, Ben-Hamu, and Gat]{lipman2024flow}
Yaron Lipman, Marton Havasi, Peter Holderrieth, Neta Shaul, Matt Le, Brian Karrer, Ricky~TQ Chen, David Lopez-Paz, Heli Ben-Hamu, and Itai Gat.
\newblock Flow matching guide and code.
\newblock \emph{arXiv}, 2024.

\bibitem[Liu et~al.(2024)Liu, Yin, Yuille, Brown, and Singh]{liu2024flowing}
Qihao Liu, Xi Yin, Alan Yuille, Andrew Brown, and Mannat Singh.
\newblock Flowing from words to pixels: A framework for cross-modality evolution.
\newblock \emph{arXiv preprint arXiv:2412.15213}, 2024.

\bibitem[Liu et~al.(2023)Liu, Gong, and Liu]{liu2022flow}
Xingchao Liu, Chengyue Gong, and Qiang Liu.
\newblock Flow straight and fast: Learning to generate and transfer data with rectified flow.
\newblock \emph{ICLR}, 2023.

\bibitem[Lu and Song(2024)]{lu2024simplifying}
Cheng Lu and Yang Song.
\newblock Simplifying, stabilizing and scaling continuous-time consistency models.
\newblock \emph{ICLR}, 2024.

\bibitem[Martin et~al.(2025)Martin, Gagneux, Hagemann, and Steidl]{martin2024pnp}
S{\'e}gol{\`e}ne Martin, Anne Gagneux, Paul Hagemann, and Gabriele Steidl.
\newblock Pnp-flow: Plug-and-play image restoration with flow matching.
\newblock \emph{ICLR}, 2025.

\bibitem[Parmar et~al.(2022)Parmar, Zhang, and Zhu]{parmar2022aliased}
Gaurav Parmar, Richard Zhang, and Jun-Yan Zhu.
\newblock On aliased resizing and surprising subtleties in gan evaluation.
\newblock In \emph{CVPR}, 2022.

\bibitem[Poli et~al.(2021)Poli, Massaroli, Yamashita, Asama, Park, and Ermon]{politorchdyn}
Michael Poli, Stefano Massaroli, Atsushi Yamashita, Hajime Asama, Jinkyoo Park, and Stefano Ermon.
\newblock Torchdyn: Implicit models and neural numerical methods in pytorch.
\newblock \emph{Physical Reasoning and Inductive Biases for the Real World at NeurIPS}, 2021.

\bibitem[Polyak et~al.(2024)Polyak, Zohar, Brown, Tjandra, Sinha, Lee, Vyas, Shi, Ma, Chuang, Yan, Choudhary, Wang, Sethi, Pang, Ma, Misra, Hou, Wang, Jagadeesh, Li, Zhang, Singh, Williamson, Le, Yu, Singh, Zhang, Vajda, Duval, Girdhar, Sumbaly, Rambhatla, Tsai, Azadi, Datta, Chen, Bell, Ramaswamy, Sheynin, Bhattacharya, Motwani, Xu, Li, Hou, Hsu, Yin, Dai, Taigman, Luo, Liu, Wu, Zhao, Kirstain, He, He, Pumarola, Thabet, Sanakoyeu, Mallya, Guo, Araya, Kerr, Wood, Liu, Peng, Vengertsev, Schonfeld, Blanchard, Juefei-Xu, Nord, Liang, Hoffman, Kohler, Fire, Sivakumar, Chen, Yu, Gao, Georgopoulos, Moritz, Sampson, Li, Parmeggiani, Fine, Fowler, Petrovic, and Du]{polyak2025moviegencastmedia}
Adam Polyak, Amit Zohar, Andrew Brown, Andros Tjandra, Animesh Sinha, Ann Lee, Apoorv Vyas, Bowen Shi, Chih-Yao Ma, Ching-Yao Chuang, David Yan, Dhruv Choudhary, Dingkang Wang, Geet Sethi, Guan Pang, Haoyu Ma, Ishan Misra, Ji Hou, Jialiang Wang, Kiran Jagadeesh, Kunpeng Li, Luxin Zhang, Mannat Singh, Mary Williamson, Matt Le, Matthew Yu, Mitesh~Kumar Singh, Peizhao Zhang, Peter Vajda, Quentin Duval, Rohit Girdhar, Roshan Sumbaly, Sai~Saketh Rambhatla, Sam Tsai, Samaneh Azadi, Samyak Datta, Sanyuan Chen, Sean Bell, Sharadh Ramaswamy, Shelly Sheynin, Siddharth Bhattacharya, Simran Motwani, Tao Xu, Tianhe Li, Tingbo Hou, Wei-Ning Hsu, Xi Yin, Xiaoliang Dai, Yaniv Taigman, Yaqiao Luo, Yen-Cheng Liu, Yi-Chiao Wu, Yue Zhao, Yuval Kirstain, Zecheng He, Zijian He, Albert Pumarola, Ali Thabet, Artsiom Sanakoyeu, Arun Mallya, Baishan Guo, Boris Araya, Breena Kerr, Carleigh Wood, Ce Liu, Cen Peng, Dimitry Vengertsev, Edgar Schonfeld, Elliot Blanchard, Felix Juefei-Xu, Fraylie Nord, Jeff Liang, John Hoffman, Jonas
  Kohler, Kaolin Fire, Karthik Sivakumar, Lawrence Chen, Licheng Yu, Luya Gao, Markos Georgopoulos, Rashel Moritz, Sara~K. Sampson, Shikai Li, Simone Parmeggiani, Steve Fine, Tara Fowler, Vladan Petrovic, and Yuming Du.
\newblock Movie gen: A cast of media foundation models.
\newblock \emph{arXiv}, 2024.

\bibitem[Pooladian et~al.(2023)Pooladian, Ben-Hamu, Domingo-Enrich, Amos, Lipman, and Chen]{pooladian2023multisample}
Aram-Alexandre Pooladian, Heli Ben-Hamu, Carles Domingo-Enrich, Brandon Amos, Yaron Lipman, and Ricky~TQ Chen.
\newblock Multisample flow matching: Straightening flows with minibatch couplings.
\newblock \emph{ICML}, 2023.

\bibitem[Radford et~al.(2021)Radford, Kim, Hallacy, Ramesh, Goh, Agarwal, Sastry, Askell, Mishkin, Clark, et~al.]{radford2021learning}
Alec Radford, Jong~Wook Kim, Chris Hallacy, Aditya Ramesh, Gabriel Goh, Sandhini Agarwal, Girish Sastry, Amanda Askell, Pamela Mishkin, Jack Clark, et~al.
\newblock Learning transferable visual models from natural language supervision.
\newblock In \emph{ICML}, 2021.

\bibitem[Rombach et~al.(2022)Rombach, Blattmann, Lorenz, Esser, and Ommer]{rombach2022high}
Robin Rombach, Andreas Blattmann, Dominik Lorenz, Patrick Esser, and Bj{\"o}rn Ommer.
\newblock High-resolution image synthesis with latent diffusion models.
\newblock In \emph{CVPR}, 2022.

\bibitem[Silvestri et~al.(2025)Silvestri, Ambrogioni, Lai, Takida, and Mitsufuji]{silvestri2025training}
Gianluigi Silvestri, Luca Ambrogioni, Chieh-Hsin Lai, Yuhta Takida, and Yuki Mitsufuji.
\newblock Training consistency models with variational noise coupling.
\newblock \emph{arXiv preprint arXiv:2502.18197}, 2025.

\bibitem[Song et~al.(2023{\natexlab{a}})Song, Dhariwal, Chen, and Sutskever]{song2023consistency}
Yang Song, Prafulla Dhariwal, Mark Chen, and Ilya Sutskever.
\newblock Consistency models.
\newblock In \emph{ICML}, 2023{\natexlab{a}}.

\bibitem[Song et~al.(2023{\natexlab{b}})Song, Gong, Xu, Cao, Lan, Ermon, Zhou, and Ma]{song2023equivariant}
Yuxuan Song, Jingjing Gong, Minkai Xu, Ziyao Cao, Yanyan Lan, Stefano Ermon, Hao Zhou, and Wei-Ying Ma.
\newblock Equivariant flow matching with hybrid probability transport for 3d molecule generation.
\newblock \emph{NeurIPS}, 36, 2023{\natexlab{b}}.

\bibitem[Tong et~al.(2024)Tong, Fatras, Malkin, Huguet, Zhang, Rector-Brooks, Wolf, and Bengio]{tong2023improving}
Alexander Tong, Kilian Fatras, Nikolay Malkin, Guillaume Huguet, Yanlei Zhang, Jarrid Rector-Brooks, Guy Wolf, and Yoshua Bengio.
\newblock Improving and generalizing flow-based generative models with minibatch optimal transport.
\newblock \emph{TMLR}, 2024.

\bibitem[Tschannen et~al.(2025)Tschannen, Gritsenko, Wang, Naeem, Alabdulmohsin, Parthasarathy, Evans, Beyer, Xia, Mustafa, et~al.]{tschannen2025siglip}
Michael Tschannen, Alexey Gritsenko, Xiao Wang, Muhammad~Ferjad Naeem, Ibrahim Alabdulmohsin, Nikhil Parthasarathy, Talfan Evans, Lucas Beyer, Ye Xia, Basil Mustafa, et~al.
\newblock Siglip 2: Multilingual vision-language encoders with improved semantic understanding, localization, and dense features.
\newblock \emph{arXiv preprint arXiv:2502.14786}, 2025.

\bibitem[Yang et~al.(2022)Yang, Yau, Fei-Fei, Deng, and Russakovsky]{yang2022study}
Kaiyu Yang, Jacqueline~H Yau, Li Fei-Fei, Jia Deng, and Olga Russakovsky.
\newblock A study of face obfuscation in imagenet.
\newblock In \emph{ICML}, 2022.

\bibitem[Yang et~al.(2024)Yang, Zhang, Zhang, Liu, Xu, Zhang, Meng, Ermon, and Cui]{yang2024consistency}
Ling Yang, Zixiang Zhang, Zhilong Zhang, Xingchao Liu, Minkai Xu, Wentao Zhang, Chenlin Meng, Stefano Ermon, and Bin Cui.
\newblock Consistency flow matching: Defining straight flows with velocity consistency.
\newblock \emph{arXiv}, 2024.

\bibitem[Yao and Wang(2025)]{yao2025reconstruction}
Jingfeng Yao and Xinggang Wang.
\newblock Reconstruction vs. generation: Taming optimization dilemma in latent diffusion models.
\newblock \emph{CVPR}, 2025.

\bibitem[Zhang et~al.(2025)Zhang, Yan, Schwing, and Zhao]{zhang2025towards}
Yichi Zhang, Yici Yan, Alex Schwing, and Zhizhen Zhao.
\newblock Towards hierarchical rectified flow.
\newblock \emph{ICLR}, 2025.

\end{thebibliography}
}

\onecolumn
\beginsupplement

\renewcommand{\contentsname}{Table of Contents}
{
\large
\tableofcontents
}
\clearpage

\section{Additional Experiments}

\subsection{Path Straightness}\label{sec:straightness}

We measure path straightness using Eq.\ (18) from \citep{pooladian2023multisample}, with 10K samples (conditions sampled from the validation sets) and 25 steps in \cref{tab:app:path_straightness}.
Specifically, we measure
\begin{equation}
    \mathbb{E}_{t, q(x_0)}\left[ \lVert v_{\theta, c}(t, \psi_t(x_0)) \rVert^2 - \lVert \psi_1(x_0) - x_0 \rVert^2 \right], 
\end{equation}
where \(\psi_t(x_0)\) denotes the numerical integration result when integrating \(x_0\) from \(0\) to \(t\) following the flow field represented by \(v_{\theta, c}\).
Our method consistently achieves straighter paths than FM. 
Although OT has even straighter paths, the end points of these paths do not accurately model the target distribution, indicated by the higher FID.

\begin{table}[h]
    \centering
    \begin{NiceTabular}{l@{\hspace{6pt}}c@{\hspace{6pt}}c@{\hspace{6pt}}c@{\hspace{6pt}}c@{\hspace{6pt}}c@{\hspace{6pt}}c}
    \toprule
    & \multicolumn{2}{c}{CIFAR} & \multicolumn{2}{c}{ImageNet32} & \multicolumn{2}{c}{ImageNet256} \\
    \cmidrule(lr{\dimexpr 4\tabcolsep-16pt}){2-3}
    \cmidrule(lr{\dimexpr 4\tabcolsep-16pt}){4-5}
    \cmidrule(lr{\dimexpr 4\tabcolsep-12pt}){6-7}
    & Straightness ↓ & FID ↓ & Straightness ↓ & FID ↓ & Straightness ↓ & FID ↓ \\
    \midrule
    FM & 110.83\(\pm\)0.20 & \textbf{5.64} & 133.95\(\pm\)0.34 & 7.17 & 4803.0\(\pm\)79.8 & 3.90 \\
    OT & \textbf{73.54}\(\pm\)0.24 & 6.82 & \textbf{97.71}\(\pm\)0.23 & 7.85 & \textbf{4088.6}\(\pm\)37.5 & 7.48 \\
    \RowStyle[rowcolor=defaultColor]{}
    \ourmethod{} (ours) & 84.99\(\pm\)0.26 & \textbf{5.64} & 122.54\(\pm\)0.34 & \textbf{7.07} & 4625.2\(\pm\)54.5 & \textbf{3.70}\\
    \midrule
    \bottomrule
\end{NiceTabular}

    \caption{Path straightness and FID of FM, OT, and \ourmethod{}.}
    \label{tab:app:path_straightness}
\end{table}

\subsection{Class-to-Image}\label{sec:c2i}

We supplement our existing CIFAR-10 class-to-image experiments (\Cref{tab:main}) with additional class-to-image results on ImageNet32. 
Results in \cref{tab:class_to_image} are consistent with existing findings.

\begin{table}[h]
    \centering
    \begin{NiceTabular}{l@{\hspace{6pt}}c@{\hspace{6pt}}c@{\hspace{6pt}}c@{\hspace{6pt}}c@{\hspace{6pt}}c@{\hspace{6pt}}c@{\hspace{6pt}}c}
    \toprule
    \multicolumn{8}{c}{\textbf{ImageNet 32\(\times\)32 Class-Conditioned Generation}}\\
    \toprule
    Method & Euler-2 & Euler-5 & Euler-10 & Euler-25 & Euler-50 & Euler-100 & Adaptive \\
    \midrule
    FM & 116.296 & 22.224 & 9.530 & 5.892 & 5.334 & 5.116 & \textbf{4.993} \\
    OT &  \textbf{71.700} & 20.703 & 11.385 & 8.065 & 7.492 & 7.303 & 7.316\\
    \RowStyle[rowcolor=defaultColor]{}
    \ourmethod{} (ours) & 81.480 & \textbf{18.285} & \textbf{8.661} & \textbf{5.607} & \textbf{5.201} & \textbf{5.055} & 5.035 \\
    \midrule
    \bottomrule
\end{NiceTabular}

    \caption{Class-to-image performance comparisons on ImageNet-32.
    }
    \label{tab:class_to_image}
\end{table}

\subsection{Predicting Conditions from Coupled Prior}
We note that OT degrades less significantly in high-dim as it skews the prior less (but still does), since mini-batch OT becomes noisy in high-dim.
To quantify, we train a condition-classifier with the coupled prior (\(x_0\)) as input (\eg, in Fig.~2 of the main paper, a classifier can perfectly predict the condition based on \(x_0\)) on CIFAR with varying resolutions and number of dims.
We collect 100K couplings for training, and divide them into an 80:20 train/test split.
\cref{tab:noise_prediction} shows the results: as the number of dimensions increases, the classifier becomes less accurate (less prior skew) for OT. 
Both FM and \ourmethod{} lead to unbiased prior, so the classifier performs at random guess accuracy (10\%).

\begin{table}[h]
    \centering
    \begin{NiceTabular}{lccccc}
    \toprule
    Method & \(2\times2\) & \(4\times4\) & \(8\times8\) & \(16\times16\) & \(32\times32\) \\
    \midrule
    FM & 10.0\%\(\pm\)0.3 & 9.9\%\(\pm\)0.1 & 10.0\%\(\pm\)0.3 & 9.9\%\(\pm\)0.2 & 9.8\%\(\pm\)0.1 \\
    OT & 29.7\%\(\pm\)0.4 & 27.3\%\(\pm\)0.3 & 24.4\%\(\pm\)0.4 & 22.6\%\(\pm\)0.3 & 20.1\%\(\pm\)0.3 \\ 
    \RowStyle[rowcolor=defaultColor]{}
    \ourmethod{} (ours) & 10.0\%\(\pm\)0.3 & 10.0\%\(\pm\)0.3 & 10.0\%\(\pm\)0.3 & 9.9\%\(\pm\)0.2 & 9.9\%\(\pm\)0.2\\
    \midrule
    \bottomrule
\end{NiceTabular}

    \caption{Test-set classification accuracy of predicting the conditioning class when given the coupled input noise in the CIFAR-10 dataset.
    }
    \label{tab:noise_prediction}
\end{table}

\subsection{Extended Results with Different OT Batch Sizes}

We compare OT batch size scaling of OT and \ourmethod{} in \Cref{tab:ot_batch_size_scaling}.
The results are consistent with our findings in \Cref{sec:vary_ot_batch_size} -- OT has better FID with few steps but does not align well with the input condition (worse CE), and increasing OT batch size improves few-step performance and slightly harms many-step performance.

\begin{table}[h]
\scriptsize
    \centering
    \begin{NiceTabular}{l@{\hspace{1.1mm}}l@{\hspace{1.6mm}}r@{}l@{\hspace{1.6mm}}r@{}l@{\hspace{1.6mm}}r@{}l@{\hspace{1.6mm}}r@{}l@{\hspace{1.6mm}}r@{}l@{\hspace{1.6mm}}r@{}l@{\hspace{1.6mm}}r@{}l@{\hspace{1.6mm}}r@{}l}
\toprule
    \toprule
    & Method (OT batch size) & \multicolumn{2}{c}{Euler-2} & \multicolumn{2}{c}{Euler-5} & \multicolumn{2}{c}{Euler-10} & \multicolumn{2}{c}{Euler-25} & \multicolumn{2}{c}{Euler-50} & \multicolumn{2}{c}{Euler-100} & \multicolumn{2}{c}{Adaptive} & \multicolumn{2}{c}{NFE ↓} \\
    \midrule
    \parbox[t]{3mm}{\multirow{10}{*}{\rotatebox[origin=c]{90}{FID ↓}}} 
    & OT (128) & 64.305&\(\pm\)0.593 & 18.033&\(\pm\)0.566 & 10.659&\(\pm\)0.354 & 6.753&\(\pm\)0.26 & 5.326&\(\pm\)0.157 & 4.639&\(\pm\)0.084 & 4.142&\(\pm\)0.015 & 125.6&\(\pm\)0.5 \\
    & OT (640) & 59.506&\(\pm\)0.976 & 18.710&\(\pm\)0.673 & 11.588&\(\pm\)0.475 & 7.555&\(\pm\)0.301 & 5.997&\(\pm\)0.222 & 5.232&\(\pm\)0.175 & 4.624&\(\pm\)0.122 & 126.1&\(\pm\)1.1 \\
    & OT (1280) & 57.656&\(\pm\)1.075 & 18.206&\(\pm\)2.063 & 11.284&\(\pm\)1.707 & 7.452&\(\pm\)1.116 & 6.036&\(\pm\)0.696 & 5.374&\(\pm\)0.408 & 4.895&\(\pm\)0.044 & 125.1&\(\pm\)6.5 \\
    & OT (2560) & 56.186&\(\pm\)0.876 & 17.976&\(\pm\)1.752 & 11.084&\(\pm\)1.682 & 7.381&\(\pm\)1.078 & 6.068&\(\pm\)0.633 & 5.473&\(\pm\)0.325 & 5.071&\(\pm\)0.074 & 125.8&\(\pm\)7.1 \\
    & OT (5120) & 54.966 &\(\pm\)0.212 & 19.399&\(\pm\)0.194 & 12.563&\(\pm\)0.112 & 8.498&\(\pm\)0.056 & 6.869&\(\pm\)0.038 & 6.042&\(\pm\)0.042 & 5.352&\(\pm\)0.074 & 128.0&\(\pm\)3.0 \\
    & \ourmethod{} (128) & 74.517&\(\pm\)0.075 & 18.281&\(\pm\)0.443 & 9.704&\(\pm\)0.326 & 5.536&\(\pm\)0.193 & 4.077&\(\pm\)0.135 & 3.391&\(\pm\)0.101 & 2.880&\(\pm\)0.059 & 124.0&\(\pm\)2.5 \\
    & \ourmethod{} (640) & 64.849&\(\pm\)0.474 & 17.572&\(\pm\)0.263 & 9.770&\(\pm\)0.195 & 5.634&\(\pm\)0.112 & 4.150&\(\pm\)0.083 & 3.446&\(\pm\)0.068 & 2.905&\(\pm\)0.047 & 127.2&\(\pm\)0.3 \\
    & \ourmethod{} (1280) & 61.850&\(\pm\)0.373 & 17.325&\(\pm\)0.407 & 9.706&\(\pm\)0.321 & 5.668&\(\pm\)0.206 & 4.200&\(\pm\)0.147 & 3.495&\(\pm\)0.102 & 2.951&\(\pm\)0.048 & 124.1&\(\pm\)0.9 \\
    & \ourmethod{} (2560) & 59.361&\(\pm\)0.324 & 17.213&\(\pm\)0.297 & 9.795&\(\pm\)0.200 & 5.712&\(\pm\)0.149 & 4.213&\(\pm\)0.102 & 3.496&\(\pm\)0.068 & 2.942&\(\pm\)0.030 & 125.5&\(\pm\)0.7 \\
    & \ourmethod{} (5120) & 56.011&\(\pm\)0.556 & 17.046&\(\pm\)0.185 & 9.904&\(\pm\)0.205 & 5.830&\(\pm\)0.126 & 4.291&\(\pm\)0.079 & 3.536&\(\pm\)0.054 & 2.945&\(\pm\)0.022 & 128.7&\(\pm\)0.7 \\
    \midrule
    \parbox[t]{3mm}{\multirow{10}{*}{\rotatebox[origin=c]{90}{CE ↓}}} 
    & OT (128) & 2.525&\(\pm\)0.015 & 0.619&\(\pm\)0.016 & 0.425&\(\pm\)0.006 & 0.368&\(\pm\)0.004 & 0.361&\(\pm\)0.004 & 0.363&\(\pm\)0.006 & 0.371&\(\pm\)0.006 & 125.6&\(\pm\)0.5 \\
    & OT (640) & 2.386&\(\pm\)0.048 & 0.672&\(\pm\)0.018 & 0.470&\(\pm\)0.011 & 0.405&\(\pm\)0.008 & 0.393&\(\pm\)0.006 & 0.392&\(\pm\)0.007 & 0.398&\(\pm\)0.008 & 126.1&\(\pm\)1.1 \\
    & OT (1280) & 2.348&\(\pm\)0.057 & 0.690&\(\pm\)0.030 & 0.487&\(\pm\)0.015 & 0.427&\(\pm\)0.009 & 0.422&\(\pm\)0.005 & 0.425&\(\pm\)0.003 & 0.435&\(\pm\)0.007 & 125.1&\(\pm\)6.5 \\
    & OT (2560) & 2.298&\(\pm\)0.031 & 0.702&\(\pm\)0.020 & 0.505&\(\pm\)0.010 & 0.443&\(\pm\)0.006 & 0.436&\(\pm\)0.008 & 0.438&\(\pm\)0.011 & 0.447&\(\pm\)0.014 & 125.8&\(\pm\)7.1 \\
    & OT (5120) & 2.277&\(\pm\)0.029 & 0.742&\(\pm\)0.006 & 0.537&\(\pm\)0.012 & 0.464&\(\pm\)0.009 & 0.453&\(\pm\)0.008 & 0.453&\(\pm\)0.007 & 0.461&\(\pm\)0.006 & 128.0&\(\pm\)3.0 \\
    & \ourmethod{} (128) & 2.636&\(\pm\)0.033 & 0.485&\(\pm\)0.007 & 0.317&\(\pm\)0.004 & 0.270&\(\pm\)0.004 & 0.265&\(\pm\)0.001 & 0.266&\(\pm\)0.002 & 0.271&\(\pm\)0.004 & 124.0&\(\pm\)2.5 \\
    & \ourmethod{} (640) & 2.270&\(\pm\)0.021 & 0.477&\(\pm\)0.012 & 0.321&\(\pm\)0.005 & 0.276&\(\pm\)0.008 & 0.272&\(\pm\)0.004 & 0.273&\(\pm\)0.005 & 0.280&\(\pm\)0.005 & 127.2&\(\pm\)0.3 \\
    & \ourmethod{} (1280) & 2.166&\(\pm\)0.032 & 0.475&\(\pm\)0.010 & 0.324&\(\pm\)0.008 & 0.279&\(\pm\)0.003 & 0.273&\(\pm\)0.004 & 0.274&\(\pm\)0.003 & 0.281&\(\pm\)0.003 & 124.1&\(\pm\)0.9 \\
    & \ourmethod{} (2560) & 2.049&\(\pm\)0.005 & 0.468&\(\pm\)0.005 & 0.321&\(\pm\)0.003 & 0.278&\(\pm\)0.006 & 0.272&\(\pm\)0.007 & 0.272&\(\pm\)0.007 & 0.277&\(\pm\)0.008 & 125.5&\(\pm\)0.7 \\
    & \ourmethod{} (5120) & 1.910&\(\pm\)0.037 & 0.463&\(\pm\)0.006 & 0.327&\(\pm\)0.008 & 0.282&\(\pm\)0.005 & 0.278&\(\pm\)0.005 & 0.279&\(\pm\)0.004 & 0.284&\(\pm\)0.004 & 128.7&\(\pm\)0.7 \\
    \midrule
    \bottomrule
\end{NiceTabular}

    \caption{Performance of OT and \ourmethod{} in CIFAR-10 when trained with different OT batch sizes. 
    }
    \label{tab:ot_batch_size_scaling}
\end{table}

\subsection{Extended Results with Different Target Ratios}

We list additional results on ImageNet-32 and ImageNet-256 in \Cref{tab:vary_r_tar}:
\(r_{\text{tar}}=0.01\) generally strikes a good balance, but we note that it might not be optimal for all datasets.

\begin{table}[h]
    \centering
    \begin{NiceTabular}{l@{\hspace{6pt}}c@{\hspace{6pt}}c@{\hspace{6pt}}c@{\hspace{6pt}}c@{\hspace{6pt}}c@{\hspace{6pt}}c@{\hspace{6pt}}c}
    \toprule
    \multicolumn{8}{c}{\textbf{ImageNet 32\(\times\)32 Caption-Conditioned Generation}}\\
    \toprule
    \(r_{\text{tar}}\) & Euler-2 & Euler-5 & Euler-10 & Euler-25 & Euler-50 & Euler-100 & Adaptive \\
    \midrule
    0.005 & 104.386 & 22.254 & 10.939 & \textbf{7.015} & \textbf{6.022} & \textbf{5.576 }& \textbf{5.284}  \\
    0.01 & 102.380 & 21.965 & \textbf{10.897} & 7.069 & 6.084 & 5.638 & 5.350 \\
    0.1 & \textbf{82.456} & \textbf{20.820} & 11.035 & 7.438 & 6.501 & 6.080 & 5.843 \\
    \toprule
    \multicolumn{8}{c}{\textbf{ImageNet 256\(\times\)256 Caption-Conditioned Generation}}\\
    \midrule
    0.005 & 201.906 & \textbf{30.159} & \textbf{9.852} & 5.114 & 3.795 & 3.373 & 3.377 \\
    0.01 & 201.010 & 30.578 & 10.032 & \textbf{5.075} & \textbf{3.702} & \textbf{3.335} & \textbf{3.290} \\
    0.1 & \textbf{199.009} & 37.014 & 13.293 & 6.277 & 4.608 & 4.268 & 4.570 \\
    \midrule
    \bottomrule
\end{NiceTabular}

    \caption{Results with different \(\targetratio\).}
    \label{tab:vary_r_tar}
\end{table}

\subsection{Reference Condition Adherence Metrics}
We measure condition adherence in \Cref{sec:expr} via two condition adherence metrics.
On CIFAR-10, we compute the average cross-entropy of the logits predicted by a pretrained classifier\footnote{\url{https://github.com/chenyaofo/pytorch-cifar-models}, commit d1c8e99b911da7d412979600c84d2a4fe3728473, ResNet56} on the generated images against the ground-truth conditioning labels.
On ImageNet, we compute the average cosine distance between the CLIP embeddings extracted from the generated images versus the conditioning captions, using SigLip-2~\cite{tschannen2025siglip}\footnote{ViT-SO400M-16-SigLIP2-256}.
For reference, we compute these metrics on the validation set using ground-truth images and present the results in \Cref{tab:reference_adherence}.

\begin{table}[h]
    \centering
    \begin{NiceTabular}{lc}
    \toprule
    Dataset & CE (↓) \\
    \midrule
    CIFAR-10 & 0.0005 \\
    \toprule
    Dataset & CLIP (↑) \\
    \midrule
    ImageNet 32\(\times\)32 & 0.1119 \\
    ImageNet 256\(\times\)256 & 0.1363 \\
    \midrule
    \bottomrule
    \end{NiceTabular}
    \caption{Reference condition adherence metrics on ground-truth images.}
    \label{tab:reference_adherence}
\end{table}

\section{Extended Plots}
\label{sec:app:extended_plots}

\Cref{fig:app:ot_batch_size_ext,fig:app:target_r_ext} extend Figures 5 and 6 of the main paper to include all data points. 

\begin{figure}[h]
    \centering
    \begin{minipage}{0.48\linewidth}
    \captionsetup{type=figure}
    \begin{adjustbox}{width=\linewidth}
    \begin{tikzpicture}
    \pgfplotsset{
        log ticks with fixed point,
        xmin=128, xmax=6400,
        xtick = {
            128, 640, 1280, 2560, 6400
        },
    }
    
    \begin{axis}[
      scale only axis,
      axis y line*=left,
      xlabel={OT batch size \(\batchsize\)},
      ylabel={\textcolor{red}{Euler-2 FID ↓}},
      enlarge x limits=0.05,   %
      grid=both,               %
      ymin=50, ymax=110,       %
      restrict y to domain=50:110,
      every axis plot/.append style={thick},
      xmode=log,
      minor tick num=4,
      every y tick label/.append style={red},
      ytick={50,55,60,65,70,75,80,85,90,95,100,105,110},
      width=9cm,height=10cm,
    ]

    \addplot[name path=upper1, draw=none] coordinates{
        (128,74.5172 + 0.0614)
        (640,64.8487 + 0.3867)
        (1280,61.8505 + 0.3042)
        (2560,59.3612 + 0.2646)
        (6400,56.0108 + 0.4537)
    };

    \addplot[name path=lower1, draw=none] coordinates{
        (128,74.5172 - 0.0614)
        (640,64.8487 - 0.3867)
        (1280,61.8505 - 0.3042)
        (2560,59.3612 - 0.2646)
        (6400,56.0108 - 0.4537)
    };
    
    \addplot[mark=x,red]
      coordinates{
        (128,74.5172)
        (640,64.8487)
        (1280,61.8505)
        (2560,59.3612)
        (6400,56.0108)
    }; \label{euler2_ext}

    \addplot[blue, fill=red, fill opacity=0.2] 
      fill between[of=upper1 and lower1];

    \addplot[only marks, mark=*, mark options={fill=red}] coordinates {(128,64.694)};
    \node[above] at (axis cs:180,64.694) {\small OT (Euler-2)};
    
    \addplot [domain=128:6400, samples=2, dashed, color=blue] {105.4807};
    \node[above] at (axis cs:4500,105.4807) {\small FM (Euler-2)};
        
    \end{axis}
    
    \begin{axis}[
      scale only axis,
      at={(current axis.south west)},
      anchor=south west,
      legend style={at={(0.88,0.87)}},
      axis y line*=right,
      axis x line=none,
      ylabel={\textcolor{blue}{Adaptive FID ↓}},
      grid=both,              %
      enlarge x limits=0.05,   %
      ymin=2.85, ymax=4.85,      %
      restrict y to domain=2.8:4.8,
      grid=none,
      tick label style={
                /pgf/number format/fixed,
                /pgf/number format/fixed zerofill,
                /pgf/number format/precision=2
            },
        ytick={2.85,3.02,3.18,3.35,3.52,3.68,3.85,4.02,4.18,4.35,4.52,4.68,4.85},
        every axis plot/.append style={thick},
        xmode=log,
        every y tick label/.append style={blue},
        width=9cm,height=10cm,
    ]
    \addlegendimage{/pgfplots/refstyle=euler2_ext}\addlegendentry{Ours, Euler-2}
    \addlegendimage{/pgfplots/refstyle=adaptive_ext}\addlegendentry{Ours, adaptive}

    \addplot[name path=upper2, draw=none] coordinates{
        (128,2.8804 + 0.0481)
        (640,2.9051 + 0.0381)
        (1280,2.9511 + 0.0386)
        (2560,2.9413 + 0.0243)
        (6400,2.9449 + 0.0174)
    };

    \addplot[name path=lower2, draw=none] coordinates{
        (128,2.8804 - 0.0481)
        (640,2.9051 - 0.0381)
        (1280,2.9511 - 0.0386)
        (2560,2.9413 - 0.0243)
        (6400,2.9449 - 0.0174)
    };
    
    \addplot[mark=o,blue]
      coordinates{
        (128,2.8804)
        (640,2.9051)
        (1280,2.9511)
        (2560,2.9413)
        (6400,2.9449)
    }; \label{adaptive_ext}
    \addplot[blue, fill=blue, fill opacity=0.2] 
      fill between[of=upper2 and lower2];

    \addplot [domain=128:6400, samples=2, dashed, color=blue] {2.9224};
    \node[below] at (axis cs:4500,2.9324) {\small FM (adaptive)};

    \addplot[only marks, mark=*, mark options={fill=blue}] coordinates {(128,4.6394)};
    \node[below] at (axis cs:180,4.6394) {\small OT (adaptive)};
      
    \end{axis}
    \end{tikzpicture}
\end{adjustbox}
    \vspace{-1em}
    \captionof{figure}{
    Changes in FID with respect to varying OT batch sizes \(\batchsize\). 
    We plot mean\(\pm\)std over three runs and represent std with a shaded region.
    }
    \label{fig:app:ot_batch_size_ext}
    \end{minipage}
    \hspace{1em}
    \begin{minipage}{0.48\linewidth}
    \captionsetup{type=figure}
    \begin{adjustbox}{width=\linewidth}
    \begin{tikzpicture}
    \pgfplotsset{
        xmode=log,
        log ticks with fixed point,
        xmin=0.0005, xmax=0.1,
        xtick = {
            0.0005, 0.001, 0.002, 0.005, 0.01, 0.02, 0.05, 0.1
        },
    }
    
    \begin{axis}[
      scale only axis,
      axis y line*=left,
      xlabel={Target ratio \(\targetratio\)},
      ylabel={\textcolor{red}{Euler-2 FID ↓}},
      enlarge x limits=0.05,   %
      grid=both,               %
      ymin=75, ymax=115,       %
      restrict y to domain=80:115,
      every axis plot/.append style={thick},
      xmode=log,
      minor tick num=4,
      every y tick label/.append style={red},
      ytick={75,80,85,90,95,100,105,110,115},
      width=9cm,height=10cm,
    ]

    \addplot[name path=upper3, draw=none] coordinates{
        (0.0005,110.148 + 0.820)
        (0.001,109.555 + 0.579)
        (0.002,107.825 + 0.588)
        (0.005,105.148 + 0.832)
        (0.01,102.380 + 0.279)
        (0.02,97.667 + 0.639)
        (0.05,90.114 + 0.181)
        (0.1,82.456 + 0.343)
    };

    \addplot[name path=lower3, draw=none] coordinates{
        (0.0005,110.148 - 0.820)
        (0.001,109.555 - 0.579)
        (0.002,107.825 - 0.588)
        (0.005,105.148 - 0.832)
        (0.01,102.380 - 0.279)
        (0.02,97.667 - 0.639)
        (0.05,90.114 - 0.181)
        (0.1,82.456 - 0.343)
    };
    
    \addplot[mark=x,red]
      coordinates{
        (0.0005,110.148)
        (0.001,109.555)
        (0.002,107.825)
        (0.005,105.148)
        (0.01,102.380)
        (0.02,97.667)
        (0.05,90.114)
        (0.1,82.456)
    }; \label{r_euler2_ext}

    \addplot[blue, fill=red, fill opacity=0.2] 
      fill between[of=upper3 and lower3];

    \addplot [domain=0.0005:0.1, samples=2, dashed, color=red] {81.3169};
    \node[below, anchor=north west] at (axis cs:0.02,81.3169) {\small OT (Euler-2)};

    \addplot [domain=0.0005:0.1, samples=2, dashed, color=red] {113.250};
    \node[above, anchor=south west] at (axis cs:0.0005,113.250) {\small FM (Euler-2)};
    
    \end{axis}
    
    \begin{axis}[
      scale only axis,
      at={(current axis.south west)},
      anchor=south west,
      legend style={at={(0.88,0.87)}},
      axis y line*=right,
      axis x line=none,
      ylabel={\textcolor{blue}{Adaptive FID ↓}},
      every tick label/.append style={blue},
      grid=both,              %
        enlarge x limits=0.05,   %
      ymin=5.15, ymax=6.3,      %
      restrict y to domain=5.15:6.3,
      grid=none,
      tick label style={
                /pgf/number format/fixed,
                /pgf/number format/fixed zerofill,
                /pgf/number format/precision=2
            },
        ytick={5.15,5.29,5.44,5.58,5.73,5.87,6.01,6.16,6.3},
        every axis plot/.append style={thick},
        xmode=log,
        width=9cm,height=10cm,
    ]
    \addlegendimage{/pgfplots/refstyle=r_euler2_ext}\addlegendentry{Ours, Euler-2}
    \addlegendimage{/pgfplots/refstyle=r_adaptive_ext}\addlegendentry{Ours, adaptive}

    \addplot[name path=upper4, draw=none] coordinates{
        (0.0005,5.341 + 0.015)
        (0.001,5.334 + 0.002)
        (0.002,5.338 + 0.027)
        (0.005,5.316 + 0.044)
        (0.01,5.350 + 0.017)
        (0.02,5.425 + 0.016)
        (0.05,5.579 + 0.038)
        (0.1,5.843 + 0.054)
    };

    \addplot[name path=lower4, draw=none] coordinates{
        (0.0005,5.341 - 0.015)
        (0.001,5.334 - 0.002)
        (0.002,5.338 - 0.027)
        (0.005,5.316 - 0.044)
        (0.01,5.350 - 0.017)
        (0.02,5.425 - 0.016)
        (0.05,5.579 - 0.038)
        (0.1,5.843 - 0.054)
    };
    
    \addplot[mark=o,blue]
      coordinates{
        (0.0005,5.341)
        (0.001,5.334)
        (0.002,5.338)
        (0.005,5.316)
        (0.01,5.350)
        (0.02,5.425)
        (0.05,5.579)
        (0.1,5.843)
    }; \label{r_adaptive_ext}
    \addplot[blue, fill=blue, fill opacity=0.2] 
      fill between[of=upper4 and lower4];

    \addplot [domain=0.0005:0.1, samples=2, dashed, color=blue] {5.358};
    \node[above, anchor=south west] at (axis cs:0.0005,5.358) {\small FM (adaptive)};

    \addplot [domain=0.0005:0.1, samples=2, dashed, color=blue] {6.2205};
    \node[below, anchor=north west] at (axis cs:0.0005,6.2205) {\small OT (adaptive)};
      
    \end{axis}
    \end{tikzpicture}
\end{adjustbox}
    \vspace{-1em}
    \captionof{figure}{
    Changes in FID with respect to varying target ratio \(\targetratio\). 
    We plot mean\(\pm\)std over three runs and represent std with a shaded region.
    }
    \label{fig:app:target_r_ext}
    \end{minipage}
\end{figure}

\section{Data Coupling in 8 Gaussians→moons}

\begin{figure}[t]
    \centering
    \begin{NiceTabular}{c@{}c@{}c@{}|@{}c@{}c@{}c@{}|@{}c@{}c@{}c}
& \Block{1-2}{\small Unconditonal} & & \Block{1-3}{\small With discrete conditions} & & & \Block{1-3}{\small With continuous conditions \\ \small (target \(x\) coordinates)}
\\
\cmidrule(lr{\dimexpr 4\tabcolsep-16pt}){2-3}
\cmidrule(lr{\dimexpr 4\tabcolsep-16pt}){4-6}
\cmidrule(lr{\dimexpr 4\tabcolsep-16pt}){7-9}
\Block{1-1}{\rotate \small \hspace{1.5em}Training}%
&
\includegraphics[width=0.118\linewidth]{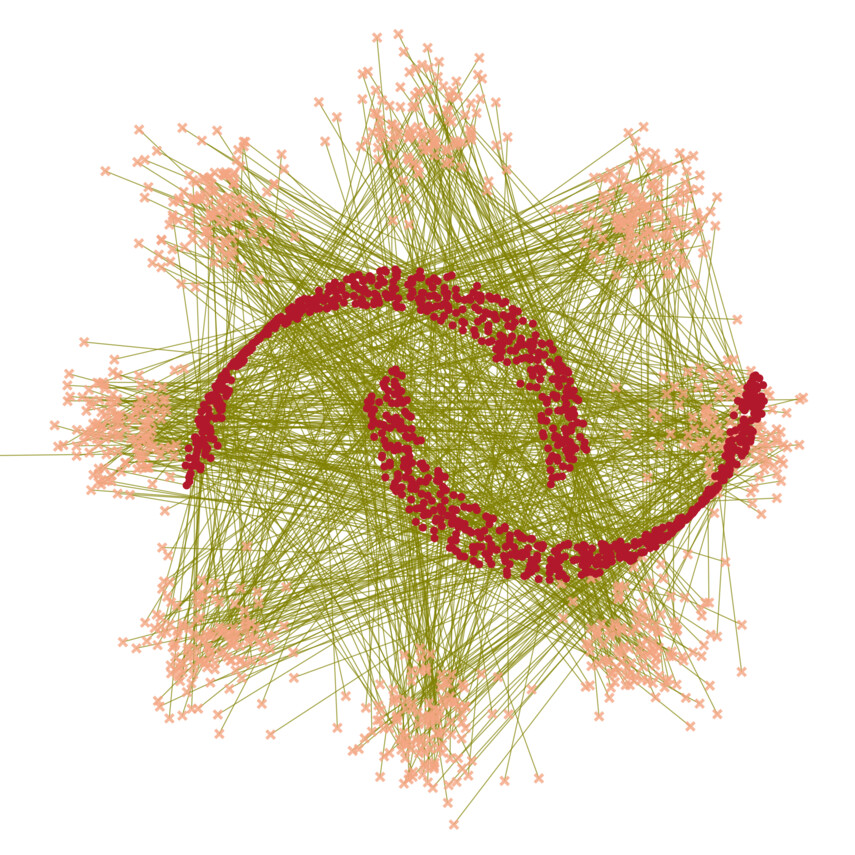} & 
\includegraphics[width=0.118\linewidth]{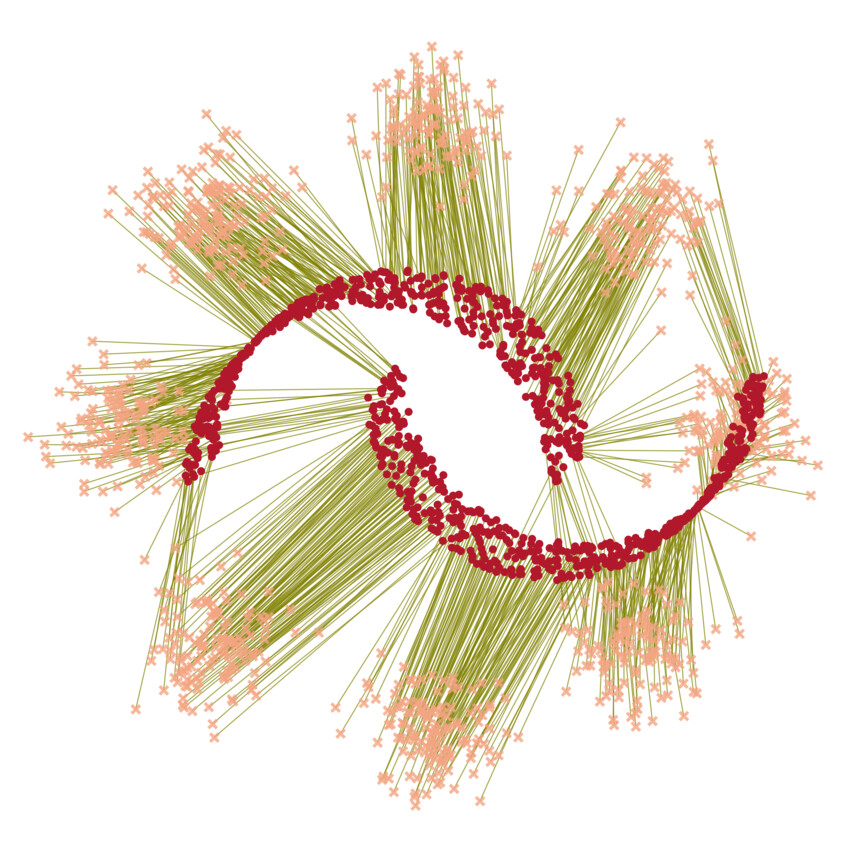} & 
\includegraphics[width=0.118\linewidth]{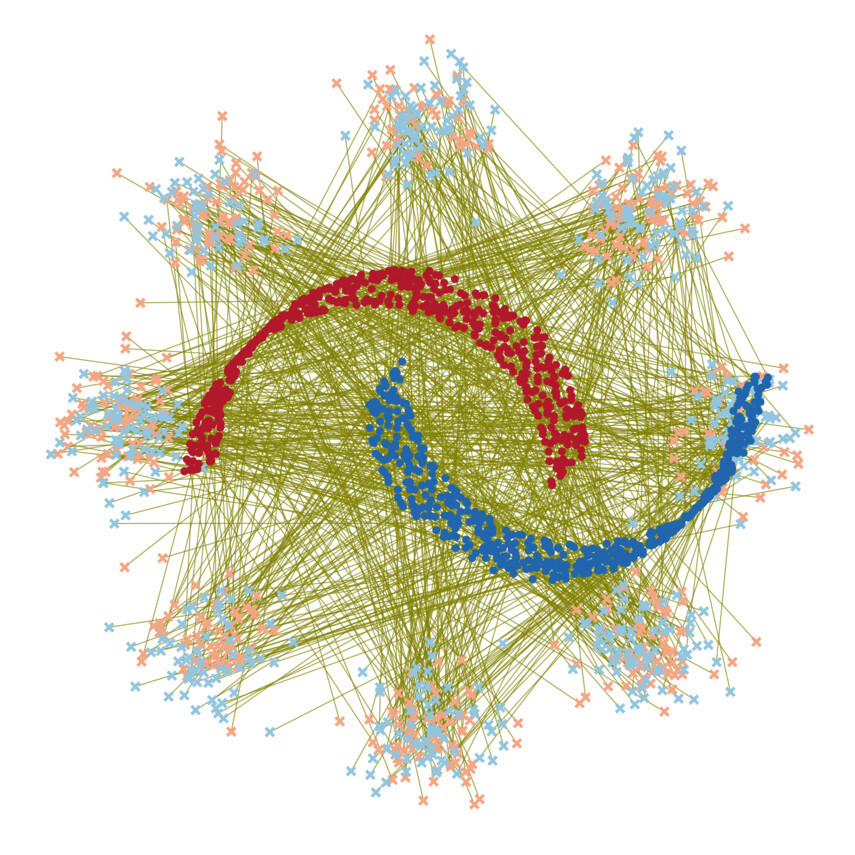} & 
\includegraphics[width=0.118\linewidth]{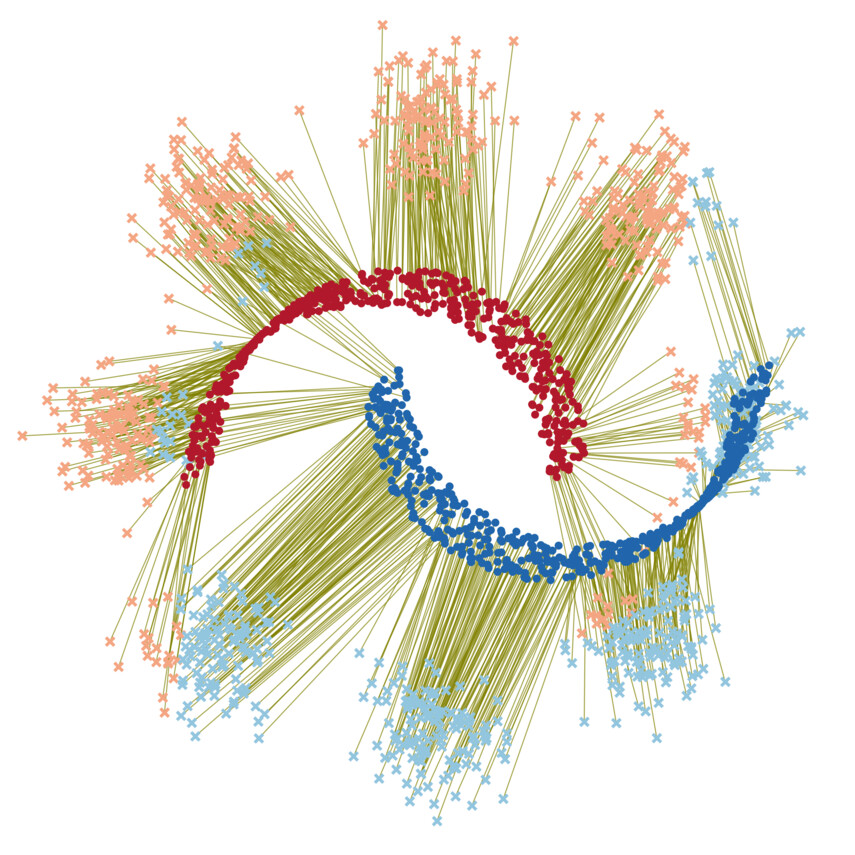} & 
\includegraphics[width=0.118\linewidth]{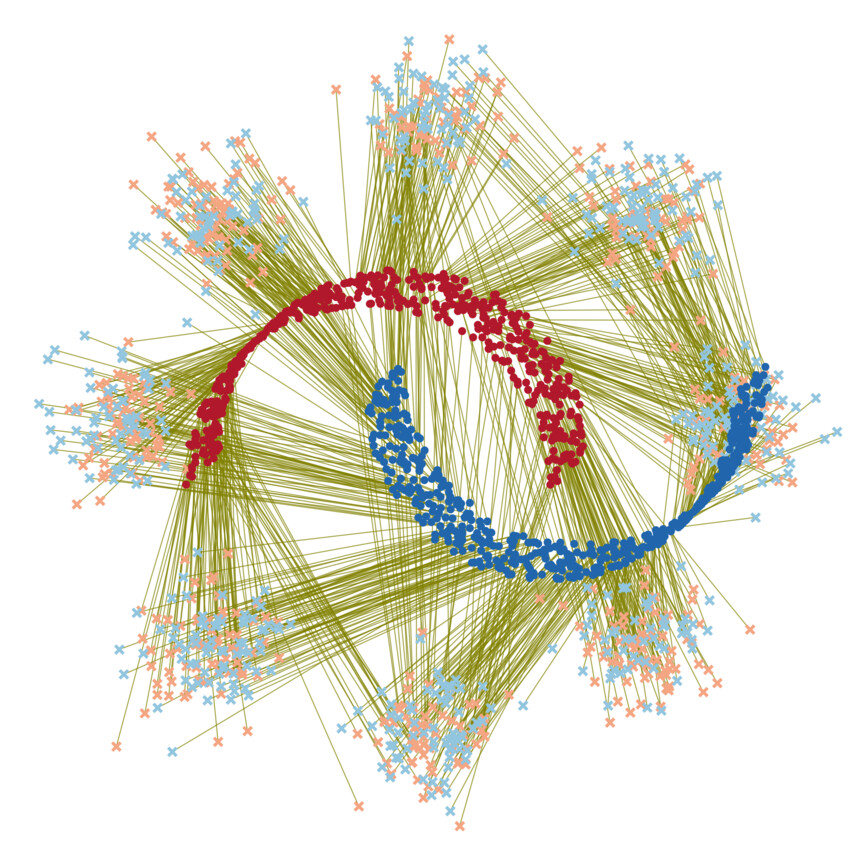} & 
\includegraphics[width=0.118\linewidth]{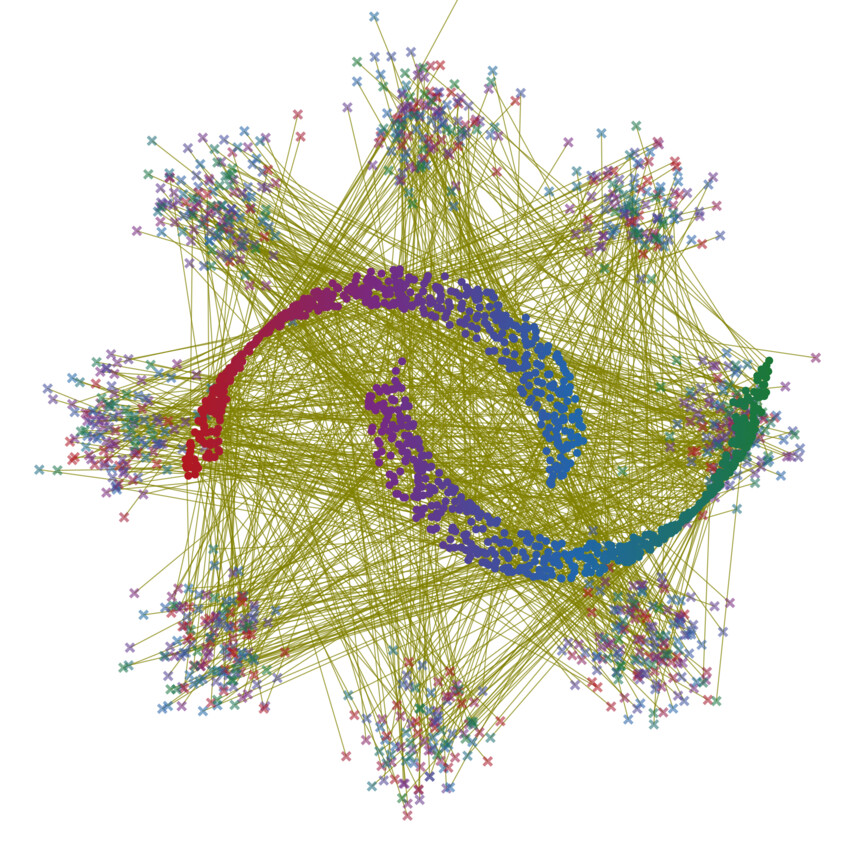} & 
\includegraphics[width=0.118\linewidth]{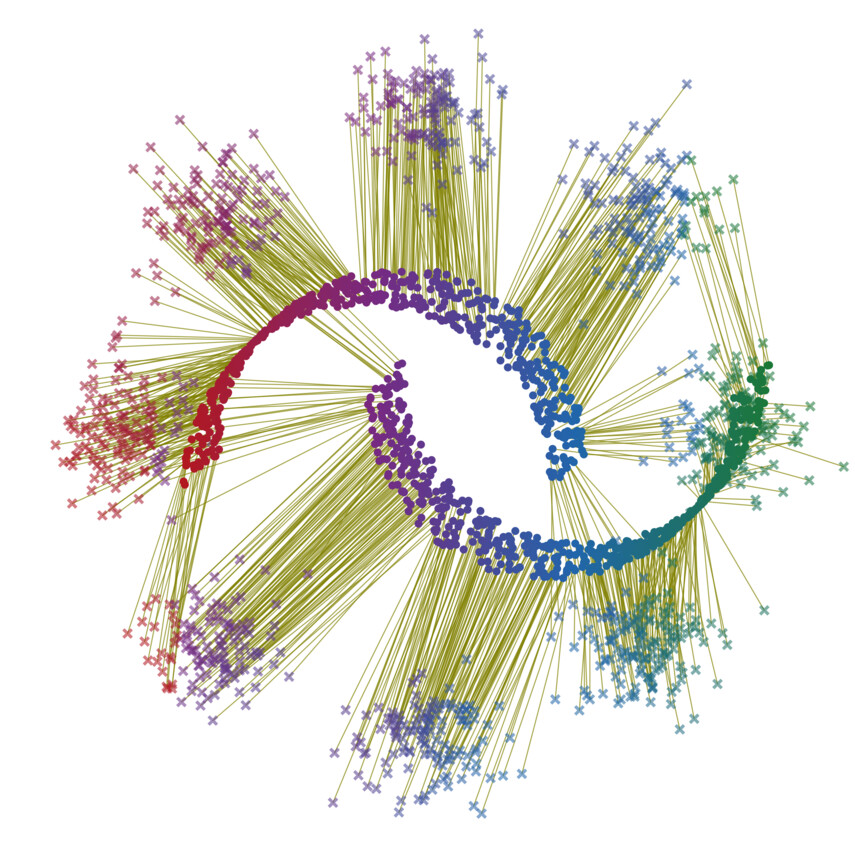} & 
\includegraphics[width=0.118\linewidth]{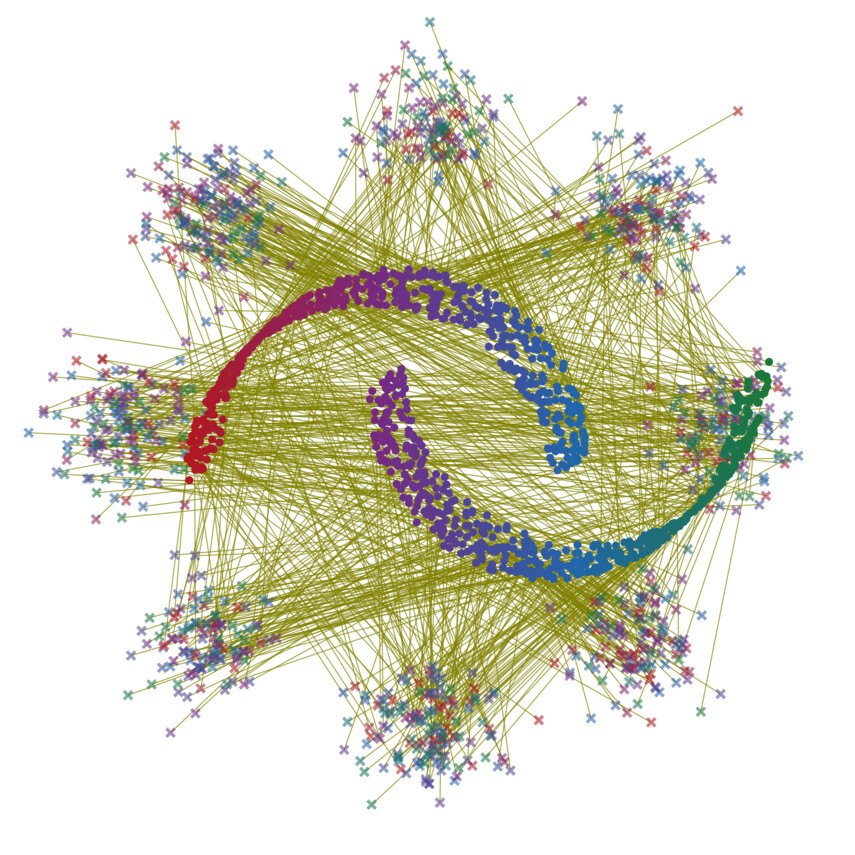}
\\
\Block{2-1}{\rotate \small Euler-1 step}
&
\includegraphics[width=0.118\linewidth]{fig/moons/moons-fm-unc_traj_euler1_20000.jpg} & 
\includegraphics[width=0.118\linewidth]{fig/moons/moons-ot-unc_traj_euler1_20000.jpg} & 
\includegraphics[width=0.118\linewidth]{fig/moons/moons-fm-cls_traj_euler1_20000.jpg} & 
\includegraphics[width=0.118\linewidth]{fig/moons/moons-ot-cls_traj_euler1_20000.jpg} & 
\includegraphics[width=0.118\linewidth]{fig/moons/moons-ot-cls-cls_traj_euler1_20000.jpg} & 
\includegraphics[width=0.118\linewidth]{fig/moons/moons-fm-x_traj_euler1_20000.jpg} & 
\includegraphics[width=0.118\linewidth]{fig/moons/moons-ot-x_traj_euler1_20000.jpg} & 
\includegraphics[width=0.118\linewidth]{fig/moons/moons-ot-cls-x_traj_euler1_20000.jpg}
\\
& \small 6.232\(\pm\)0.186
& \small \textbf{0.072\(\pm\)0.025}
& \small 2.573\(\pm\)0.112
& \small 0.483\(\pm\)0.059
& \small \textbf{0.048\(\pm\)0.010}
& \small 0.732\(\pm\)0.052
& \small 8.276\(\pm\)6.510
& \small \textbf{0.077\(\pm\)0.024}
\\
\Block{2-1}{\rotate \small Adaptive}
&
\includegraphics[width=0.118\linewidth]{fig/moons/moons-fm-unc_traj_20000.jpg} & 
\includegraphics[width=0.118\linewidth]{fig/moons/moons-ot-unc_traj_20000.jpg} & 
\includegraphics[width=0.118\linewidth]{fig/moons/moons-fm-cls_traj_20000.jpg} & 
\includegraphics[width=0.118\linewidth]{fig/moons/moons-ot-cls_traj_20000.jpg} & 
\includegraphics[width=0.118\linewidth]{fig/moons/moons-ot-cls-cls_traj_20000.jpg} & 
\includegraphics[width=0.118\linewidth]{fig/moons/moons-fm-x_traj_20000.jpg} & 
\includegraphics[width=0.118\linewidth]{fig/moons/moons-ot-x_traj_20000.jpg} & 
\includegraphics[width=0.118\linewidth]{fig/moons/moons-ot-cls-x_traj_20000.jpg}
\\
& \small 0.120\(\pm\)0.045 
& \small \textbf{0.050\(\pm\)0.040} 
& \small 0.059\(\pm\)0.029 
& \small 0.462\(\pm\)0.025 
& \small \textbf{0.018\(\pm\)0.006} 
& \small 0.028\(\pm\)0.010 
& \small 2.143\(\pm\)1.993 
& \small \textbf{0.013\(\pm\)0.003} 
\\
& \small FM
& \small OT
& \small FM
& \small OT
& \small C\(^2\)OT (ours)
& \small FM
& \small OT
& \small C\(^2\)OT (ours)
\end{NiceTabular}

    \caption{
    We visualize the flows learned by different algorithms using the {\tt 8gaussians\(\to\)moons} dataset.
    Below each plot, we show the 2-Wasserstein distance (lower is better; mean\(\pm\)std over 10 runs).
    Compared to Figure 1, we have added a first row illustrating the prior-data coupling during training.
    Note that the OT coupled paths during training (first row) do cross. 
    This is expected -- the commonly referred to ``no-crossing'' property of OT coupling refers to the uniqueness of the pair \((x_0, x_1)\) given \(x\) and \(t\) -- at the same timestep \(t\), no two paths may cross at \(x\)
    (see Proposition 3.4 in \citet{tong2023improving}
     and Theorem D.2 in \citet{pooladian2023multisample}).
    Since we plot all timesteps simultaneously in this figure, there are apparent crossings.
    However, the intersecting paths do not share the same timestep \(t\) at the point of intersection.
    }
    \label{fig:app:teaser-ext}
\end{figure}

\Cref{fig:app:teaser-ext} extends Figure 1 with an additional row that shows coupling during training.
Clearly, OT samples form a biased distribution at training time in conditional generation as discussed.
Since we cannot sample from this biased distribution at test-time, we obtain a gap between training and testing.
This gap degrades the performance of OT.

\section{Implementation Details}\label{sec:app:implementation}

\subsection{Two-Dimensional Data}

\paragraph{Data.}
Following the implementation of~\citet{tong2023improving}, we generate the ``moons'' data using the {\tt torchdyn} library~\cite{politorchdyn}, and the ``8 Gaussians'' using {\tt torchcfm}~\cite{tong2023improving}.

\paragraph{Network.}
We employ a simple multi-layer perceptron (MLP) network for this dataset.
Initially, the two-dimensional input (x and y coordinates) and the flow timestep (a scalar uniformly sampled from \([0,1]\)) are projected into the hidden dimension using individual linear layers. 
When an input condition is provided, it is similarly projected into the hidden dimension. 
Discrete conditions are encoded as -1 or +1, while continuous conditions are represented by the x-coordinate of the target data point. 
After projection, all input features are summed and processed through a network comprising three MLP modules. 
Each MLP module consists of two linear layers, where the first layer uses an expansion ratio of 4, and is followed by a GELU activation function~\cite{hendrycks2016gaussian}. 
We use a residual connection to incorporate the output of each MLP block.
Finally, another linear layer projects the features to two dimensions to produce the output velocity.
The hidden dimension is set to 128.

\paragraph{Training.}
We train each network for 20,000 iterations with the Adam~\cite{kingma2014adam} optimizer, a learning rate of \(3\text{e-}4\) without weight decay, and a ``deep net'' batch size of 256 for computing forward/backward passes/gradient updates.
We use an OT batch size \(\batchsize\) of 1024 and a target ratio \(\targetratio\) of 0.01.

\begin{table}[t]
    \centering
    \begin{tabular}{lcc}
    \toprule
    & CIFAR-10 & ImageNet-32  \\
    \midrule
    Channels & 128 & 256 \\
    Depth & 2 & 3 \\
    Channels multiple & 1, 2, 2, 2 & 1, 2, 2, 2 \\
    Heads & 4 & 4 \\
    Heads channels & 64 & 64 \\
    Attention resolution & 16 & 4\\
    Dropout & 0.0 & 0.0 \\
    Use scale shift norm & True & True \\
    Batch size / GPU & 128 & 128 \\
    GPUs & 2 & 4 \\
    Effective batch size & 256 & 512 \\
    Iterations & 100k & 300k \\
    Learning rate & 2.0e\(-\)4 & 1.0e\(-\)4 \\
    Learning rate scheduler & Warmup then constant & Warmup then linear decay \\
    Warmup steps & 5k & 20k \\
    OT batch size (per GPU, for \ourmethod{}) & 640 & 6400 \\
    \midrule
    \bottomrule
    \end{tabular}
    \caption{Hyperparameter settings for training on CIFAR-10 and ImageNet-32.}
    \label{tab:app:cifar_imagenet_hyperparameter}
\end{table}

\subsection{CIFAR-10}
\label{sec:app:cifar}

In CIFAR-10, we employ the UNet architecture used by~\citet{tong2023improving}.
We list our hyperparameters in \Cref{tab:app:cifar_imagenet_hyperparameter}, following the format in~\cite{tong2023improving,pooladian2023multisample,lipman2022flow}.
To accelerate training, we use {\tt bf16}.
We use the Adam~\cite{kingma2014adam} optimizer with the following parameters: \(\beta_1=0.9\), \(\beta_2=0.95\), weight decay=0.0, and \(\epsilon=1\textrm{e}{-8}\).
For learning rate scheduling, we linearly increase the learning rate from 1.0e\(-\)8 to 2.0e\(-\)4 over 5,000 iterations and then keep the learning rate constant.
The ``use scale shift norm'' denotes employing adaptive layer normalization to incorporate the input condition, as implemented in~\cite{dhariwal2021diffusion}.
To stabilize training, we clip the gradient norm to 1.0.
We report the results using an exponential moving average (EMA) model with a decay factor of 0.9999.
For FID computation, we use the {\tt clean\_fid} library~\cite{parmar2022aliased} in {\tt legacy\_tensorflow} mode following~\cite{tong2023improving}.

\subsection{ImageNet-32}
\label{sec:app:imagenet32}

\paragraph{Data.}
We use face-blurred ImageNet-1K~\cite{deng2009imagenet,yang2022study} following~\cite{lipman2024flow}, and apply the downsampling script from~\cite{chrabaszcz2017downsampled}.
Images are downsampled to 32××32 using the `box' algorithm, and reference FID statistics are computed with respect to the downsampled validation set images. 
For text input, we use the captions provided by \cite{imagenetcaptioned}.

\paragraph{Network and Training.}
We largely follow the training pipeline described in \Cref{sec:app:cifar}.
We use a larger network following~\cite{pooladian2023multisample} and list our hyperparameters in \Cref{tab:app:cifar_imagenet_hyperparameter}.
To encode text input, we use the {\tt openclip} library~\cite{ilharco_gabriel_2021_5143773} and the text encoder of the ``DFN5B-CLIP-ViT-H-14'' checkpoint~\cite{fang2023data}, a pretrained CLIP-like model.
CLIP feature vectors are normalized to unit norm before being used as input conditions.
For learning rate scheduling, after the initial warmup phase, we linearly decay the learning rate to 1.0e\(-\)8 over time.

\paragraph{Evaluation.}
As stated in the main paper, we use 49,997 images from the validation set to compute FID.
This is because the fine-grained nature of image captions might lead to overfitting, \ie, memorizing the training set.
For CLIP score computation, we evaluate the cosine similarity between the input caption and the generated image using SigLIP-2~\cite{tschannen2025siglip}, with the {\tt ViT-SO400M-16-SigLIP2-256} checkpoint via the {\tt openclip} library~\cite{ilharco_gabriel_2021_5143773}.

\subsection{ImageNet-256}
For this dataset, we use the open-source implementation of LightningDiT~\cite{yao2025reconstruction} and train the models under the `64 epochs' setting with minimal modifications to change the network from class-conditioned to caption-conditioned.
In addition to integrating the coupling algorithms (OT and \ourmethod{}, while the original LightningDiT~\cite{yao2025reconstruction} already employs FM), our modifications include:
\begin{enumerate}
    \item Changing the input conditional mapping layer from an embedding layer (that takes a class label as input) to a linear layer (that takes CLIP features as input).
    \item Adjusting the classifier-free guidance (CFG) scale. We find that the model benefits from a higher CFG scale when using caption conditioning. 
    Specifically, we increase the CFG scale from 10.0 to 17.0, and adjust the CFG interval start parameter from 0.11 to 0.10.
\end{enumerate}

For data and evaluation, we follow the same setup as described in \Cref{sec:app:imagenet32}.

\section{Additional Generated Images}

We present additional image generation results in this section. All showcased images are uncurated, meaning they were sampled completely at random. For consistency and direct comparison, we used the same random seed for each generation across different methods.

\clearpage
\subsection{CIFAR-10}
\begin{figure}[ht]
    \centering
    \begin{NiceTabular}{c@{}cc}
    & Euler-10 & Adaptive \\
    \Block{1-1}{\rotate \small \hspace{0.16\linewidth}FM}
    &
    \includegraphics[width=0.35\linewidth]{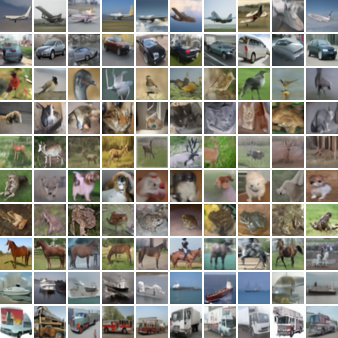} & 
    \includegraphics[width=0.35\linewidth]{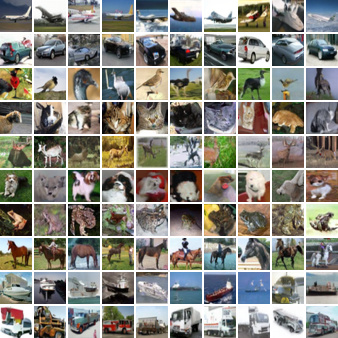} \\ \\
    \Block{1-1}{\rotate \small \hspace{0.16\linewidth}OT}
    &
    \includegraphics[width=0.35\linewidth]{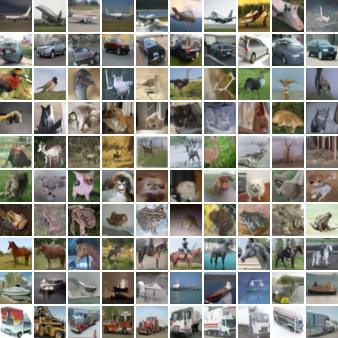} & 
    \includegraphics[width=0.35\linewidth]{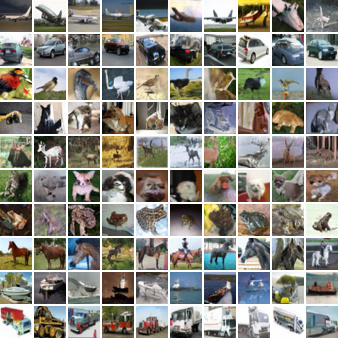} \\ \\
    \Block{1-1}{\rotate \small \hspace{0.16\linewidth}\ourmethod{} (ours)}
    &
    \includegraphics[width=0.35\linewidth]{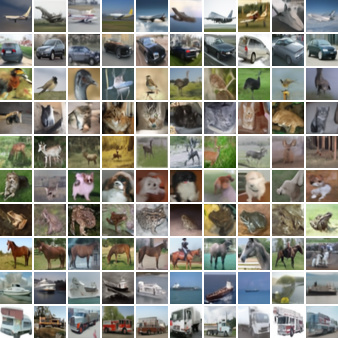} & 
    \includegraphics[width=0.35\linewidth]{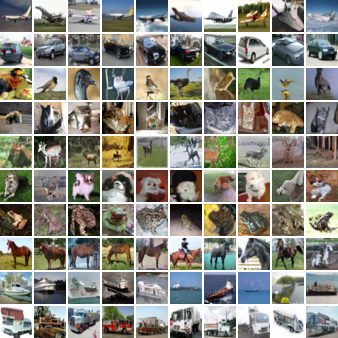} \\
    \end{NiceTabular}
    \caption{\emph{Uncurated} generations in CIFAR-10, 10-per-class.
    We compare FM, OT, and \ourmethod{} with both 10-step Euler's method and an adaptive solver for test-time numerical integration.
    }
    \label{fig:app:vis-cifar}
\end{figure}

\clearpage

\subsection{ImageNet-32}
\begin{figure}[ht]
    \centering
    \begin{NiceTabular}{c@{}cc}
    & Euler-10 & Adaptive \\
    \Block{1-1}{\rotate \small \hspace{0.16\linewidth}FM}
    &
    \includegraphics[width=0.35\linewidth]{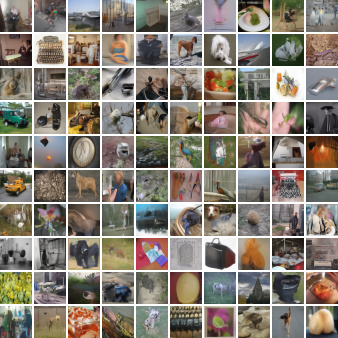} & 
    \includegraphics[width=0.35\linewidth]{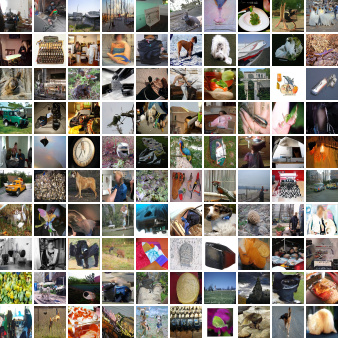} \\ \\
    \Block{1-1}{\rotate \small \hspace{0.16\linewidth}OT}
    &
    \includegraphics[width=0.35\linewidth]{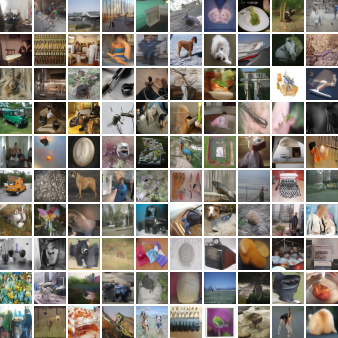} & 
    \includegraphics[width=0.35\linewidth]{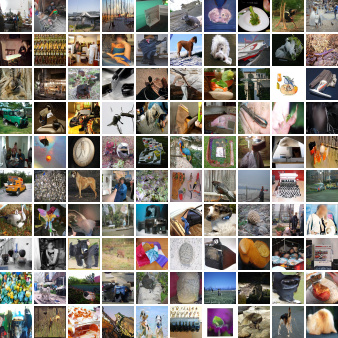} \\ \\
    \Block{1-1}{\rotate \small \hspace{0.16\linewidth}\ourmethod{} (ours)}
    &
    \includegraphics[width=0.35\linewidth]{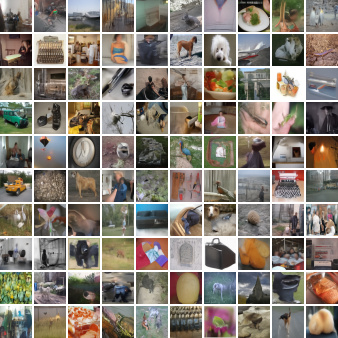} & 
    \includegraphics[width=0.35\linewidth]{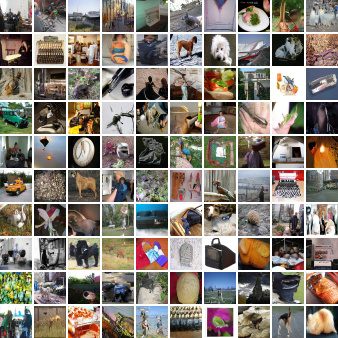} \\
    \end{NiceTabular}
    \caption{\emph{Uncurated} generations in ImageNet-32.
    We compare FM, OT, and \ourmethod{} with both 10-step Euler's method and an adaptive solver for test-time numerical integration.
    }
    \label{fig:app:vis-imagenet32}
\end{figure}

\clearpage

\subsection{ImageNet-256}
\begin{figure}[ht]
    \centering
    \includegraphics[width=0.8\linewidth]{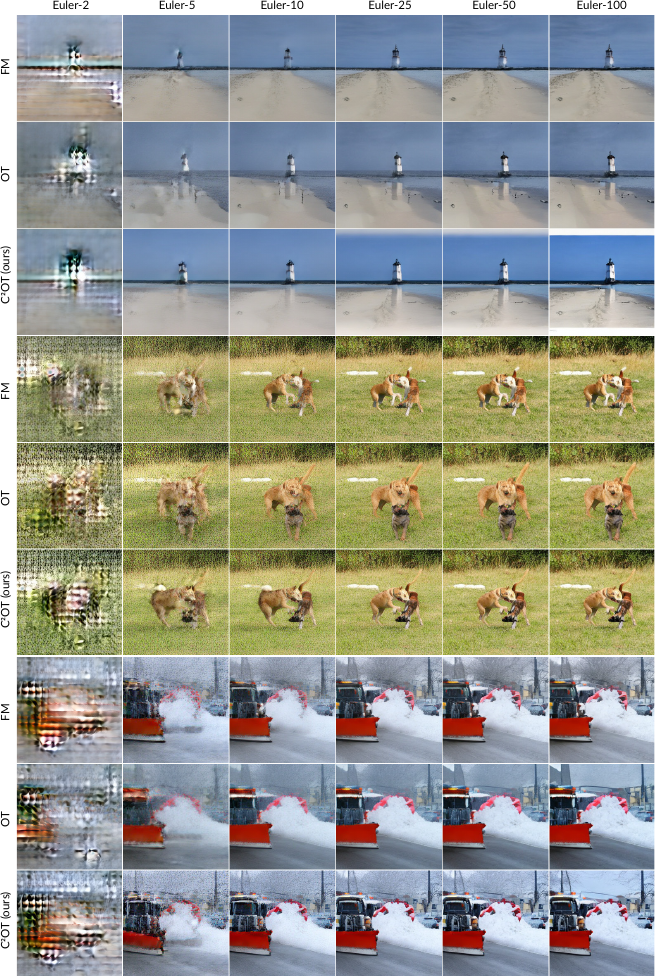}
    \caption{\emph{Uncurated} generations in ImageNet-256.
    We compare FM, OT, and \ourmethod{} with different amounts of sampling steps.
    }
    \label{fig:app:vis-imagenet256-steps}
\end{figure}

\begin{figure}[ht]
    \centering
    \begin{NiceTabular}{c@{}c}
    \Block{1-1}{\rotate \small \hspace{0.16\linewidth}FM}
    &
    \includegraphics[width=0.75\linewidth]{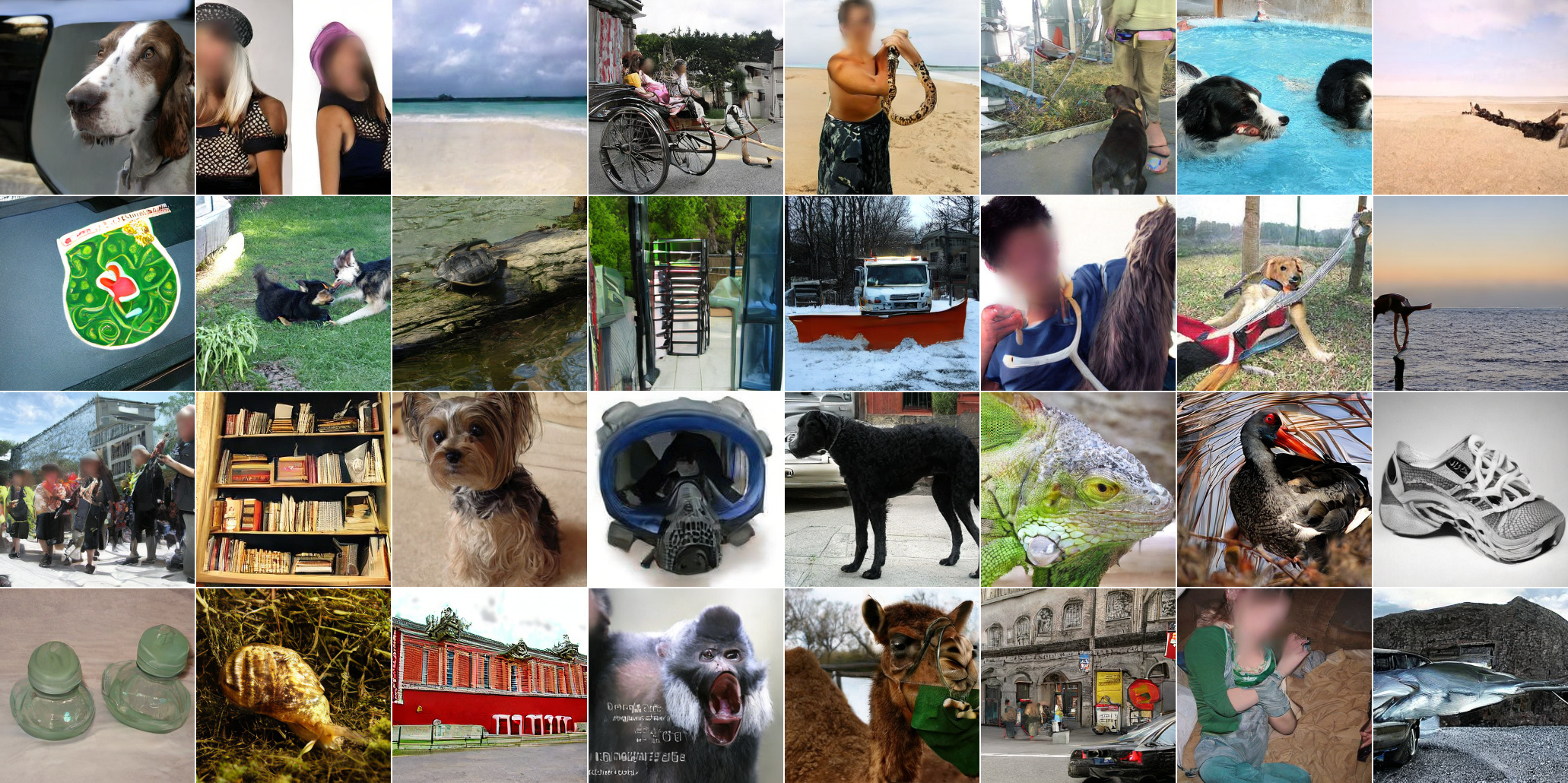} \\ \\
    \Block{1-1}{\rotate \small \hspace{0.16\linewidth}OT}
    &
    \includegraphics[width=0.75\linewidth]{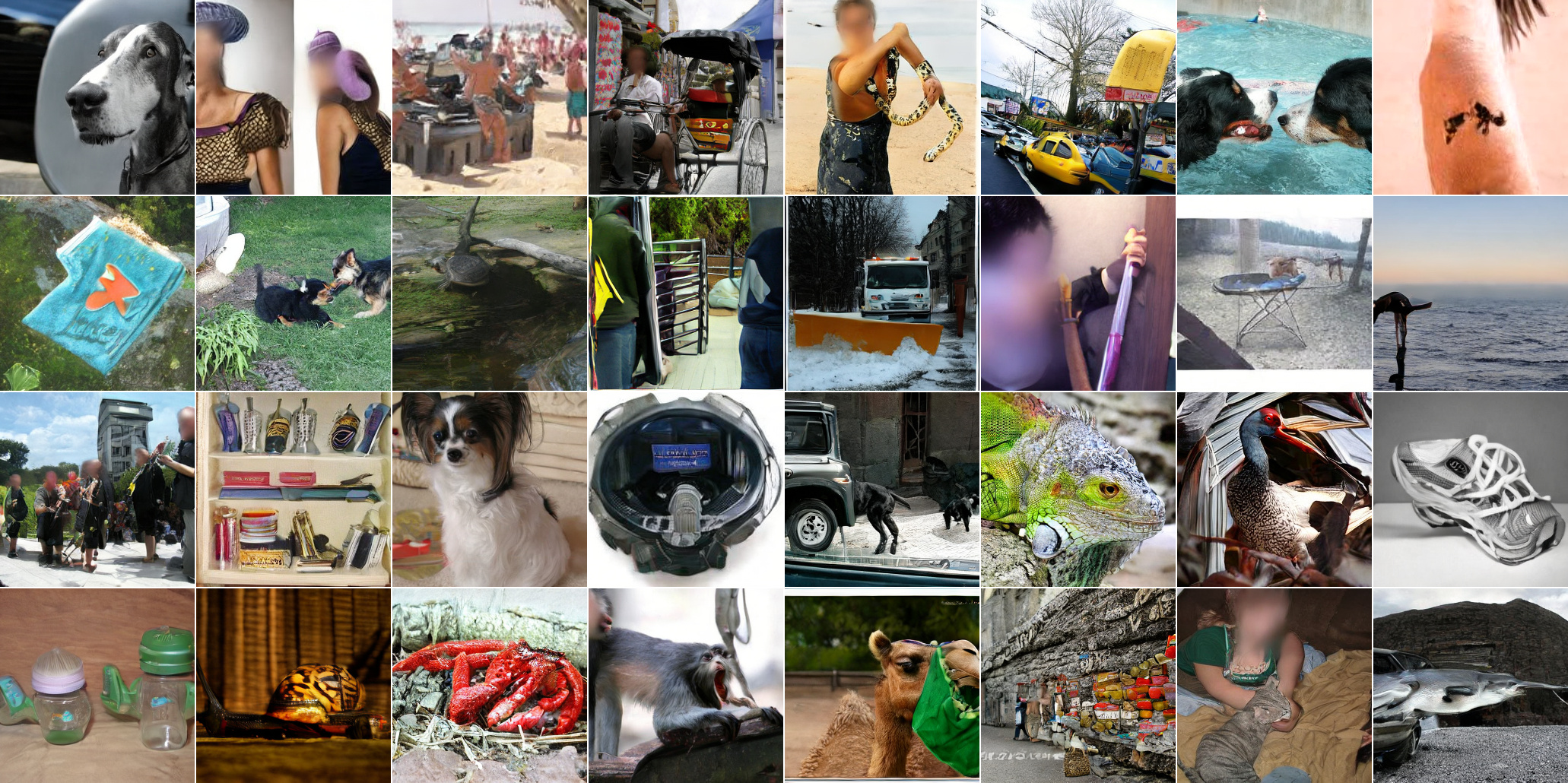} \\ \\
    \Block{1-1}{\rotate \small \hspace{0.16\linewidth}\ourmethod{} (ours)}
    &
    \includegraphics[width=0.75\linewidth]{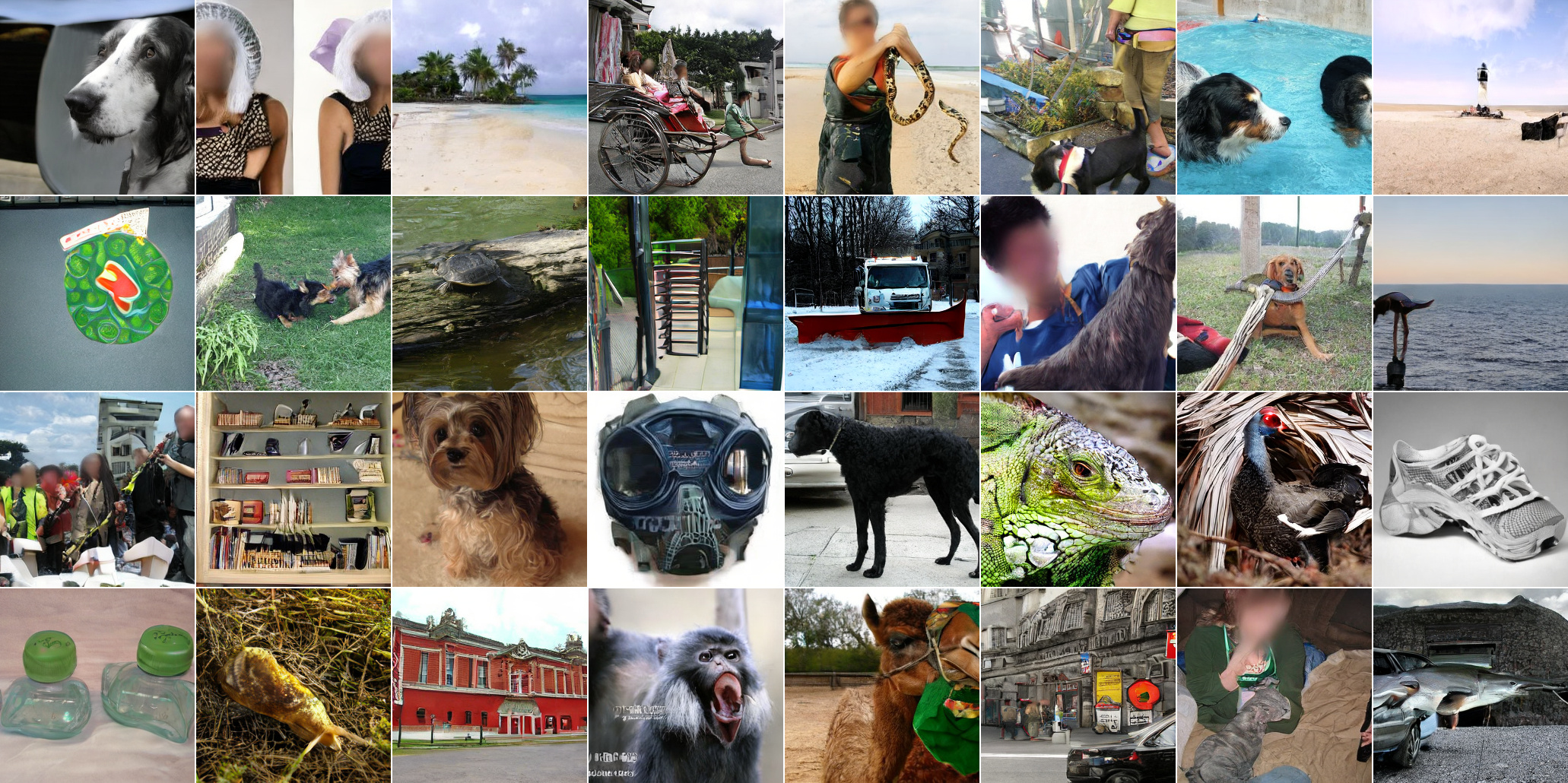} \\ \\
    \end{NiceTabular}
    \caption{\emph{Uncurated} generations in ImageNet-256.
    We compare FM, OT, and \ourmethod{} with an adaptive solver for test-time numerical integration.
    }
    \label{fig:app:vis-imagenet256}
\end{figure}

\clearpage
\section{DeltaAI Acknowledgment}
This work used the DeltaAI system at the National Center for Supercomputing Applications through allocation CIS250008 from the Advanced Cyberinfrastructure Coordination Ecosystem: Services \& Support (ACCESS) program, which is supported by National Science Foundation grants 2138259, 2138286, 2138307, 2137603, and 2138296.

\end{document}